%% file: Main_Document__PDF_.tex
\documentclass[acmsmall,screen]{acmart}

\AtBeginDocument{%
  \providecommand\BibTeX{{%
    \normalfont B\kern-0.5em{\scshape i\kern-0.25em b}\kern-0.8em\TeX}}}

\setcopyright{acmcopyright}
\copyrightyear{2022}
\acmYear{2022}
\acmDOI{ }

\acmConference[ACM Computing Survey]{ACM Computing Survey}
\acmBooktitle{Manuscript}

\usepackage{multirow}
\usepackage{multicol}
\usepackage{url}
\usepackage{tabularx}
\usepackage{makecell}
\usepackage{diagbox}
\usepackage[figuresright]{rotating}
\begin{document}

\newcommand\norm[1]{\left\lVert#1\right\rVert}
\newcommand{\Lapl}{\mathbf{\mathop{\mathcal{L}}}}
\newcommand{\Trans}[1]{{#1}^{\top}}
\newcommand{\Trace}[1]{tr\left({#1}\right)}
\newcommand{\Bracs}[1]{\left({#1}\right)}
\newcommand{\Mat}[1]{\mathbf{#1}}
\newcommand{\MatS}[3]{\mathbf{#1}^{#2}_{#3}}
\newcommand{\Space}[1]{\mathbb{#1}}
\newcommand{\Set}[1]{\mathcal{#1}}
\newcommand{\vectornorm}[1]{\left|\left|#1\right|\right|}
\newcommand{\bpi}{\boldsymbol{\pi}}
\newcommand{\BlockMat}[2]{\left[\begin{matrix}#1\\#2\end{matrix}\right]}
\newcommand{\BlockMatSquare}[4]{\left[\begin{matrix}#1 & #2\\#3 & #4\end{matrix}\right]}

\newcommand{\ie}{i.e., }
\newcommand{\eg}{e.g., }
\newcommand{\etal}{et al.}
\newcommand{\etc}{etc.}
\newcommand{\wrt}{w.r.t. }
\newcommand{\cf}{\emph{cf. }}
\newcommand{\aka}{a.k.a. }

\title{A Comprehensive Survey on Deep Clustering: \\ Taxonomy, Challenges, and Future Directions}

\author{Sheng Zhou, Hongjia Xu, Zhuonan Zheng, Jiawei Chen, Zhao li, Jiajun Bu}
\email{\{zhousheng\_zju, xu\_hj,zhengzn,sleepyhunt, zhao_li, bjj\}@zju.edu.cn}
\affiliation{%
  \institution{Zhejiang University}
  \country{China}
}
\author{Jia Wu}
\email{jia.wu@mq.edu.au}
\affiliation{
\institution{Macquarie University}
\country{Australia}
}

\author{Xin Wang, Wenwu Zhu}
\email{\{xin\_wang,wwzhu\}@tsinghua.edu.cn}
\affiliation{
  \institution{Tsinghua University}
  \country{China}
}
\author{Martin Ester}
\email{ester@cs.sfu.ca}
\affiliation{%
  \institution{Simon Fraser University}
  \country{Canada}
}

\renewcommand{\shortauthors}{Zhou and Xu, et al.}


\begin{abstract}
Clustering is a fundamental machine learning task which has been widely studied in the literature.
Classic clustering methods follow the assumption that data are represented as features in a vectorized form through various representation learning techniques. 
As the data become increasingly complicated and complex, the shallow (traditional) clustering methods can no longer handle the high-dimensional data type.
With the huge success of deep learning, especially the deep unsupervised learning, many representation learning techniques with deep architectures have been proposed in the past decade.
One straightforward way to incorporate the benefit of deep learning is to first learn the deep representation before feeding it into shallow clustering methods.
However, this is suboptimal due to: i) the representation is not directly learned for clustering which limits the clustering performance; 
ii) the clustering relies on the relationship among instances which is complicated rather than linear; iii) the clustering and representation learning is dependent on each other which should be mutually enhanced. 
To tackle the above challenges, the concept of \textbf{Deep Clustering}, i.e., jointly optimizing the representation learning and clustering, has been proposed and hence attracted growing attention in the community.
Motivated by the tremendous success of deep learning in clustering, one of the most fundamental machine learning tasks, and the large number of recent advances in this direction,
in this paper we conduct a comprehensive survey on deep clustering by proposing a new taxonomy of different state-of-the-art approaches.
We summarize the essential components of deep clustering and categorize existing methods by the ways they design interactions between deep representation learning and clustering.
Moreover, this survey also provides the popular benchmark datasets, evaluation metrics and open-source implementations to clearly illustrate various experimental settings.
Last but not least, we discuss the practical applications of deep clustering and suggest challenging topics deserving further investigations as future directions.
\end{abstract}


\maketitle

\clearpage

\section{Introduction}
\input{introduction.tex}

\section{Preliminary}
\input{definition.tex}
\label{sec:definition}

\section{Representation Learning Module}
\label{sec:representation}

\input{representation.tex}

\section{Clustering Module}
\label{sec:clustering}
\input{clustering.tex}

\section{Taxonomy of Deep Clustering}
\label{sec:work}
In this section, we summarize existing deep clustering methods into four branches based on the interaction between the representation learning module and the clustering module:
\begin{enumerate}
    \item Multi-stage deep clustering: the representation learning module is sequential connected with the clustering module.
    \item Iterative deep clustering: the representation learning module and the clustering module are iteratively updated.
    \item Generative deep clustering: the clustering module is modeled as a prior representation module.
    \item Simultaneous deep clustering: the representation learning module and the clustering module are simultaneously updated.
\end{enumerate}

\subsection{Multistage Deep Clustering}
\label{multistage}
\input{multistage.tex}

\subsection{Iterative Deep Clustering}
\input{iterative.tex}

\subsection{Generative Deep Clustering}
\input{generative.tex}

\subsection{Simultaneous Deep Clustering}
\input{simultaneous.tex}

\input{Advantages-and-disadvantages.tex}
\input{application.tex}

\section{Datasets and Evaluation Metrics}
\label{sec:datasets}
\input{dataset-evaluation.tex}

\section{Applications}
\label{sec:application}
\input{real-application.tex}

\section{Future Directions}
\label{sec:future}
\input{future.tex}

\section{Conclusion}
\input{conclusion}


\bibliographystyle{ACM-Reference-Format}
\bibliography{reference}


\end{document}

%% file: introduction.tex
Clustering is a fundamental problem in machine learning and frequently serves as an important preprocessing step in many data mining tasks.
The primary purpose of clustering is to assign the instances into groups so that the similar samples belong to the same cluster while dissimilar samples belong to different clusters.
The clusters of samples provide a global characterization of data instances, which can significantly benefit the further analysis on the whole dataset, such as anomaly detection \cite{saeedi2018novel,wibisono2021multivariate}, domain adaptation \cite{tang2020unsupervised,zhou2021cluster}, community detection \cite{su2022comprehensive,liu2020deep} and discriminative representation learning \cite{yang2016joint, meng2020hierarchical,rezaei2021learning}, etc.

Although shallow clustering methods have achieved tremendous success, they assume that instances are already represented in a latent vectorized space with a good shape.
With the rapid development of internet and web services in the past decades, the research community is showing an increasing interest in discovering new machine learning models capable of handling unstructured data without explicit features, such as images, and high-dimensional data with thousands of dimensions, etc.
As such, shallow clustering methods can no longer be directly applied to deal with such data.
Recent years have witnessed the success of representation learning with deep learning, especially in the unstructured and high-dimensional data \cite{saeedi2018novel,wibisono2021multivariate}. However, the deep learning technique was not explored in the clustering process. The complex relations among instances can not be well captured, which results in sub-optimal clustering results.
To address the issues, \textbf{Deep Clustering}, which aims at joint optimization of deep representation learning and clustering, arises and has attracted increasing attention recently in the community.
More specifically, the deep clustering methods focus on the following research challenges:
\begin{enumerate}
    \item How to learn discriminative representations that can yield better clustering performance?
    \item How to efficiently conduct clustering and representation learning in a unified framework?
    \item How to break down the barriers between clustering and representation learning, enabling them to enhance each other in an interactive and iterative manner?
\end{enumerate}
To tackle the above challenges, numerous deep clustering methods have been proposed with variant deep architectures and data types.
Motivated by the tremendous success of deep learning in clustering, one of the most fundamental machine learning tasks, and the large number of recent advances in this direction,
in this paper we conduct a comprehensive survey on deep clustering by proposing a new taxonomy of various state-of-the-art approaches.

\textbf{Related Surveys} Overview papers on clustering tend to have close relations to our work, however, we clearly state their distinctions from our survey as follows.  
There have been surveys on classic clustering~\cite{frades2010overview,xu2015comprehensive},
particularly summarizing classic clustering methods into the following main groups: 
i) partition based methods \cite{lloyd1982least,park2009simple} that assign the data instances to the closest cluster centers, 
ii) hierarchy based methods \cite{johnson1967hierarchical,zhang1996birch,guha1998cure} that construct the hierarchical relationships among data instances, 
iii) density based methods \cite{ester1996density,sander1998density,comaniciu2002mean,cao2006density,ester2018density} that take the instances in the region with high density as a cluster,
and iv) generative methods \cite{xu1998distribution,rasmussen1999infinite} that assume data belonging to the same cluster are generated from the same distribution.
Generally, these methods take the feature representations as input and output the cluster assignment of each instance.
To the best of our knowledge, very few works have been proposed to survey the deep clustering approaches. 
Min \etal~\cite{min2018survey} in their survey paper states that the essence of deep clustering is to learn clustering-oriented representations, so the literature should be classified according to network architecture. However, we discover that the importance of clustering constraint and the interaction between representation and clustering play central roles in modern deep clustering.
Nutakki \etal~ \cite{nutakki2019introduction} review some basic methods, excluding the most recent advanced techniques for representation learning and clustering. 
Aljalbout \etal~ \cite{aljalbout2018clustering} design the taxonomy mainly based on the DNN architecture and the loss function, discussing the comparisons among methods only on the MNIST and COIL20 datasets. 
Furthermore, all these works \cite{min2018survey,nutakki2019introduction,aljalbout2018clustering} merely cover methods prior to the year of 2018, missing more recent advances in the past four years. 
Therefore, we strongly believe that this survey with a new taxonomy of literature and more than 200 scientific works will strengthen the focus and promote the future research on deep clustering.

\begin{figure}
    \centering
    \includegraphics[width=\textwidth]{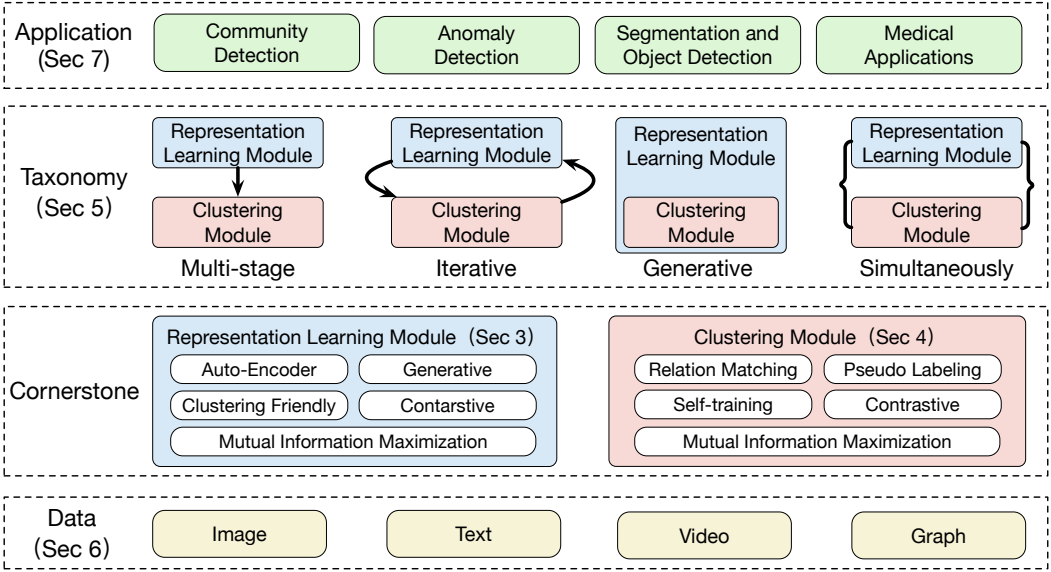}
    \caption{Overall organization of this survey.}
    \label{fig:framework}
\end{figure}

\textbf{Contributions.} In summary, this paper aims at supporting the potential readers understanding the panorama of deep clustering with respect to the following aspects:
\begin{itemize}
    \item \textbf{Cornerstones of Deep Clustering.} We summarize two cornerstones of deep clustering, namely the representation learning module and the clustering module. For each module, we highlight the representative and universal designs summarized from existing methods, which can be easily generalized to new models. 
    \item \textbf{Systematic Taxonomy.} We propose a systematic taxonomy of existing deep clustering methods based on the ways of interactions between representation learning module and clustering module through providing four representative branches of methods. We also compare and analyse the properties of each branch within different scenarios.
    \item \textbf{Abundant Resources and References.} We collect various types of benchmark datasets, evaluation metrics and open-source implementations of latest publications on deep clustering, which are organized together with references on the Github (1.8K Star)\footnote{\url{https://github.com/zhoushengisnoob/DeepClustering}}.
    \item \textbf{Future Directions.} 
    Based on the properties of representation learning module and clustering module as well as their interactions, we discuss the limitations and challenges of existing methods, followed by our insights and thoughts on promising research directions deserving future investigations. 
\end{itemize}

\textbf{Scope.}
In this survey, we focus on the clustering with deep learning techniques, especially the interactions between deep representation learning and clustering with deep neural networks.
For the other fundamental research problems such as initialization of clustering, automatic identifying number of clusters, \etc , we provide discussions in Section \ref{sec:future} and leave them in the future work. The comparisons between surveys on shallow clustering, deep clustering and representation learning can be found in Table \ref{table:comparisons}.

\begin{table*}[!htbp]
\setlength\tabcolsep{1pt}
\footnotesize
\centering
\caption{Comparison with related surveys. 
}
    \begin{tabularx}{\textwidth}{|X|X<{\centering}|p{0.045\textwidth}<{\centering}|p{0.045\textwidth}<{\centering}|p{0.045\textwidth}<{\centering}|p{0.045\textwidth}<{\centering}|p{0.045\textwidth}<{\centering}|p{0.045\textwidth}<{\centering}|p{0.045\textwidth}<{\centering}|p{0.045\textwidth}<{\centering}|p{0.045\textwidth}<{\centering}|p{0.045\textwidth}<{\centering}|p{0.045\textwidth}<{\centering}|p{0.045\textwidth}<{\centering}|p{0.045\textwidth}<{\centering}|p{0.045\textwidth}<{\centering}|}
 \hline
        \multicolumn{2}{|c|}{\multirow{2}*{\diagbox{Comparisons}{References}}}
        & \multicolumn{5}{c|}{Deep Clustering}
        &\multicolumn{4}{c|}{Shallow Clustering}
        &\multicolumn{5}{c|}{Unsupervised Learning}

\\  \cline{3-16}
\multicolumn{2}{|c|}{}	& \textbf{Ours} &\cite{min2018survey}	&\cite{aljalbout2018clustering}	&\cite{nutakki2019introduction}	&\cite{karoly2018unsupervised}	&\cite{xu2015comprehensive}	&\cite{xu2005survey}	&\cite{jain1999data}	&\cite{berkhin2006survey}	&\cite{abukmeil2021survey}	&\cite{le2020contrastive}	&\cite{bengio2013representation}	&\cite{jing2020self}	&\cite{liu2021self}	\\	\cline{1-16}
\multirowcell{4}[0pt][c]{Deep \\ representation \\ learning design}&Auto Encoder	&\checkmark	&\checkmark	&\checkmark	&\checkmark	&\checkmark	&-	&-	&-	&-	&\checkmark	&-	&\checkmark	&\checkmark	&\checkmark	\\	\cline{2-16}
&Generative	&\checkmark	&\checkmark	&\checkmark	&-	&\checkmark	&-	&-	&-	&-	&\checkmark	&\checkmark	&\checkmark	&\checkmark	&\checkmark	\\	\cline{2-16}
&Contrastive	&\checkmark	&-	&-	&-	&-	&-	&-	&-	&-	&-	&\checkmark	&-	&\checkmark	&\checkmark	\\	\cline{2-16}
&Cluster Based	&\checkmark	&\checkmark	&\checkmark	&-	&-	&-	&-	&-	&-	&-	&\checkmark	&-	&\checkmark	&\checkmark	\\	\cline{1-16}
\multicolumn{2}{|c|}{Clustering with DNN}&\checkmark	&\checkmark	&\checkmark	&\checkmark	&\checkmark	&-	&\checkmark	&\checkmark	&\checkmark	&-	&-	&-	&-	&-	\\	\cline{1-16}
\multirowcell{4}[0pt][c]{Interaction \\ between \\ Representation \\ and Clustering}&Multistage	&\checkmark	&-	&\checkmark	&\checkmark	&\checkmark	&\checkmark	&\checkmark	&\checkmark	&\checkmark	&\checkmark	&-	&-	&-	&-	\\	\cline{2-16}
&Iterative	&\checkmark	&-	&\checkmark	&\checkmark	&-	&-	&\checkmark	&\checkmark	&-	&-	&\checkmark	&-	&\checkmark	&\checkmark	\\	\cline{2-16}
&Generative	&\checkmark	&\checkmark	&-	&-	&\checkmark	&-	&-	&-	&-	&-	&-	&-	&-	&-	\\	\cline{2-16}
&Simultaneously	&\checkmark	&\checkmark	&\checkmark	&\checkmark	&-	&-	&-	&-	&-	&-	&-	&-	&-	&-	\\	\cline{1-16}
\multicolumn{2}{|c|}{Application}	&\checkmark	&-	&-	&- &-	&-	&\checkmark	&\checkmark	&\checkmark	&\checkmark	&\checkmark	&\checkmark	&\checkmark	&-	\\	\cline{1-16}
\multicolumn{2}{|c|}{Dataset}	&\checkmark	&-	&-	&- &-	&-	&-	&-	&-	&-	&\checkmark	&-	&-	&-	\\	\cline{1-16}
\multicolumn{2}{|c|}{Evaluation}	&\checkmark	&\checkmark	&-	&- &-	&\checkmark	&-	&-	&-	&-	&-	&-	&\checkmark	&-	\\	\cline{1-16}
\multicolumn{2}{|c|}{Implementation}	&\checkmark	&-	&-	&- &-	&-	&-	&-	&-	&-	&-	&-	&-	&-	\\	\hline

\multicolumn{16}{l}{\multirow{3}*{\makecell[l]{* Each column indicates a survey paper being compared,  `\checkmark' means the term of corresponding row has been surveyed or\\ analysed in this paper while `-' means not.}} }\\
\multicolumn{16}{l}{}\\

\end{tabularx}%

\label{table:comparisons}
\end{table*}

\textbf{Organization.}
The rest of this survey is organized as follows:
Section \ref{sec:definition} introduces the basic definitions and notations used in this paper.
Section \ref{sec:representation} summarizes the representative design of representation module, along with different data types.
Section \ref{sec:clustering} summarizes the representative design of clustering module, which mainly focuses on the 
basic modules defined in the deep clustering methods.
Section \ref{sec:work} summarizes the representative ways of interactions between the two modules, which covers most existing literature. 
Section \ref{sec:datasets} introduces the widely used benchmark datasets and evaluation metrics. 
Section \ref{sec:application} discusses the applications of deep clustering.
Section \ref{sec:future} discusses limitations, challenges, and suggests future research directions that deserve further explorations.
The overall organization of this survey is illustrated in Figure \ref{fig:framework}.





%% file: definition.tex
In this section, we first briefly introduce some definitions in deep clustering that need to be clarified, then we illustrate the notations used in this paper in Table \ref{table:notation}.

\begin{table*}[]
\renewcommand\arraystretch{1.1}
\centering
\caption{Important notations used in this paper}
\scalebox{0.97}{

\begin{tabular}{c c | c c}
\hline
Notations & Explanations & Notations & Explanations \\ \hline
$N$  &  The number of data instances & $K$  &  The number of clusters  \\ \hline
$x$  &  Data instance  &  $\hat{x}$  &  Data reconstruction  \\ \hline
$x^{\mathcal{T}}$  &  Augmented instance  &  $\cdot^T$  &  Transpose of a matrix(vector) \\ \hline
$h$  &  Representation  &  $z$  &  Soft assignment  \\ \hline
$\tilde{y}$  &  Predicted hard label  &  $y$  &  Ground truth label  \\ \hline
$c$  &  Representation(centroid) of cluster  &  $\tau$  &  Temperature parameter   \\ \hline
$\mathbb{P,Q}$  &  Probability distribution  &  $\mathcal{X}=\{x_{i}\}^{N}_{i=1}$  &  Data instances set   \\ \hline
$\|\cdot\|_F$  &  Frobenius norm of a matrix(vector)  &  $|\cdot|$  &  The number of elements in a set  \\ \hline
\end{tabular}
}
\label{table:notation}
\end{table*}

\textbf{Deep Clustering and Shallow (Non-deep) Clustering.}
Given a set of data instances $\mathcal{X}=\{x_{i}\}^{N}_{i=1}$, clustering aims to automatically assign each instance $x$ into groups so that instances in the same group are similar while instances from different groups are dissimilar.
The \textit{shallow (non-deep) clustering} takes the feature vectors of instances as input and outputs the clustering result without deep neural networks.
The \textit{deep clustering} aims at clustering unstructured data or high-dimensional data with deep neural networks.
It is worth noting that deep clustering is not narrowly defined as applying deep learning techniques in representation learning, 
instead, the clustering itself can be conducted by deep neural networks and benefits from the interaction with deep representation learning.

\textbf{Hard Clustering and Soft Clustering.}
The clustering methods can be categorized into hard clustering and soft clustering according to the output type.
The output of hard clustering is a discrete one-hot cluster label $\tilde{y}_{i}$ for each instance $x_{i}$, while the output of soft clustering is a continuous cluster assignment probabilistic vector $z_{i}\in \mathcal{R}^{K}$.
The discrete assignment of instance is usually hard to optimize, especially for deep neural networks with backpropagation. 
As a result, most existing deep clustering methods belong to the soft clustering category where the clustering results are produced by a deep neural network $f$ whose outputs are softmax activated $K$-dimension logits. 
For the final evaluation purpose and discrete clustering category, the hard label can be obtained by selecting the dimension with maximum probability.

\textbf{Partitioning Clustering and Overlapping Clustering.}
The above-defined hard clustering by maximizing the probability vector is designed for the \textit{partitioning clustering} task where each data instance belongs to only one cluster.
This means the clustering method participates the whole dataset into $K$ disjoint groups.
The majority of existing deep clustering methods are disjoint since the major evaluated datasets are disjoint ones.
For the overlapping clustering setting, each data instance may belong to more than one cluster.
This has brought extra difficulty to the clustering methods which will be discussed in section \ref{sec:future}.



%% file: representation.tex
\begin{figure}
    \centering
    \includegraphics[width=\textwidth]{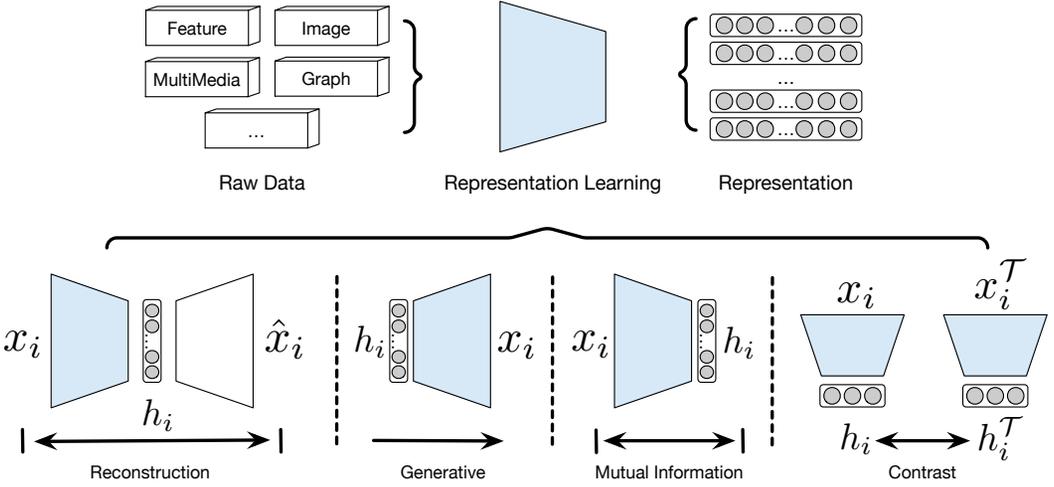}
    \caption{Representative representation learning modules}
    \label{fig:representation}
\end{figure}

Recent decades have witnessed the rapid development of deep representation learning \cite{bengio2013representation,zhou2022network,usman2019survey}, especially the unsupervised ones.
Intuitively, all unsupervised representation learning methods can serve as an input generator and be directly incorporated into deep clustering framework (further discussion on this issue in section \ref{sec:clustering}).  
However, most existing methods are not intently designed for clustering task, and unable to integrate the potential clustering information to learn better representations.
In this section, we introduce the \textit{representation learning module} in deep clustering, which takes the raw data as input and outputs the low-dimensional representation (\aka embedding).
Figure \ref{fig:representation} illustrates the representative representation learning modules described in this section.

\subsection{Auto-Encoder based Representation Learning.}
Auto-Encoder \cite{rumelhart1985learning} is one of the most widely adopted unsupervised deep representation learning methods for its simplicity and effectiveness.
The auto-encoder is a linear stack of two deep neural networks named encoder and decoder. 
The encoder $\mathbf{f}_{e}$ encodes the input data $x$ into low-dimensional representation $h = \mathbf{f}_{e}(x)$, and
the decoder $\mathbf{f}_{d}$ decodes the low-dimensional representation $h$ to the input data space $\hat{x}=\mathbf{f}_{d}(h)$.
A good auto-encoder is expected to reconstruct the input data without dropping any information, thus the optimization target of auto-encoder is usually formulated as:
\begin{equation}
    \mathcal{L}_{AE}=\sum_{i}^{N}\left\|x_{i}-\hat{x}_{i}\right\|_{2}^{2} = \sum_{i}^{N}\left\|x_{i}-\mathbf{f}_{d}\left(\mathbf{f}_{e}\left(x\right)\right)\right\|_{2}^{2}
\end{equation}
where $\left\|\cdot\right\|_{2}$ is the L2-norm, $\hat{x}_{i}$ is the reconstructed data.
The auto-encoder is a general structure and can be customized for different data types.
For example, deep neural networks for vectorized features \cite{zhai2018autoencoder}, convolutional networks for images and graph neural networks for graphs \cite{pan2018adversarially}, 3D convolutional networks and LSTM auto-encoder for videos \cite{srivastava2015unsupervised} \etc.

\textbf{Analysis.}
The auto-encoder based representation learning framework enjoys the property of easy implementation and efficient training, and has been widely adopted in the early works of deep clustering.
However, the representations are learned in an instance-wise manner while the relationships among different instances are largely ignored.
As a result, the instances in the embedding space may not be well discriminated from each other, resulting in poor clustering performance.


\subsection{Deep Generative Representation Learning.}
Another line of deep unsupervised representation learning lies in the generative model. 
Generative methods assume that the data $x$ are generated from latent representation $h$ and then reversely deduce the posterior of the representation $p(h|x)$ from the data. 
Among these, the most typical method is variational auto-encoder (VAE) \cite{kingma2013auto}. VAE resorts to the variance inference technique and maximizes the evidence lower bound (ELBO) on the data likelihood:
\begin{equation}
\log p(x) \geq\mathbb{E}_{q(h|x)}[\log p(x|h)]-D_{K L}(q(h|x) \| p(h)) \label{eq:vae}
\end{equation}
where $D_{KL}{(\cdot\|\cdot)}$ denotes the KL-divergence between two distributions, $p(h)$ is the prior distribution of the latent representation, $q(h|x;\varphi)$ is the variational posterior of the representation to approximate the true posterior (\textit{i.e.,} $q(h|x;\varphi)\approx p(h|x)$), which can be modelled with the recognition networks $\varphi$. By utilizing the reparameterization trick \cite{kingma2013auto} and Monte Carlo approximation \cite{kalos2009monte}, the posterior can be efficiently learned from the equation (\ref{eq:vae}) via backpropagation. 

\textbf{Analysis.} Deep generative models enjoy some merits such as being flexible, explainable and capable to recreate data points. It would be promising to transform the generative representation models for deep clustering task so that the clustering models could inherit these advantages.

\subsection{Mutual Information Maximization Representation Learning.}
\label{subsec:mi_rl}
Mutual information (MI) \cite{kinney2014equitability} is a fundamental quantity for measuring the dependence between random variables $X$ and $Y$, which is formulated as:
\begin{equation}
\mathcal{I}(X ; Y)=\int \log \frac{d \mathbb{P}_{X Y}}{d \mathbb{P}_{X} \otimes \mathbb{P}_{Y}} d \mathbb{P}_{X Y}
\end{equation}
where $\mathbb{P}_{X Y}$ is the joint distribution, $\mathbb{P}_{X} = \int_{Y} d \mathbb{P}_{X Y}$ and $\mathbb{P}_{Y} = \int_{X} d \mathbb{P}_{X Y}$ are the marginal distribution, $\mathbb{P}_{X} \otimes \mathbb{P}_{Y}$ is the product of the marginal distributions.
Traditional mutual information estimations \cite{kraskov2004estimating} are only tractable for discrete
variables or known probability distributions.
Recently, MINE \cite{belghazi2018mutual} is proposed to estimate the mutual information with deep neural networks.
The widely used mutual information estimation is the Jensen-Shannon divergence (JSD) \cite{nowozin2016f} formulated as:
\begin{equation}
    \mathcal{I}_{JSD}(X;H)=\mathbb{E}_{\mathbb{\mathbb{P}}_{X H}}[-\operatorname{sp}(-D(x, h))]-\mathbb{E}_{\mathbb{P}_{X}\times \mathbb{P}_{H}}[\operatorname{sp}(D(x, h))]
\end{equation}
where $\operatorname{sp}(x)=\log \left(1+e^{x}\right)$ is the softplus function.
$D$ is a discrimintor function modeled by a neural network.
Another popular mutual information estimation is InfoNCE \cite{oord2018representation} which will be introduced in subsection \ref{subsec:contrastive}.
Benefiting from the neural estimation, the mutual information has been widely adopted in unsupervised representation learning \cite{hjelm2018learning,bachman2019learning}.
More specifically, the representation is learned by maximizing the mutual information between different layers \cite{bachman2019learning} or different parts of the data instances \cite{hjelm2018learning}, so that the consistency of representation can be guaranteed.
This can be viewed as an early attempt at self-supervised learning which has an extensive impact on the later works.


\textbf{Analysis.}
As a fundamental measure of correlation and dependence, the mutual information has several advantages.
The major advantage taken by the deep clustering task is that variables measured by mutual information are not restricted to same dimension and semantic space, such as instances and clusters.
The detailed applications will be introduced in subsection \ref{subsec:mi_clustering} and \ref{subsec:mi_model}.
Similar to auto-encoder based and deep generative representation learning, the objective of mutual information maximization methods is also instance-wise, which may also have the aforementioned problems in capturing the relationships among instances.
However, the marginal distribution in mutual information estimation depends on all the observed samples.
In other words, the relationships among instances are captured in an implicit way, which has also boosted the performance in deep clustering.

\subsection{Contrastive Representation Learning}
\label{subsec:contrastive}
Contrastive learning is one of the most popular unsupervised representation learning techniques in the past few years.
The basic idea is to pull positive pair of instances close while push negative pair of instances far away, which is also known as instance discrimination.
The representative target of contrastive learning is the InfoNCE loss \cite{oord2018representation} formulated as:
\begin{equation}
    \mathcal{L}_{InfoNCE}=-\log \sum^{N}_{i=1}\frac{\exp \left(f\left(h_{i}, h^{\mathcal{T}}_{i}\right) / \tau\right)}{\sum_{j=1}^{N} \exp \left(f\left(h_{i}, h_{j}^{\mathcal{T}}\right) / \tau\right)}
\end{equation}
where $h_{i}$ is the representation of the anchor sample, $h^{\mathcal{T}}_{i}$ is the positive sample representation and $h_{j}$, $h_{j}^{\mathcal{T}}$ are the negative sample representations,  $f$ is a similarity function, $\tau$ is the temperature parameter \cite{hinton2015distilling}.
The positive samples are usually conducted by data augmentation which may vary from different data types.
For example, the flip, rotate and crop augmentation for image data \cite{chen2020simple}, the node dropping, edge perturbation, attribute masking and subgraph sampling for graph data \cite{you2020graph,li2021disentangled}.
The negative samples are selected from augmented view of other instances in the dataset \cite{chen2020simple} or a momentum updated memory bank of old negative representations \cite{he2020momentum}, which can be viewed as approximation of noise.

\textbf{Analysis.}
There has been several theoretical analyses on contrastive learning, and substantial evidences have shown that the representation learned by contrastive learning can benefit the clustering task.
In \cite{wang2020understanding}, contrastive learning is explained with two properties: alignment of features from positive pairs and uniformity of the feature distribution on the hypersphere.
The alignment property encourages the samples with similar features or semantic categories to be close in the low-dimensional space, which is essential for clustering.
Such discriminative power has also been proved in the supervised manner \cite{khosla2020supervised}.

The former work \cite{oord2018representation} has proved that minimizing the InfoNCE loss is equivalent to maximizing the lower bound of mutual information, while the data augmentation is not considered in the mutual information maximization.
Recent study \cite{wang2021chaos} has shown that the augmentation has played the role of `ladder' in connecting instances within same category. As a result, instances from same cluster may be pulled closer which can benefit the clustering in the discriminative space.

\subsection{Clustering Friendly Representation Learning.}
\label{subsec:cluster_friend}
Although the aforementioned representation learning methods have somehow implicitly boosted the performance of clustering, they are not explicitly designed for the clustering task.
In this subsection, we will introduce some representation learning methods that explicitly support the clustering task.

K-means \cite{lloyd1982least} friendly representation is first defined in DCN \cite{yang2017towards} where samples in the low-dimensional space are expected to scatter around their corresponding cluster centroids.
This can well support the assumption of K-means that each sample is assigned to the cluster with the minimum distance to the centroid.
The objective $\mathcal{L}_{KF}$ can be formulated as:
\begin{equation}
    \mathcal{L}_{KF} = \sum^{N}_{i=1}\left\|\mathbf{f}\left(x_{i}\right)-\tilde{y}_{i}\mathbf{C} \right\|_{2}^{2}
\end{equation}
where $\mathbf{f}(\cdot)$ is the deep neural network for representation learning, 
$\tilde{y}_{i}$ is the hard assignment vector of data instance $x_{i}$ which has only one non-zero elements, 
$\mathbf{C}$ is the cluster representation matrix where $k$-th column of $\mathbf{C}$, i.e., $c_{k}$ denotes the centroid of the $k$-th cluster.

Spectral clustering friendly representation learning is inspired by the eigen decomposition in spectral clustering \cite{ng2001spectral} that projects instances into the space with orthogonal bases.
In deep clustering, the orthogonal basis is modeled explicitly by reducing correlations within features \cite{tao2021clustering}, which can be formulated as:
\begin{equation}
    \mathcal{L}_{SF}=\sum_{m=1}^{d}\left(-h_{m}^{T} h_{m} / \tau+\log \sum_{n}^{d} \exp \left(h_{n}^{T} h_{m} / \tau\right)\right)
\end{equation}
where $d$ is the number of feature dimensions, $h_{m}$ is the $m$-th dimension feature vector, $\tau$ is the temperature parameter.
The objective is to learn the independent features so that redundant information is reduced.

\textbf{Analysis.}
The clustering friendly representation learning benefits from the direct optimization for clustering, which may significantly boost the corresponding clustering performance.
However, such simplicity also limits the generalization to other clustering methods.
Currently, the research community has put more efforts on the inspirations of clustering methods and express them in a deep learning perspective, rather than learning specific representations for each clustering method.

\subsection{Subspace Representation Learning}
\label{subsec:subspace}
Subspace representation learning is the early stage of subspace clustering \cite{vidal2009sparse}, which aims at mapping the data instances into a low-dimensional subspace where instances can be separated. 
Basically, current subspace representation learning methods \cite{zhang2019self,zhang2019neural,ji2017deep,zhang2021learning,peng2016deep,zhou2018deep} have relied on the self-expressiveness assumption where a data instance can be expressed as a linear combination of other data instances from the same subspace,
i.e., $\mathbf{X=X \Theta_c}$, where $\mathbf{X}$ is the data matrix and $\mathbf{\Theta_c}$ is the self-expression coefficient matrix. For representation learning, the self-expression property leads to the following objective:
\begin{equation}
    \mathop{\text{min}}_\mathbf{\Theta_c}\ \|\mathbf{\Theta_c}\|_p + \frac{\lambda}{2} \|\mathbf{H-H \Theta_c} \|_F^2 \quad \text{s.t.} \quad \text{diag}(\mathbf{\Theta_c})=\textbf{0}
\end{equation}
where $\|\cdot\|_p$ is a matrix norm, $\lambda$ controls the weight balance. $\textbf{H}$ denotes the sample representations learned by a network. Further, $\mathbf{\Theta_c}$ can be implemented as the parameters of an additional network layer \cite{ji2017deep}.

\textbf{Analysis.}
The success of subspace representation learning relies on the theoretical assurance, which provides explicit modeling of relationships among data instances.
However, it suffers from the limitation of solving the coefficient matrix of size $N \times N$, which is computationally difficult for large-scale data.

\subsection{Data type specific representation learning}
The former subsections have summarized the universal architectures of representation learning module.
In real-world scenarios, representation learning for different data types can be viewed as the variants of the aforementioned architectures.
In this subsection, we summarize four widely studied data types and their corresponding representation learning methods in deep clustering.

\subsubsection{Image Representation Learning}
Learning representations of images using CNN \cite{lecun1998gradient} and ResNet \cite{he2016deep} as backbone has achieved great success in the past decades.
In the image deep clustering, they still play active roles as feature extractors or backbones in representation learning module.
Beyond the above two methods, recent advances have been made by introducing modern representation learning techniques such as vision transformer \cite{dosovitskiy2020image} to deep clustering.
As one of most popular directions, the unsupervised representation learning for image data will play a central role in deep clustering and affect the other data types.

\subsubsection{Text Representation Learning}
The early attempts of text representation learning have utilized the statistical based methods like TF-IDF \cite{steinbach2000comparison}, Word2Vec\cite{le2014distributed} and Skip-Gram \cite{mikolov2013efficient}.
Later, some works focus on the topic modeling \cite{hofmann2013probabilistic} and semantic distances \cite{dzisevivc2019text,prasad2019hybrid} for text representation learning, and more \cite{chung2019unsupervised} on unsupervised scenarios.
Recently, the pre-trained language models like BERT \cite{devlin2018bert} and GPT-3 \cite{brown2020language} are gradually dominating the area of text representation learning.
However, the fine tuning \cite{zhang2021fast} of these methods in deep clustering task is still an open question. 

\subsubsection{Video Representation Learning}
The video representation learning is a challenging task which combines the spatial-temporal learning, multi-model learning (with audio) \cite{alwassel2020self}, and natural language processing (with video abstract and subtitles) into one place.
The early methods utilize LSTM Autoencoder \cite{srivastava2015unsupervised}, 3D-ResNets \cite{qian2021spatiotemporal} and 3D-U-Net \cite{peng2021deep} as feature extractor.
The recent methods have focused on spatial-temporal modeling \cite{feichtenhofer2021large,srivastava2015unsupervised,zhang2016unsupervised,zhang2018salient} and Qian \etal~ \cite{qian2021spatiotemporal} in particular incorporates contrastive learning for self-supervision.

\subsubsection{Graph Representation Learning}
\label{community}
The classic graph representation learning aims at learning low dimensional representation for nodes so that the proximity among nodes can be preserved in the embedding space.
Graph Neural Networks (GNNs) \cite{zhang2020deep,wang2016structural} are widely used including 
GCN \cite{kipf2016semi}, GraphSAGE \cite{hamilton2017inductive} and GAT \cite{velivckovic2017graph}, brings infinite possibility of graph node representation learning combining node features and graph topology\cite{wang2021explainable,zhang2020learning,zhu2020deep,zhou2020dge}.
Furthermore, the graph-level information also has great potential in tasks like proteins classification \cite{borgwardt2005protein}, which has drown increasing attention in graph-level representation learning \cite{zhang2019hierarchical,zhang2018anrl,wu2014bag,guo2017combining}.

\textbf{Analysis.}
The data type specific representation learning mentioned above can be naive backbone for feature extraction or end-to-end unsupervised representation learning, which are most active research directions in deep learning.
With more types of data being collected and fast development of deep learning, we believe that the deep clustering with grow along with the data type specific representation learning techniques.

%% file: clustering.tex
In this section, we introduce the representative clustering modules in deep clustering, which takes the low-dimensional representations as input, and outputs the cluster labels for hard clustering or cluster assignment probabilities for soft clustering.
Although many shallow clustering methods can be directly employed for clustering, they are hard to be trained with deep representation learning in a unified framework.
More importantly, they can not well interact with representation learning module and mutually enhanced.
For more shallow clustering methods, please refer to the former surveys \cite{jain1999data,xu2005survey}.
In deep clustering, a more `deep' way of clustering is as follows:
the vectorized features are directly fed forward through deep neural networks and reduce the dimension to the cluster number $k$, then the softmax function is applied on the last layer so that the assignment distribution can be established.

\begin{figure}
    \centering
    \includegraphics[width=\textwidth]{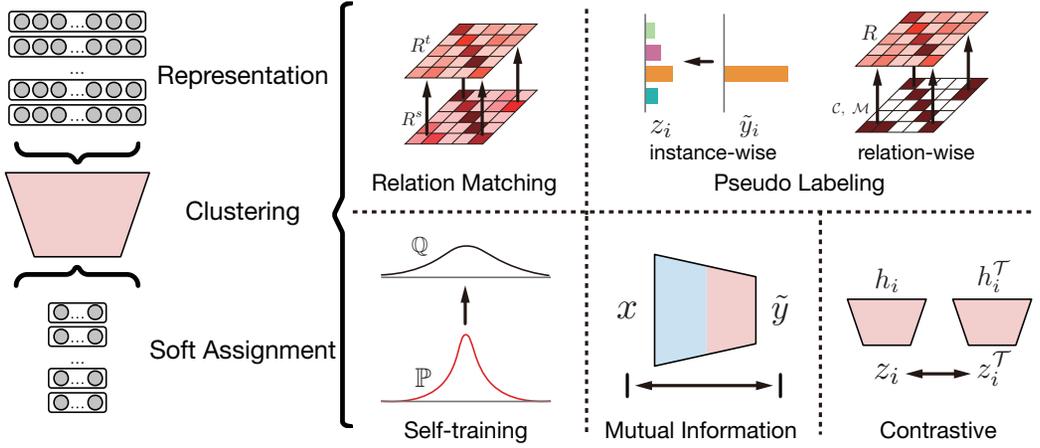}
    \caption{Representative clustering modules}
    \label{fig:clustering}
\end{figure}

Although the $k$-dimensional representations are in the form of probability distribution, they may not represent the cluster distribution without explicit constraints and intensify the \textit{degenerate problem} where all instances are assigned to the same cluster.
In this section, we assume the clustering process is conducted by the deep neural networks and focus on the clustering targets.
Figure \ref{fig:clustering} illustrates the representative clustering module described in this section.

\subsection{Relation Matching Deep Clustering}
In deep clustering, each data instance can be represented in two spaces, namely the $d$-dimensional \textit{embedding space} and $K$-dimensional \textit{label space}.
The relationships among instances are expected to be consistent during dimension reduction, which has been utilized to bridge the semantic gap between representation learning and clustering.
Borrowing the idea from the domain adaptation \cite{long2015learning}, relation matching can be realized in bidirectional ways:
\begin{equation}
    \mathcal{L}_{RM} = \sum^{N}_{i}\sum^{N}_{j}\ell(R^{s}_{ij},R^{t}_{ij})
\end{equation}
where $\ell$ is a measure of relation matching, \eg cosine similarity or Euclidean distance, $R^{s}$ and $R^{t}$ are the relations in source and target space, which can be either embedding space or label space.
It is worth noting that the relations defined here narrowly refers to the continuous ones and we introduce the discrete relations as pseudo labeling in subsection \ref{subsec:pseudo}.

\textbf{Analysis.} 
The relation matching deep clustering explicitly connect the representation learning and clustering, which is straightforward and easy to implement.
However, calculating $N^{2}$ pairs of instances is computational inefficient.
To tackle this challenge, some methods only preserve the k-nearest-neighbor relations \cite{van2020scan,dang2021nearest} for each instance or the relations with high confidence \cite{van2020scan}.
Although this can somehow improve the training efficiency, the extra hyper-parameter is hard to set in the unsupervised manner.
Furthermore, among all pairs of relations, many of them are noisy especially in the early training phases with limited capability. How to filter out the clean relations to boost the performance while discard the noisy relations is still an open research question.

\subsection{Pseudo Labeling Deep Clustering}
\label{subsec:pseudo}
The pseudo labeling can be viewed as another type of relation matching where the relations are discrete based on the consistency of labels.
It has been widely studied in semi-supervised learning \cite{lee2013pseudo}, and recently being introduced to deep clustering.
According to the way of utilizing pseudo labels, existing methods can be largely divided into two groups: instance-wise pseudo labeling \cite{van2020scan,niu2021spice,caron2018deep} and relation-wise pseudo labeling \cite{chang2017deep,niu2020gatcluster}.

The instance-wise pseudo labeling filters out a subset of instances with high confidence and trains the network in a supervised manner with cross-entropy loss:
\begin{equation}
    \mathcal{L}_{IPL} = -\frac{1}{|\mathcal{X}^{c}|} \sum^{\mathcal{X}^{c}}_{i} \sum_{k=1}^{K} \tilde{y}_{i k} \log \left(z_{i k}\right)
\end{equation}
where $\mathcal{L}_{IPL}$ denotes the loss of instance-wise pseudo labeling, $\mathcal{X}^{c}$ denotes the subset of instances with high confidence, $\tilde{y}_{i k}$ and $z_{i k}$ are the predicted hard label and soft cluster assignment.
The confidence is usually estimated by entropy or maximum of the assignment probabilistic distribution.

The general idea of relation-wise pseudo labeling is to enforce instances with same pseudo labels closer while instances with different pseudo labels away from each other in the embedding space.
Given the filtered instances, relation-wise pseudo labeling construct the discrete relations among instances to guide the representation learning: the \textit{must-link} for pairs of instances with same pseudo labels and \textit{cannot-link} for pairs of instances with different pseudo labels:
\begin{equation}
    \mathcal{L}_{RPL} = \frac{1}{|\mathcal{C}|}\sum_{\{i,k\}\in \mathcal{C}}  R_{ik} - \frac{1}{|\mathcal{M}|}\sum_{\{i,j\}\in \mathcal{M}}  R_{ij}
\end{equation}
where $\mathcal{M}$ is the set of must-link relations and $\mathcal{C}$ is the set of cannot-link relations, 
$R_{ij}$ is the similarity of instance $x_{i}$ and $x_{j}$ in the low-dimensional embedding space.

\textbf{Analysis.}
The pseudo labeling has brought the power of semi-supervised learning into the unsupervised clustering task.
However, the performance highly relies on the quality of filtered pseudo labels which is susceptible to model capability and hyper-parameter tuning, especially in the unsupervised manner. The current methods \cite{van2020scan,niu2021spice} have taken the pre-training as the early stage before pseudo labeling to tackle this challenge, but more attention deserved to be paid on it.

\subsection{Self-training Deep Clustering}
\label{subsec:self-training}
The self-training strategy is introduced to the deep clustering task \cite{xie2016unsupervised} and open up an active branch of methods named self-training deep clustering. 
More specifically, the cluster assignment distribution is optimized by minimizing the KL-divergence with an auxiliary distribution:
\begin{equation}
    \mathcal{L}=D_{K L}(\mathbb{P} \| \mathbb{Q})=\sum^{N}_{i} \sum^{K}_{k} p_{i k} \log \frac{p_{i k}}{q_{i k}}
\end{equation}
where $\mathbb{Q}$ is the cluster assignment distribution and $\mathbb{P}$ is the auxiliary distribution.
$q_{ik}$ and $p_{ik}$ denotes the probability of instance $x_{i}$ belong to cluster $k$.
The assignment distribution $\mathbb{Q}$ follows the assumption of K-means and is produced by the embedding distance between instance and cluster centroids:
\begin{equation}
    q_{i k}=\frac{\left(1+\left\|h_{i}-c_{k}\right\|_{2}^{2} / \alpha\right)^{-\frac{\alpha+1}{2}}}{\sum^{K}_{j}\left(1+\left\|h_{i}-c_{j}\right\|_{2}^{2} / \alpha\right)^{-\frac{\alpha+1}{2}}}
\end{equation}
where $h_{i}$ is the representation of data instance $x_{i}$ and $c_{k}$ is the representation of cluster $k$, $\alpha$ is the freedom degree of the Student's t-distribution \cite{van2008visualizing}.
The auxiliary distribution $\mathbb{P}$ is a variant of assignment distribution $\mathbb{Q}$ with both instance-wise and cluster-wise normalization:
\begin{equation}
    p_{i k}=\frac{q^{2}_{i k} / f_{k}}{\sum^{K}_{j} q_{i j}^{2} / f_{j}}
\end{equation}
where $f_{k}=\sum^{N}_{i}q_{i k}$ are soft cluster frequencies.

\textbf{Analysis.}
The self-training deep clustering has been widely adopted in several existing works \cite{xie2016unsupervised,guo2017improved,guo2018deep,ghasedi2017deep,li2018discriminatively,rezaei2021learning,peng2017cascade}.
The success relies on the following properties:
First, the square of cluster assignment probability with cluster-wise normalization will guide the model put more attention (gradient) on the instances with higher confidence, which in turn reduce the impact of low confidence ones.
As a result, the cluster assignment vector tends to be one-hot.
Second, the soft cluster frequencies $f_{k}$ can be viewed as the sum of the probability that instance belongs to the $k$-th cluster.
This can prevent the degenerate solution that all instances belong to the same cluster.
The above two properties have a profound effect on the later works and many deep clustering methods can be viewed as variants of it.

\subsection{Mutual Information Maximization based Clustering}
\label{subsec:mi_clustering}
As introduced in subsection \ref{subsec:mi_rl}, mutual information maximization has achieved tremendous success in representation learning.
Benefiting from its unrestriction to the feature dimension and semantic meaning, it has also been introduced to the clustering module for measuring the dependence between instance and cluster assignment.
The mutual information maximization based clustering can be formulated as:
\begin{equation}
        \mathcal{I}(X, \tilde{Y}) =D_{K L}(p(x, \tilde{y}) \| p(x) p(\tilde{y})) =\sum_{x \in \mathcal{X}} \sum_{\tilde{y} \in \mathcal{Y}} p(x, \tilde{y}) \log \frac{p(x, \tilde{y})}{p(x) p(\tilde{y})}.
\end{equation}
where $X$ is the original data, $\tilde{Y}$ is the predicted labels, $p(x,\tilde{y})$ is the joint distribution and $p(x), p(\tilde{y})$ are the marginal distribution.
The estimation of mutual information is same as that in representation learning.

\textbf{Analysis.} The mutual information maximization based clustering and representation learning are quite similar despite the dimension of optimization object.
The major advantage is that it overcomes the gap between the representation learning and clustering.
As a consequence, the rapid development of deep representation learning techniques can be naturally introduced to the clustering task and optimized in a unified framework.

\subsection{Contrastive Deep Clustering}
Inspired by the success of contrastive representation learning and mutual information maximization based deep clustering, contrastive learning has also been introduced to deep clustering. 
Similar to contrastive representation learning, the target of contrastive deep clustering is to pull the positive pairs close while pushing the negative pairs far away.
The major difference lies in the definition of positive pairs and negative pairs, which can be further divided into three groups:

\subsubsection{Instance-Instance contrast}
The instance-instance contrast treats the cluster assignment of each instance as the representation and directly reuses the contrastive representation learning loss:
\begin{equation}
    \mathcal{L}_{IIC}=-\log \sum^{N}_{i=1}\frac{\exp \left(f\left(z_{i}, z^{\mathcal{T}}_{i}\right) / \tau\right)}{\sum_{j=1}^{N} \exp \left(f\left(z_{i}, z_{j}^{\mathcal{T}}\right) / \tau\right)}
\end{equation}
where $z_{i}$ is the cluster assignment of instance $x_{i}$ predicted by the clustering module.

\subsubsection{Cluster-Cluster contrast}
The cluster-cluster contrast treats each cluster as an instance in the embedding space, the target is pulling the cluster and its augmented version close while pushing different clusters far away, which can be formulated as:
\begin{equation}
    \mathcal{L}_{CCC}=-\log \sum^{K}_{k=1}\frac{\exp \left(f\left(c_{k}, c^{\mathcal{T}}_{k}\right) / \tau\right)}{\sum_{j=1}^{K} \exp \left(f\left(c_{k}, c_{j}^{\mathcal{T}}\right) / \tau\right)}
\end{equation}
where $c_{k}$ is the representation of cluster $k$.
It is worth noting that cluster-cluster contrast satisfy the basic requirement of clustering that each cluster should be dissimilar, which agrees with the clustering friendly representation learning described in subsection \ref{subsec:cluster_friend}.

\subsubsection{Instance-Cluster contrast}
The instance-cluster contrast is similar to the K-means which utilizes the cluster centroid as an explicit guidance.
Given the representation of each instance and cluster centroid in the same low-dimensional space, each instance is expected to be close to the corresponding cluster centroid while far from the other cluster centroids.
Such similarity and dissimilarity can be naturally modeled by the contrastive learning:
\begin{equation}
    \mathcal{L}_{ICC}=-\log \sum^{N}_{i=1}\frac{\exp \left(f\left(h_{i}, c'_{i}\right) / \tau\right)}{\sum_{j=1}^{K} \exp \left(f\left(h_{i}, c_{j}\right) / \tau\right)}
\end{equation}
where $c'_i$ is the corresponding cluster centroid of $x_i$ which is usually estimated by an alternative clustering method.
This can be also understood as maximizing the mutual information between the representation and cluster assignment with data augmentations.

\textbf{Analysis.}
Besides the advantages inherited from the mutual information maximization clustering, 
the primary advantages of contrastive deep clustering
is that the data augmentation helps improve the robustness of clustering, which has been ignored by most existing methods.
For detailed advantages of contrastive learning, please refer to the former contrastive learning surveys \cite{jing2020self,jaiswal2020survey}.

%% file: multistage.tex
Multistage deep clustering refers to the methods where two modules are separately optimized and sequentially connected. 
One straightforward way is to employ the deep unsupervised representation learning techniques to learn the representations (embedding) for each data instance first, 
and then feed the learned representations into the classic clustering models to obtain the final clustering results.
Such detachment of data processing and clustering facilitate researchers to perform clustering analysis.
More specifically, all existing clustering algorithms can be of service to any research scenarios.


Early multi-stage deep clustering methods \cite{tian2014learning,huang2014deep} have trained a deep auto-encoderto learn the representations, which can be directly packed as input of the K-means method to obtain the clustering results.
Later, 
deep subspace clustering was proposed to learn an affinity matrix and instance representations first, and then performs clustering by spectral clustering on the affinity matrix \cite{ji2017deep,zhang2021learning,zhou2018deep} or the K-means on the instance representations \cite{peng2016deep}.
Thanks to the contribution of scikit-learn \cite{scikit-learn} and many other open-source machine learning libraries, clustering algorithms has been applied to many fields with a limited cost of programming. For example, in the scenario of textual/video/graph data clustering, relation (similarity) matching was used in \cite{chang2005using, tseng2010generic,kong2020learning}, K-Means in \cite{cerquitelli2017data}, Spectral Clustering in \cite{janani2019text,peng2021deep,alami2021unsupervised} and Hierarchical Agglomerative Clustering in \cite{shen2018improving}, so as many other clustering algorithms being directly applied. In special, graph cut based node clustering like Metis \cite{karypis1998fast}, Graclus \cite{dhillon2007weighted} and Balance Normalized Cut (BNC) \cite{chiang2012scalable} were used in graph clustering applications \cite{chiang2019cluster,nasrazadani2022sign,wu2016cccf}.

The most recent multistage methods have explicitly incorporated the clustering prior into the representation learning, then conduct clustering on the target friendly representations.
For example, 
IDFD \cite{tao2021clustering} learn representations with two aims: learning similarities among instances and reducing correlations within features. 
With the above explicit purposes, a naive K-means on the learned representations can also achieve competitive clustering results over many existing deep clustering methods.


\textbf{Summary.}
Multistage methods enjoy the property of fast deployment, programming friendly, and intuitive understanding.
However, such a simple combination of deep representation learning and shallow (traditional) clustering has the following weakness:
    1) Most representation learning methods are not intently designed for clustering tasks, which can not provide sufficient discriminative power for clustering. Specifically, clustering reflects global patterns among data instances, while existing representation learning methods have largely focused on the individual patterns.
    2) The clustering results can not be further utilized to guide the representation learning, which is of great importance for comprehensive representation. In particular, the cluster structure implies the inherent relationships among data instances, and can in turn serve as critical guidance for improving representation learning.
To conclude, such straightforward cascade connection will cut off the information flow and interactions between representation learning and clustering, thus the limitations of both side will influence the final performance together.

%% file: iterative.tex
The key motivation of iterative deep clustering is that good representations can benefit clustering while clustering results reversely provide supervisory to representation learning. 
Briefly speaking, most existing iterative deep clustering pipeline is iteratively updated between two steps: 1) calculating clustering results given current representations and 2) updating the representations given the current clustering results.
Since the representation module only provides input for the clustering module in iterative deep clustering, in this subsection, we classify existing iterative deep clustering methods according to the information provided by the clustering module.

\subsubsection{Iterative deep clustering with individual supervision}
The individual supervision in iterative deep clustering depends on the pseudo labels generated by the clustering module, which can be utilized to train the representation learning module in a supervised manner.

In early works \cite{yang2017towards,song2013auto}, the cluster centroids and assignments are updated in a K-means way. 
$\text{S}^2\text{ConvSCN}$ \cite{zhang2019self} and PSSC \cite{lv2021pseudo} combine subspace clustering and pseudo labeling, which obtain pseudo labels by spectral clustering or partitioning the pseudo similarity graph. 
Later, a lot of works \cite{caron2018deep,van2020scan,niu2021spice} tend to utilize neural networks for both representation learning and clustering, where these two parts are combined together as one neural network. The clustering module is usually a multilayer perceptron(MLP) that produces soft clustering assignments. In this way, the hard pseudo labels can guide both clustering and representation learning through gradient backpropagation with proper constraints.
The representative method is DeepCluster \cite{caron2018deep},
which alternates between K-means clustering 
and updating the backbone along with the classifier by minimizing the gap between predicted clustering assignments and pseudo labels. 
In fact,  DeepCluster has already been applied as a mature clustering algorithm in video clustering \cite{alwassel2020self}.

Recently, SCAN \cite{van2020scan} follows a pretraining-with-finetuning framework. The clustering results are fine-tuned with self-labeling, which selects the highly confident instances by thresholding the soft assignment probability, and updates the whole network by minimizing the cross-entropy loss on the selected instances.
SPICE \cite{niu2021spice} is another representative iterative deep clustering method, where the classification model is first trained with the guidance of pseudo labels and then retrained by the semi-supervised training on the set of reliably labeled instances.


\subsubsection{Iterative deep clustering with relational supervision}
The relational supervision in iterative deep clustering refers to the relationship based on the pesudo labels, which provides pairwise guidance to the representation learning module.
More specifically, the relationship is usually modeled by whether two instances have same discrete pseudo labels \cite{yang2016joint,chang2017deep} and the model is trained as a binary classification task.
Another popular branch of methods \cite{zhang2019neural, niu2020gatcluster} model the relationship by the similarity of cluster assignment probabilities, which train the representation learning as a regression task.

\textbf{Summary.}
Iterative deep clustering methods benefit from the mutual promotion between representation learning and clustering. 
However, they also suffer from the error propagation in the iterative process. More specifically, the inaccurate clustering results can lead to chaotic representations where the performance are limited by the self-labeling effectiveness.
Furthermore, this will in turn affect the clustering results especially in the early stage of training.
Therefore, existing iterative clustering methods heavily rely on the pretraining of representation module.

%% file: generative.tex
Generative models are able to capture, represent and recreate data points, and thus are drawing increasing attention from both academia and industry. They would make the hypotheses about the latent cluster structure and then infer the clustering assignment by estimation of data density. The most representative model is Gaussian Mixture Model (GMM) \cite{reynolds2009gaussian}, which assumes that the data points are generated from a Mixture-of-Gaussians. Specifically, suppose there are $K$ clusters, and an observed sample $x$ is generated from the following process:
\begin{enumerate}
    \item Choose a cluster: $c \sim \operatorname{Mult}(\pi)$
 \item Draw a sample: $\mathbf{x}|c \sim \mathcal{N}\left(\mathbf{\mu}_{c}, \mathbf{\sigma}_{c}^{2} \mathbf{I}\right)$
\end{enumerate}
where $\pi$ denotes the prior probability for clustering; $\operatorname{Mult}(\pi)$ is the multinomial distribution with the parameter $\pi$; $\mathbf{\mu}_{c}$ and $\mathbf{\sigma}_{c}$ are the mean and variance parameters of the Gaussian distribution corresponding to the cluster $c$. The well-known expectation maximization algorithm can be employed to learn the optimal parameters and clustering assignment. 

While GMM has gained successful applications, its shallow structure is usually insufficient for capturing the nonlinear patterns of the data, adversely affecting its performance on complex data (\eg images, texts, graphs and etc). To address this problem, deep generative models have been proposed to combine the generative model with the powerful deep neural networks, which have enough capacity to model the non-linear and complex data. This kind of methods can be classified into two types: the methods based on Variational Auto-Encoder (VAE) and the methods based on Generative Adversarial Networks (GAN).

\subsubsection{Deep Generative clustering based on Variational Auto-Encoder}

For clustering of high dimensional and complex data, one promising solution is to directly stack GMM with a deep neural network --- GMM generates a latent vector $z$, and the deep neural network further transforms the latent vector $z$ into the complex data instance $x$. In this way, the stacked model can enjoy the merit of the latent cluster structure and meanwhile has sufficient capacity to model the complex data. For example, the representative models, VaDE \cite{jiang2016variational} and GMVAE \cite{dilokthanakul2016deep}, assume the following generative process for each instance:
\begin{enumerate}
    \item Choose a cluster: $c \sim \operatorname{Mult}(\pi)$
 \item Draw a latent vector: $\mathbf{z}|c \sim \mathcal{N}\left(\mathbf{\mu_z}(c;\beta), \mathbf{\sigma}_z^{2}((c;\beta)) \mathbf{I}\right)$
 \item Draw a sample: $\bf{x}|\bf{z} \sim \mathcal{N}\left(\mathbf{\mu_x}(\mathbf{z};\theta), \mathbf{\sigma}_x^{2}(\mathbf{z};\theta) \mathbf{I}\right)$
\end{enumerate}
where $\mathbf{\mu_z}(.;\beta)$, $\bf{\sigma_z^2}(.;\beta)$, $\mathbf{\mu_x}(.;\theta)$ and $\mathbf{\sigma_x^2}(.;\theta)$ are given by neural networks with parameters $\beta$ and $\theta$, which determinize the mean and variance of Gaussian distributions, respectively. Given the above generative process, the optimal parameters and cluster assignment can be obtained by maximizing the likelihood of the given data points as:
\begin{align}
    \log p({\bf{x}}) = \log \int_{\bf{z}} {\sum\limits_c {p ({\bf{x}}|{\bf{z}})p(\mathbf{z}|c)p(c)\bf{d}{\bf{z}}}}
    \end{align}

However, directly optimizing the above likelihood is intractable as it involves integration and complex neural network. Variational Auto-Encoder (VAE)\cite{kingma2013auto} sheds a light to tackle this problem such that the parameters and the posterior can be efficiently estimated via backpropagation. Specifically, the generative model is trained with the following variational inference objective, \aka the evidence lower bound (ELBO):
\begin{align}
{{\mathcal L}_{{\mathop{\rm ELBO}\nolimits} }}({\bf{x}}) = {\mathbb E_{q({\bf{z}},c| {\bf{x}})}}\left[ {\log \frac{{p({\bf{x}},{\bf{z}},c)}}{{q({\bf{z}},c|{\bf{x}})}}} \right]
\end{align}
where $q(\bf{z}, c|\bf{x};\varphi)$ is the variational posterior to approximate the true posterior, which can be modelled with the recognition networks $\varphi$. Monte Carlo \cite{kalos2009monte} and the reparameterization trick \cite{kingma2013auto} can be employed to learn the parameters.  

More recently, upon VaDE and GMVAE, some improved variants have been proposed. For example, Prasad \etal~ \cite{prasad2020variational} introduced a data augmentation technique that constrains an input instance (\eg image) to share a similar clustering distribution with its augmented one; Li \etal~ \cite{li2020unsupervised} employed Monte Carlo objective and the Std Annealing track for optimizing mixture models, which would generate better-separated embeddings than the basic VAE-based methods; Ji \etal~ \cite{ji2021decoder} proposed to replace decoder in VAE with an improved mutual-information-based objective; Wang \etal~ \cite{wang2022neural} proposed to separate the latent embeddings into two parts which capture the particularity and commonality of the clusters, respectively.

\subsubsection{Deep Generative clustering based on Generative Adversarial Network}

Recent years have witnessed the great success of Generative Adversarial Network (GAN) in estimating complex data distribution \cite{goodfellow2014generative,jabbar2021survey,arjovsky2017wasserstein}. A standard GAN contains two components: a generator $G$ that targets synthesizing “real” samples to fool the discriminator, and a discriminator $D$ tries to discriminate the real data from the generated samples. With the adversary between the two components, the generator could generate samples that have a similar distribution to the data. Inspired by such outstanding ability, it would be promising to integrate GAN into generative clustering models. Specifically, Ben-Yosef \etal~ \cite{ben2018gaussian} proposed to stack a GMM with a GAN, where GMM serves as a prior distribution for generating data instances. Formally, they optimized the following objective function:
\begin{align}
\min _{G} \max _{D} V(D, G)=\underset{\mathbf{x} \sim p_{\mathcal{X}}(\mathbf{x})}{\mathbb{E}}[\log D(\mathbf{x})]+\underset{\mathbf{z} \sim p_{\mathcal{Z}}(\mathbf{z})}{\mathbb{E}}[\log (1-D(G(\mathbf{z})))]
\end{align}
where $p_{\mathcal{X}}(\mathbf{x})$ denotes the training data distribution; $p_{\mathcal{Z}}(\mathbf{z})$ is a prior distribution of $G$ and defined as mixture of Gaussians:
\begin{align}
    {p_{\mathcal Z}}({\bf{z}}) = \sum\limits_{k = 1}^K {{\pi _k}} * {\mathcal N}\left( {{\bf{\mu }}_c,{{\bf{\sigma }}^2_c}{\bf{I}}} \right)
\end{align}
By equipping GAN with such multi-modal probability distribution, the model could provide a better fit to the complex data distribution especially when the data includes many different clusters. 

There are also some improved variants. For example, Yu \etal~\cite{yu2018mixture} proposed to directly replace the Gaussian distribution of GMM with a GAN and developed a $\epsilon$-expectation- maximization learning algorithm to forbid early convergence issues; Ntelemis \etal~ \cite{ntelemis2021image} proposed to employ Sobel operations prior to the discriminator of the GAN; Mukherjee \etal~ \cite{mukherjee2019clustergan} proposed to sample the latent vector $z$ from a mixture of one-hot encoded variables and continuous latent variables. An inverse network with a clustering-specific loss is introduced to make the model more friendly to the clustering task. Analogously, an inverse network is introduced in \cite{ghasedi2019balanced,gan2021learning} for the feature-level (\ie latent vector) adversary. 

\textbf{Summary.} Although deep generative clustering models can generate samples while completing clustering, they also have some weaknesses: 1) Training a generative model usually involves Monte Carlo sampling, which may incur training unstable and high computational complexity; 2) The mainstream generative models are based on VAE and GAN, and inevitably inherit the same disadvantages of them. VAE-based models usually require prior assumptions on the data distributions, which may not be held in real cases; although GAN-based algorithms are more flexible and diverse, they usually suffer from mode collapse and slow convergence, especially for the data with multiply clusters.


%% file: simultaneous.tex
Simultaneously deep clustering is the most active direction in current deep clustering, where representation learning module and the clustering module are simultaneously optimized in an end-to-end manner.
Although most iterative deep clustering methods also optimize both two modules with a single objective, the two modules are optimized in an explicit iterative manner and can not be updated simultaneously.   
In this subsection, we introduce the representative architectures of simultaneously deep clustering.





\subsubsection{Auto-encoder with self-training}
Auto-encoder is a powerful tool to learn data representations in an unsupervised way, and has been utilized since the first attempts of simultaneously deep clustering \cite{huang2014deep}.

The representative method is DEC \cite{xie2016unsupervised} which combines the auto-encoder with the self-training strategy.
Such simple but effective strategy has deeply influenced the follow-up works.
The auto-encoder is pre-trained and only the encoder is utilized as the initialization of the representation learning module.
The self-training strategy mentioned in subsection \ref{subsec:self-training} is then introduced to optimize the clustering and representation learning simultaneously.

Based on the vanilla DEC method, many variants and improvements are proposed. 
To preserve the local structure of each instance,
IDEC \cite{guo2017improved} further integrates the reconstruction loss to the auto-encoder.
The general formulation of auto-encoder with self-training can be summarized as:
\begin{equation}
    \mathcal{L}_{AEST}=\mathcal{L}_{AE}+\mathcal{L}_{ST}
\end{equation}
where $\mathcal{L}_{AE}$ is the loss of auto-encoder and $\mathcal{L}_{ST}$ is the loss of clustering oriented self-training, e.g. the neighborhood constraint in DEC.
To improve the capability of auto-encoder, some efforts are made to adapt different data types.
In \cite{ghasedi2017deep,li2018discriminatively,guo2018deep}, the linear layer of auto-encoder is replaced with fully convolutional layers so that the image feature can be well captured.
In CCNN \cite{hsu2017cnn}, the clustering convolutional neural network is proposed as a new backbone to extract the representations which are friendly to the clustering task.
In DEPICT \cite{ghasedi2017deep}, an additional noisy encoder is introduced, and the robustness of auto-encoder is improved by minimizing the reconstruction error of every layer between the noisy decoder and the clean encoder.


Although the self-training strategy has achieved success, later works also make attempts for specific problem. 
To boost the robustness of clustering, self-training is applied between two branches: clean and augmented \cite{guo2018deep} (noisy \cite{ghasedi2017deep}). Specifically, the target distribution $\mathbb{P}$ is computed via the clean branch to guide the soft assignments $\mathbb{Q}$ of the augmented or noisy branch.
Self-training strategy can be combined with subspace clustering. CSC \cite{peng2017cascade} introduces the assumption named invariance of distribution, i.e. the target distribution $\mathbb{P}$ should be invariant to different distance metrics in subspace space. Therefore, two metrics (Euclidean and Cosine distance) are used to compute the target distributions $\mathbb{P}_E$ and $\mathbb{P}_C$, with the KL divergence between them minimized.

According to aforementioned analysis, self-training is similar to the K-means clustering and suffers from the unbalance problem \cite{buda2018systematic} between different clusters.
To solve the problem of the unbalanced data and out-of-distribution samples, StatDEC \cite{rezaei2021learning} improves target distribution by adding normalized instance frequency of clusters.
In this way, the model can preserve discrimination of small groups and form a local clustering boundary which is insensitive to unbalanced clusters.


\subsubsection{Mutual Information Maximization based Clustering}
\label{subsec:mi_model}
As illustrated in subsection \ref{subsec:mi_rl} and \ref{subsec:mi_clustering}, the mutual information maximization has been successfully applied in both representation and clustering.
The unified form of mutual information in both modules has brought convenience for understanding and implementation.

The representative mutual information maximization based clustering method is DCCM \cite{wu2019deep}.
For each data instance, the mutual information between deep and shallow layer representation is maximized so that the consistency of representation can be guaranteed.
Such consistency is further extended to the cluster assignment space by encouraging the instances with the same pseudo labels to share similar representations.
Many later works can be regarded as variants of this method.
In VCAMI \cite{ji2021variational} and IIC \cite{ji2019invariant}, the augmented mutual information (AMI) is introduced to improve the robustness.
Such augmentation invariant has inspired the later contrastive deep clustering methods, which will be introduced in the next subsection.
In ImC-SWAV \cite{ntelemis2021information}, the mutual information between the integrated discrete representation and a discrete probability distribution is maximized, which improves the vanilla SWAV \cite{caron2020unsupervised} method.


\subsubsection{Contrastive Deep Clustering}
Similar to mutual information maximization based deep clustering, contrastive learning has also been successfully applied in both representation learning module and clustering module.
The main idea of contrastive learning is pulling similar instances closer while pushing different instances away, which is in the spirit of clustering that instances from the same cluster should be close while instances from different clusters should be apart.

The representative contrastive deep clustering method is CC \cite{li2021contrastive}.
The basic idea is to treat each cluster as the data instance in the low-dimensional space.
The instance discrimination task in contrastive representation learning can be migrated to the clustering task by discriminating different clusters, which is the fundamental requirement of clustering.
Furthermore, the advantage of augmentation invariant and local robustness can be preserved in the clustering task.

Take CC as the fundamental architecture, many contrastive deep clustering can be viewed as variants of it.
PICA \cite{huang2020deep} can be viewed as the degeneration of CC without augmentation, it directly separates different clusters by minimizing the cosine similarity between the cluster-wise assignment statistic vectors.
In DCCM \cite{wu2019deep}, the augmentation is introduced to guarantee the local robustness of the learned representation. 
DRC \cite{zhong2020deep} has the same contrastive learning as CC where the cluster representation is called assignment probability.
The difference lies in the cluster regularization which is inspired by group lasso \cite{meier2008group}.
In CRLC \cite{do2021clustering}, contrastive learning is performed between the cluster assignments of two augmented versions of the same instance, rather than the cluster representations. 
Also, the dot product in contrastive learning is replaced by the log-dot product, which is more suitable for the probabilistic contrastive learning.
SCCL \cite{zhang2021supporting} extends this approach with textual data augmentation, which proves that this contrastive learning based self-training framework is universally applicable.


The later works further adopt the contrastive clustering in the semantic space.
In SCL \cite{huang2021deep}, the negative samples are limited by different pseudo labels,
so that instances from different clusters can be further distinguished.
MiCE \cite{tsai2020mice} follows a divide-and-conquer strategy where the gating function divides the whole dataset into clusters and the experts in each cluster aim at discriminating the instances in the cluster.
However, compared with InfoNCE that models alignment and uniformity implicitly\cite{wang2020understanding}, MiCE model these two properties in a more explicit way.
Recently, TCC \cite{shen2021you} further improve the efficiency with the reparametrization trick and explicitly improve the cluster discriminability.

Some other methods make attempts on overcoming the problems of vanilla contrastive learning. 
In GCC \cite{zhong2021graph}, the positive pair and negative pair are selected by the KNN graph constructed on the instance representation. 
This may be related to overcoming the `false negative' problem in contrastive learning. 
In NCC \cite{huang2021exploring}, the contrastive learning module is replaced from SimCLR to BYOL \cite{grill2020bootstrap}, so that the over-reliance on the negative samples can be solved.

\subsubsection{Hybrid simultaneous deep clustering}
The aforementioned simultaneous deep clustering methods have remarkable characteristics and advantages, some other works are hybrids of the above techniques.
SCCL \cite{zhang2021supporting} and Sundareswaran \etal~ \cite{sundareswaran2021cluster} combine contrastive representation learning and self-training.
DDC \cite{chang2019deep} and RCC \cite{shah2017robust} combine relation matching clustering and pseudo labeling to boost the clustering performance. DCC \cite{shah2018deep} combines the auto-encoder based representation and the relation matching clustering.
The auto-encoder based representation learning and spectral clustering are combined in \cite{yang2019deep} by incorporating the augmentations from the contrastive learning.

\textbf{Summary.} The simultaneous deep clustering has attracted the most attention for its unified optimization.
Intuitively, the learned representation is clustering oriented and the clustering is conducted on the discriminative space.
However, it may incur from undesired prejudice of optimization focus between representation learning module and clustering module, which can only be mitigated by the manually setting of the balanced parameter for now.
Also, the model is easy to sink into degenerate solutions where all instances are assigned into one single cluster.

%% file: application.tex
\begin{sidewaystable} 
\centering
\caption{The proposed taxonomy of existing methods and their properities}
\label{Label}       
\scalebox{0.95}{
\begin{tabularx}{\textwidth}{|p{0.095\textwidth}<{\centering}|p{0.20\textwidth}<{\centering}|p{0.245\textwidth}<{\centering}|p{0.16\textwidth}<{\centering}|p{0.05\textwidth}<{\centering}|X<{\centering}|}
 \hline 
 Taxonomy&Advantages & Disadvantages&Clustering Module&Type&Citation  \\ \hline
    \multirow{7}{*}{Multi Stage} & \multirowcell{7}[0pt][l]{ 1. Easy to implement  \\
2. Better interpretability} &\multirowcell{7}[0pt][l]{1. Sub-optimal representations \\
2. Limit on clustering performance} &   \multirow{7}{*}{Multi Stage}  &  \multirow{2}{*}{Image} &\cite{tian2014learning}\cite{ji2017deep}\cite{zhang2021learning}\cite{zhou2018deep}\newline \cite{peng2016deep}\cite{tao2021clustering}\cite{dang2021nearest}    \\ \cline{5-6} 
                          &     &  &                                    &  Graph  &  \cite{huang2014deep}\cite{chiang2019cluster}\cite{nasrazadani2022sign}\cite{wu2016cccf}  \\ \cline{5-6} 
                          &     &  &                                   &  \multirow{2}{*}{Textual}  &  \cite{chang2005using}\cite{tseng2010generic}\cite{cerquitelli2017data}\cite{shen2018improving}\newline \cite{janani2019text} \cite{li2020text}\cite{kong2020learning}\cite{alami2021unsupervised}   \\ \cline{5-6} 
                          &     &  &                                   &  Video  &  \cite{peng2021deep}\cite{wang2020cluster}   \\ \cline{5-6} 
                          &     &  &                                   &  Medical  &  \cite{hassan2021novel}\cite{dao2010inferring}   \\ \hline
    \multirow{5}{*}{Generative}& \multirowcell{5}[0pt][l]{ 1. Latent cluster structure \\ discovery \\ 2. Able to model complex \\ data distribution} & \multirowcell{5}[0pt][l]{1. Unstable training \\ 2. High computational complexity \\ 3. Sensitive to prior distribution \\ assumptions}& \multirow{2}{*}{VAE} &\multirow{2}{*}{Image}&\cite{dilokthanakul2016deep}\cite{yang2019deep}\cite{jiang2016variational}\cite{prasad2020variational}\newline \cite{ji2021decoder}  \cite{li2020unsupervised}\cite{wang2022neural}   \\ \cline{4-6}
     &  &  & \multirow{3}{*}{GAN} &\multirow{2}{*}{Image}&\cite{mukherjee2019clustergan}\cite{ben2018gaussian}\cite{yu2018mixture}\cite{ntelemis2021image}\newline \cite{gan2021learning} \cite{ghasedi2019balanced}  \\ \cline{5-6}
      &  &  & & Graph & \cite{jia2019communitygan}   \\ \hline
    \multirow{5}{*}{Iterative}  & \multirowcell{5}[0pt][l]{ 1. Clustering oriented \\ representation learning \\ 2. Discriminative space for \\ clusteing} & \multirowcell{5}[0pt][l]{ 1. Error propagation \\ 2. Limit on pseudo-label quality} & \multirow{4}{*}{Individual supervision} & \multirow{2}{*}{Image} & \cite{yang2016joint}\cite{yang2017towards}\cite{song2013auto}\cite{zhang2019self}\newline  \cite{lv2021pseudo}\cite{van2020scan}\cite{niu2021spice}  \\ \cline{5-6}
                          &  &   &                                &Video&\cite{alwassel2020self} \\ \cline{5-6}
                          &  &   &                                &Medical&\cite{kart2021deepmcat}\cite{mittal2021new}   \\ \cline{4-6}
                          &  & &\multirow{1}{*}{Relational supervision}&Image&\cite{chang2017deep}\cite{caron2018deep}\cite{zhang2019neural}\cite{niu2020gatcluster}   \\  \hline
    \multirow{12}{*}{Simultaneous} & \multirowcell{12}[0pt][l]{ 1. Clustering oriented \\ representation learning  \\ 2.Discriminative space for \\ clusteing\\ 3. Unified optimization \\ framework} & \multirowcell{12}[0pt][l]{1. Undesired prejudice of \\ optimization  focus \\ 2. Sensitive to degenerate solutions    }  &\multirow{4}{*}{Auto-encoder based}&\multirow{2}{*}{Image}& \cite{xie2016unsupervised}\cite{guo2017improved}\cite{ghasedi2017deep}\cite{guo2018deep}\cite{hsu2017cnn}\newline
    \cite{li2018discriminatively}\cite{peng2017cascade}\cite{rezaei2021learning} \\ \cline{5-6}
    &  & & &Graph&\cite{bo2020structural}   \\ \cline{5-6}
    &  & & &Medical&\cite{tian2019clustering}\cite{hu2020iterative}   \\ \cline{4-6}
    &  & & Mutual Information based clustering &\multirow{2}{*}{Image}&\cite{wu2019deep}\cite{ji2021variational}\cite{ji2019invariant}\cite{ntelemis2021information}   \\ \cline{4-6}
    &  & &\multirow{5}{*}{Contrastive}&\multirow{3}{*}{Image}&\cite{li2021contrastive}\cite{dang2021doubly}\cite{huang2020deep}\cite{wu2019deep}\newline \cite{zhong2020deep}  \cite{do2021clustering}\cite{huang2021deep}\cite{tsai2020mice}\newline \cite{shen2021you}\cite{huang2021exploring}  \cite{dang2021nearest}   \\ \cline{5-6}
    &  & & &Textual&\cite{zhang2021supporting}   \\ \cline{5-6}
    &  && &Graph&\cite{zhong2021graph}    \\ \cline{4-6}
    &  & &\multirow{2}{*}{Hybrid methods}& Image &\cite{sundareswaran2021cluster}\cite{chang2019deep}\cite{shah2017robust}   \\ \cline{5-6}
    &  &  & &Textual &\cite{zhang2021supporting}\cite{meng2020hierarchical} \cite{thirumoorthy2021hybrid}  \\ \hline
    \end{tabularx}%
    }
\end{sidewaystable}

%% file: dataset-evaluation.tex
In this section, we introduce benchmark datasets and evaluation metrics that are widely used in existing deep clustering methods.

\subsection{Datasets}

\subsubsection{Image Datasets}

Image is the most commonly used data type in real-world deep clustering. The early attempts of deep clustering are applied to image datasets including COIL-20\footnote{http://www.cs.columbia.edu/CAVE/software/softlib/coil-20.php}, CMU PIE\footnote{http://www.cs.cmu.edu/afs/cs/project/PIE/MultiPie/Multi-Pie/Home.html}, Yale-B\footnote{http://vision.ucsd.edu/\textasciitilde leekc/ExtYaleDatabase/Yale\%20Face\%20Database.htm}, MNIST\footnote{http://yann.lecun.com/exdb/mnist/index.html}, CIFAR\footnote{http://www.cs.toronto.edu/\textasciitilde kriz/cifar.html} and  STL-10\footnote{https://cs.stanford.edu/\textasciitilde acoates/stl10/}. 
Recently, efforts has been paid to perform clustering on large volume vision datasets (e.g. ImageNet). Although existing methods have achieved promising performance on ImageNet-10 and ImageNet-Dogs dataset, clustering on Tiny-ImageNet (200 clusters) or full-size ImageNet is still challenging.

\subsubsection{Textual Datasets}
The widely used textual datasets in early applications of texutal data clustering include
Reuters-21578\footnote{https://archive.ics.uci.edu/ml/datasets/Reuters-21578+Text+Categorization+Collection} and 20 Newsgroups\footnote{http://qwone.com/\textasciitilde jason/20Newsgroups/ } datasets, which have already been vectorized and little feature engineering is needed. Currently, raw textual datasets including IMDB\footnote{http://ai.stanford.edu/\textasciitilde amaas/data/sentiment/}, stackOverflow\footnote{https://github.com/jacoxu/StackOverflow}, and more in nlp-datasets github repository\footnote{https://github.com/niderhoff/nlp-datasets} is still challenging for deep textual clustering. 

\subsubsection{Video Datasets}
The ultimate task of video clustering varies from action classification \cite{peng2021deep} to video anomaly detection \cite{wang2020cluster}. Kinetics-400 and Kinetics-600\footnote{https://www.deepmind.com/open-source/kinetics} are two of the most famous video datasets.
The others include UCF-101 dataset\footnote{https://www.crcv.ucf.edu/research/data-sets/ucf101/} and HMDB-51 dataset\footnote{https://serre-lab.clps.brown.edu/resource/hmdb-a-large-human-motion-database/\#dataset}.

\subsubsection{Graph Datasets}
Commonly used graph datasets for node clustering can be referred from the following papers \cite{shun2016parallel,chiang2019cluster,nasrazadani2022sign} and Stanford Network Analysis Project\footnote{http://snap.stanford.edu/}. And there are also graph-level classification datasets like PROTINS\cite{borgwardt2005protein} and MUTAG\cite{debnath1991structure}, which can be used to perform and evaluate graph-level clustering.

\subsection{Evaluation Metrics}
Evaluation metrics aim to evaluate the validity of methods. In the field of Deep Clustering, three standard clustering performance metrics are widely used: Accuracy(ACC), Normalized Mutual Information(NMI) and Adjusted Rand Index(ARI).

\subsubsection{Accuracy}
ACC indicates the average correct classification rate of clustering samples. Given the ground truth labels $Y=\left\{ y_i|1\leq i\leq N\right\}$ and the predicted hard assignments $\tilde{Y}=\left\{ \tilde{y}_i|1\leq i\leq N \right\}$, ACC can be computed as follows:
\begin{equation}
    ACC(\tilde{Y}, Y)=\mathop{\max}_{g} \frac{1}{N} \sum_{i=1}^{N} \mathbf{1}\{ y_i = g(\tilde{y}_i)\}
\end{equation}
where $g$ is the set of all possible one-to-one mappings between the predicted labels and ground truth labels. The optimal mapping can be efficiently obtained by the Hungarian algorithm \cite{kuhn1955hungarian}.

\subsubsection{Normalized Mutual Information}
NMI quantifies the mutual information between the predicted labels and ground truth labels into $[0,1]$:
\begin{equation}
    NMI(\tilde{Y}, Y)=\frac{\mathcal{I}(\tilde{Y}; Y)}{\frac{1}{2} \left[H(\tilde{Y})+H(Y)\right]}
\end{equation}
where $H(Y)$ is the entropy of $Y$ and $\mathcal{I}(\tilde{Y}; Y)$ is the mutual information between $\tilde{Y}$ and $Y$.

\subsubsection{Adjusted Rand Index}
ARI comes from Rand Index(RI), which regards the clustering result as a series of pair-wise decisions and measures it according to the rate of correct decisions:
\begin{equation}
    RI=\frac{TP + TN}{C_N^2}
\end{equation}
where $TP$ and $TN$ denote the number of true positive pairs and true negative pairs, $C_N^2$ is the number of possible sample pairs. However, the RI value of two random partitions is not a constant approaching 0, thus ARI was introduced:
\begin{equation}
    ARI=\frac{RI-\mathbb{E}(RI)}{\max(RI)-\mathbb{E}(RI)}
\end{equation}

Both ACC and NMI $\in [0, 1]$ while ARI $\in [-1, 1]$, in which higher values indicate better performance.

%% file: real-application.tex
Despite the success of deep clustering in mining global pattern among instances, it has also benefit various downstream tasks.
In this section, we discuss some typical applications of deep clustering.

\subsection{Community Detection}
Community Detection \cite{fortunato2010community,liu2020deep,jin2021survey} aims at partitioning the graph network into several sub-graphs mainly according to connection density, which can be treated as node-level graph clustering task. 
Early works are mainly based on modularity measurement \cite{girvan2002community,blondel2008fast}, Maximum flows \cite{flake2002self}, graph cut \cite{shi2000normalized} and its extension, spectral methods \cite{von2007tutorial}.
With the development of Graph Nerual Networks (GNNs)\cite{ma2019disentangled,zhou2019hahe}, nodes are represented as individual instances in the low-dimensional space.
As a result, the border between modern community detection \cite{su2022comprehensive} and graph clustering\cite{li2017semi,li2020schain,zhou2020cross} is gradually getting blurry, and GNN based graph clustering \cite{tsitsulin2020graph,bianchi2020spectral} have already been applied to many applications.
However, different from early community detection that focus on the network topology, the graph clustering usually incorporates the node attributes and other side information.
How to release the power of GNNs while reserve the topology characteristic is still under study.




\subsection{Anomaly Detection}
\label{app:anomaly}
Anomaly Detection (\aka Outlier Detection, Novelty Detection) is a technique for identifying abnormal instances or patterns among data. 
Early before deep clustering, density based clustering methods \cite{ester1996density,schubert2017dbscan,comaniciu2002mean} have specifically mentioned and addressed the problem of noise during clustering, which has enlighted a group of density based anomaly detection methods \cite{ccelik2011anomaly,chen2011anomaly}. 
The later anomaly detection methods \cite{markovitz2020graph,wang2020cluster,ma2021comprehensive} have utilized the clustering results and identify the anomaly instances as those far from the cluster centroids or border of each cluster.
Currently, the deep clustering has shown great potential in forming a better clustering space for anomaly detection.
Instead of performing anomaly detection after the deep clustering, the recent efforts have been put on conducting them in a unified framework: identify and remove the anomaly instances to reduce the impact on clustering \cite{liu2019clustering}, and anomaly detection can be further improved with better clustering results.

\subsection{Segmentation and Object Detection}
Image Segmentation is one of the most important approaches to simulating human understanding of images which aims at dividing the pixels into disjoint regions.
Generally speaking, image segmentation is doing pixel classification in a supervised manner and pixel clustering in an unsupervised manner \cite{coleman1979image,arifin2006image}. 
Currently, the deep clustering has been successfully applied in segmentation by 
utilizing the clustered regions to generate Scene Graph \cite{yang2018graph}.
Yi \etal~ \cite{yi2012image} surveyed graph cut based image segmentation, where graph cut is one of the most fundamental solutions to perform clustering (Spectral Clustering). 
3D clustering can be a solution to 3D Object Detection, like in \cite{chen2022self}, 3D points were clustered to represent an object with geometric consistency. 
But such clustering based segmentation and object detection has no guarantee for small region and object, where the expected clustering result is highly unbalanced. And the global positional information of a pixel may be ignored when peoforming clustering.

\subsection{Medical Applications}
\label{medical}
The convolutional neural network has successfully promoted the development of medical image processing in a supervised manner.
However, the manual dataset labeling process is often labor-intensive and requiring expert medical knowledge \cite{kart2021deepmcat}, which is hard to realize in the real-world scenarios. 
Recently, the deep clustering has been introduced to automatically categorize large-scale medical images \cite{kart2021deepmcat}.
Mit al \etal~ \cite{mittal2021new} introduce medical image clustering analysis for faster COVID-19 diagnostic.
In the biological science field, single cell RNA sequencing (scRNA-seq) \cite{eraslan2019single} gives an output cell-gene matrix for cell population and behaviour analysis and even new cell discovery. For this purpose, ScDeepCluster \cite{tian2019clustering} and ItClust \cite{hu2020iterative} develop their models based on DEC \cite{xie2016unsupervised} to cluster scRNA-seq data, and MARS \cite{brbic2020mars} combining transfer learning and clustering to discover novel cell types. More application can be found in gene data clustering area\cite{gao2006discovering}.

\subsection{Summary}
In addition to the above successful deep clustering applications, the clustering also has great potential in many other domains, such as
financial analysis \cite{jaiswal2020green,govindasamy2018cluster,close2020combining}, 
trajectory analysis \cite{bandaragoda2019trajectory,olive2019trajectory}, 
and social media understanding \cite{yue2019survey,orkphol2019sentiment,zhu2014tripartite}.
Although most of the aforementioned methods has not incorporate deep learning techniques, with the increasing data volume and dimension, we believe that deep clustering will pave the way for these scenarios.

%% file: future.tex
In this section, we conduct some future directions of deep clustering based on the above cornerstone, taxonomy and real-world applications.

\subsection{Initialization of Deep Clustering Module}
The initialization of deep neural networks usually play an important role in training efficiency and stability \cite{glorot2010understanding}.
This is more critical in deep clustering where both the representation learning module and clustering module are modeled by deep neural networks.
Recently, the model pre-training \cite{qiu2020pre} has been a popular network initialization technique which has also been introduced to the deep clustering \cite{van2020scan}.
However, the pre-training based initialization is appropriate for representation learning module but has not been well studied for the clustering module. 
Although there has been some initialization schemes on the shallow clustering \cite{celebi2013comparative}, the initialization for the cluster module with deep neural networks is still under investigation.

\subsection{Overlapping Deep Clustering}
The deep clustering methods discussed in this paper largely focus on the partitioning clustering where each instance belong to only one cluster.
Meanwhile, in the real-world scenario, each instance may belong to multiple clusters, \eg users in a social network \cite{xie2013overlapping} may belong to several communities, and the video/audio in the social media may have several tags \cite{wu2020multi}.
Among the deep clustering methods discussed in this paper, if the clustering constraints are conducted on the cluster assignment probabilistic matrix, they can be directly adapted to the overlapping clustering setting.
However, if the training relies on the pseudo hard label of data instances, they may fail in the overlapping clustering setting.
Although the multi-label classification has been widely studied in the literature \cite{liu2021emerging,zhang2013review}, how to adapt to unsupervised clustering learning is still an open research question.

\subsection{Degenerate Solution VS Unbalanced Data}
The degenerate solution \cite{caron2018deep} has been one of the most noteworthy concerns in deep clustering, where all instances may be assigned to a single cluster.
Many deep clustering methods have added extra constraints to overcome this problem \cite{huang2014deep, caron2018deep,ji2019invariant,huang2020deep,van2020scan,li2021contrastive,park2021improving}, among which the entropy of cluster size distribution is the most widely used one.
By maximizing the entropy, the instances are expected to be evenly assigned to each cluster and avoid degenerating to single cluster.
It is worth noting that the success relies on the uniform distribution of ground truth labels which is coincidentally satisfied by the benchmark dataset such as CIFAR10 and CIFAR100.
However, in real-world scenarios, such assumption is too strict and most datasets are unbalanced or long tailed \cite{wang2016training}.
The divergent target of uniform distribution and unbalanced dataset will seriously weaken the deep clustering.
Recently, clustering unbalanced data \cite{fang2021unbalanced,liang2020lr} has attracted increasing attention and this will boost the performance of clustering in real-world applications.


\subsection{Boosting Representation with Deep Clustering}
Throughout this paper, we can find that a good representation is essential for clustering.
Although the clustering friendly representation learning has been studied in the literature, they are designed for the specific shallow clustering method.
On the contra, the clustering structure denotes the high-order pattern of the dataset, which should be preserved in the comprehensive representation \cite{li2020prototypical}. 
The deep clustering methods discussed in this paper focus on how to incorporate the representation learning to boost the clustering, meanwhile, how to in turn boost the representation learning by the clustering is still to be studied.

\subsection{Deep Clustering Explanation}
As an unsupervised task, the clustering process usually lacks human priors such as label semantics, number of clusters\cite{moser2007joint}, which makes the clustering results hard to explain or understand.
Some methods \cite{zhang2021deep} have already combined the tags provided by the users to boost the explanation of the clustering results.
However, it relies on the accurate human tagging which may not be realized in practical.
With the development of causal inference in deep learning, the explanation of clustering among instances is hopefully to be improved.
Both the research and industry community are expecting a generalized explanation framework for clustering, especially on the high-dimensional data. 
To conclude, how to utilize the casual inference techniques in the clustering is of great importance and deserve more attention.

\subsection{Transfer Learning with Deep Clustering}
Transfer learning \cite{pan2009survey} aims at bridging the gap between the training and test datasets with distribution shift.
The general idea is to transfer the knowledge from the known data to the unknown test data.
Recently, deep clustering is playing an increasing important role in unsupervised transfer learning \cite{xue2016multi,deng2019cluster,tang2020unsupervised,zhou2021cluster,Zhou_2021_ICCV}, where the target domain is in the unsupervised manner.
For example, ItClust \cite{hu2020iterative} and MARS \cite{brbic2020mars} has achieved success in scRNA-seq clustering (section \ref{medical}), AD-Cluster \cite{zhai2020ad} that use clustering has improved domain adaptive person re-identification.
On the other way around, unsupervised transfer learning methods can also benefit deep clustering. 
Take UCDS \cite{menapace2020learning} as an example, the unsupervised domain adaptation is used to perform clustering among variant domains. Distribution shift is one of the key factor affecting the performance of machine learning models, including deep clustering. Perform clustering analysis can give further understanding on unsupervised target domain, but how to transfer clustering result to knowledge and effectively minimize the distribution shift based on clustering result can be further explored.

\subsection{Clustering with Anomalies}
In subsection \ref{app:anomaly}, we have discussed the applications of deep clustering in anomaly detection where the instances are well clustered.
Concerning the existence of anomaly instances in the dataset, the clustering may also be influenced and mutually restrict, since most existing deep clustering methods have no specific response to the outliers.
The classic K-means method is known to be sensitive to outliers, although there has been a few works on overcoming this problem \cite{liu2019clustering}, they are designed for shallow clustering methods. 
To this end, how to improve the deep clustering robustness to the anomaly instances and gradually improve the clustering performance by reducing the detected anomaly instances is still an open research question.

\subsection{Efficient Training VS Global Modeling}
To improve the training efficiency and scalability, most existing deep clustering methods have utilized the mini-batch training strategy, where the instances are separated into batches and the model is updated with each batch.
This is suitable for the task where the instances are independent on each other, such as classification and regression.
However, as the deep clustering heavily relies on the complicated relationship among instances, such mini-batch training may lose the ability of global modeling.
Although some existing methods have utilized the cluster representations or prototypes \cite{li2020prototypical} to store the global information, how to balance the training efficiency and model capability is still worth studying.

%% file: conclusion.tex
In this survey, we present a comprehensive and up-to-date overview of the deep clustering research field.
We first summarize the cornerstones of deep clustering namely the representation learning module and the clustering module, followed by the representative design.
Based on the ways of interaction between representation learning module and clustering module, we present the taxonomy of existing methods namely: multi-stage deep clustering, iterative deep clustering, generative deep clustering and simultaneous deep clustering.
Then, we collect the benchmark datasets, metrics for evaluation and applications of deep clustering.
Last but not least, we discuss the future directions in deep clustering that have potential opportunities.

%% file: Main_Document__PDF_.bbl

\begin{thebibliography}{246}


\ifx \showCODEN    \undefined \def \showCODEN     #1{\unskip}     \fi
\ifx \showDOI      \undefined \def \showDOI       #1{#1}\fi
\ifx \showISBNx    \undefined \def \showISBNx     #1{\unskip}     \fi
\ifx \showISBNxiii \undefined \def \showISBNxiii  #1{\unskip}     \fi
\ifx \showISSN     \undefined \def \showISSN      #1{\unskip}     \fi
\ifx \showLCCN     \undefined \def \showLCCN      #1{\unskip}     \fi
\ifx \shownote     \undefined \def \shownote      #1{#1}          \fi
\ifx \showarticletitle \undefined \def \showarticletitle #1{#1}   \fi
\ifx \showURL      \undefined \def \showURL       {\relax}        \fi
\providecommand\bibfield[2]{#2}
\providecommand\bibinfo[2]{#2}
\providecommand\natexlab[1]{#1}
\providecommand\showeprint[2][]{arXiv:#2}

\bibitem[\protect\citeauthoryear{Abukmeil, Ferrari, Genovese, Piuri, and
  Scotti}{Abukmeil et~al\mbox{.}}{2021}]%
        {abukmeil2021survey}
\bibfield{author}{\bibinfo{person}{Mohanad Abukmeil}, \bibinfo{person}{Stefano
  Ferrari}, \bibinfo{person}{Angelo Genovese}, \bibinfo{person}{Vincenzo
  Piuri}, {and} \bibinfo{person}{Fabio Scotti}.}
  \bibinfo{year}{2021}\natexlab{}.
\newblock \showarticletitle{A survey of unsupervised generative models for
  exploratory data analysis and representation learning}.
\newblock \bibinfo{journal}{\emph{Acm computing surveys (csur)}}
  \bibinfo{volume}{54}, \bibinfo{number}{5} (\bibinfo{year}{2021}),
  \bibinfo{pages}{1--40}.
\newblock


\bibitem[\protect\citeauthoryear{Alami, Meknassi, En-nahnahi, El~Adlouni, and
  Ammor}{Alami et~al\mbox{.}}{2021}]%
        {alami2021unsupervised}
\bibfield{author}{\bibinfo{person}{Nabil Alami}, \bibinfo{person}{Mohammed
  Meknassi}, \bibinfo{person}{Noureddine En-nahnahi}, \bibinfo{person}{Yassine
  El~Adlouni}, {and} \bibinfo{person}{Ouafae Ammor}.}
  \bibinfo{year}{2021}\natexlab{}.
\newblock \showarticletitle{Unsupervised neural networks for automatic Arabic
  text summarization using document clustering and topic modeling}.
\newblock \bibinfo{journal}{\emph{Expert Systems with Applications}}
  \bibinfo{volume}{172} (\bibinfo{year}{2021}), \bibinfo{pages}{114652}.
\newblock


\bibitem[\protect\citeauthoryear{Aljalbout, Golkov, Siddiqui, Strobel, and
  Cremers}{Aljalbout et~al\mbox{.}}{2018}]%
        {aljalbout2018clustering}
\bibfield{author}{\bibinfo{person}{Elie Aljalbout}, \bibinfo{person}{Vladimir
  Golkov}, \bibinfo{person}{Yawar Siddiqui}, \bibinfo{person}{Maximilian
  Strobel}, {and} \bibinfo{person}{Daniel Cremers}.}
  \bibinfo{year}{2018}\natexlab{}.
\newblock \showarticletitle{Clustering with deep learning: Taxonomy and new
  methods}.
\newblock \bibinfo{journal}{\emph{arXiv preprint arXiv:1801.07648}}
  (\bibinfo{year}{2018}).
\newblock


\bibitem[\protect\citeauthoryear{Alwassel, Mahajan, Korbar, Torresani, Ghanem,
  and Tran}{Alwassel et~al\mbox{.}}{2020}]%
        {alwassel2020self}
\bibfield{author}{\bibinfo{person}{Humam Alwassel}, \bibinfo{person}{Dhruv
  Mahajan}, \bibinfo{person}{Bruno Korbar}, \bibinfo{person}{Lorenzo
  Torresani}, \bibinfo{person}{Bernard Ghanem}, {and} \bibinfo{person}{Du
  Tran}.} \bibinfo{year}{2020}\natexlab{}.
\newblock \showarticletitle{Self-supervised learning by cross-modal audio-video
  clustering}.
\newblock \bibinfo{journal}{\emph{Advances in Neural Information Processing
  Systems}}  \bibinfo{volume}{33} (\bibinfo{year}{2020}),
  \bibinfo{pages}{9758--9770}.
\newblock


\bibitem[\protect\citeauthoryear{Arifin and Asano}{Arifin and Asano}{2006}]%
        {arifin2006image}
\bibfield{author}{\bibinfo{person}{Agus~Zainal Arifin} {and}
  \bibinfo{person}{Akira Asano}.} \bibinfo{year}{2006}\natexlab{}.
\newblock \showarticletitle{Image segmentation by histogram thresholding using
  hierarchical cluster analysis}.
\newblock \bibinfo{journal}{\emph{Pattern recognition letters}}
  \bibinfo{volume}{27}, \bibinfo{number}{13} (\bibinfo{year}{2006}),
  \bibinfo{pages}{1515--1521}.
\newblock


\bibitem[\protect\citeauthoryear{Arjovsky, Chintala, and Bottou}{Arjovsky
  et~al\mbox{.}}{2017}]%
        {arjovsky2017wasserstein}
\bibfield{author}{\bibinfo{person}{Martin Arjovsky}, \bibinfo{person}{Soumith
  Chintala}, {and} \bibinfo{person}{L{\'e}on Bottou}.}
  \bibinfo{year}{2017}\natexlab{}.
\newblock \showarticletitle{Wasserstein generative adversarial networks}. In
  \bibinfo{booktitle}{\emph{International conference on machine learning}}.
  PMLR, \bibinfo{pages}{214--223}.
\newblock


\bibitem[\protect\citeauthoryear{Bachman, Hjelm, and Buchwalter}{Bachman
  et~al\mbox{.}}{2019}]%
        {bachman2019learning}
\bibfield{author}{\bibinfo{person}{Philip Bachman}, \bibinfo{person}{R~Devon
  Hjelm}, {and} \bibinfo{person}{William Buchwalter}.}
  \bibinfo{year}{2019}\natexlab{}.
\newblock \showarticletitle{Learning representations by maximizing mutual
  information across views}.
\newblock \bibinfo{journal}{\emph{Advances in neural information processing
  systems}}  \bibinfo{volume}{32} (\bibinfo{year}{2019}).
\newblock


\bibitem[\protect\citeauthoryear{Bandaragoda, De~Silva, Kleyko, Osipov,
  Wiklund, and Alahakoon}{Bandaragoda et~al\mbox{.}}{2019}]%
        {bandaragoda2019trajectory}
\bibfield{author}{\bibinfo{person}{Tharindu Bandaragoda},
  \bibinfo{person}{Daswin De~Silva}, \bibinfo{person}{Denis Kleyko},
  \bibinfo{person}{Evgeny Osipov}, \bibinfo{person}{Urban Wiklund}, {and}
  \bibinfo{person}{Damminda Alahakoon}.} \bibinfo{year}{2019}\natexlab{}.
\newblock \showarticletitle{Trajectory clustering of road traffic in urban
  environments using incremental machine learning in combination with
  hyperdimensional computing}. In \bibinfo{booktitle}{\emph{2019 IEEE
  intelligent transportation systems conference (ITSC)}}. IEEE,
  \bibinfo{pages}{1664--1670}.
\newblock


\bibitem[\protect\citeauthoryear{Belghazi, Baratin, Rajeshwar, Ozair, Bengio,
  Courville, and Hjelm}{Belghazi et~al\mbox{.}}{2018}]%
        {belghazi2018mutual}
\bibfield{author}{\bibinfo{person}{Mohamed~Ishmael Belghazi},
  \bibinfo{person}{Aristide Baratin}, \bibinfo{person}{Sai Rajeshwar},
  \bibinfo{person}{Sherjil Ozair}, \bibinfo{person}{Yoshua Bengio},
  \bibinfo{person}{Aaron Courville}, {and} \bibinfo{person}{Devon Hjelm}.}
  \bibinfo{year}{2018}\natexlab{}.
\newblock \showarticletitle{Mutual information neural estimation}. In
  \bibinfo{booktitle}{\emph{International conference on machine learning}}.
  PMLR, \bibinfo{pages}{531--540}.
\newblock


\bibitem[\protect\citeauthoryear{Ben-Yosef and Weinshall}{Ben-Yosef and
  Weinshall}{2018}]%
        {ben2018gaussian}
\bibfield{author}{\bibinfo{person}{Matan Ben-Yosef} {and}
  \bibinfo{person}{Daphna Weinshall}.} \bibinfo{year}{2018}\natexlab{}.
\newblock \showarticletitle{Gaussian mixture generative adversarial networks
  for diverse datasets, and the unsupervised clustering of images}.
\newblock \bibinfo{journal}{\emph{arXiv preprint arXiv:1808.10356}}
  (\bibinfo{year}{2018}).
\newblock


\bibitem[\protect\citeauthoryear{Bengio, Courville, and Vincent}{Bengio
  et~al\mbox{.}}{2013}]%
        {bengio2013representation}
\bibfield{author}{\bibinfo{person}{Yoshua Bengio}, \bibinfo{person}{Aaron
  Courville}, {and} \bibinfo{person}{Pascal Vincent}.}
  \bibinfo{year}{2013}\natexlab{}.
\newblock \showarticletitle{Representation learning: A review and new
  perspectives}.
\newblock \bibinfo{journal}{\emph{IEEE transactions on pattern analysis and
  machine intelligence}} \bibinfo{volume}{35}, \bibinfo{number}{8}
  (\bibinfo{year}{2013}), \bibinfo{pages}{1798--1828}.
\newblock


\bibitem[\protect\citeauthoryear{Berkhin}{Berkhin}{2006}]%
        {berkhin2006survey}
\bibfield{author}{\bibinfo{person}{Pavel Berkhin}.}
  \bibinfo{year}{2006}\natexlab{}.
\newblock \showarticletitle{A survey of clustering data mining techniques}.
\newblock In \bibinfo{booktitle}{\emph{Grouping multidimensional data}}.
  \bibinfo{publisher}{Springer}, \bibinfo{pages}{25--71}.
\newblock


\bibitem[\protect\citeauthoryear{Bianchi, Grattarola, and Alippi}{Bianchi
  et~al\mbox{.}}{2020}]%
        {bianchi2020spectral}
\bibfield{author}{\bibinfo{person}{Filippo~Maria Bianchi},
  \bibinfo{person}{Daniele Grattarola}, {and} \bibinfo{person}{Cesare Alippi}.}
  \bibinfo{year}{2020}\natexlab{}.
\newblock \showarticletitle{Spectral clustering with graph neural networks for
  graph pooling}. In \bibinfo{booktitle}{\emph{International Conference on
  Machine Learning}}. PMLR, \bibinfo{pages}{874--883}.
\newblock


\bibitem[\protect\citeauthoryear{Blondel, Guillaume, Lambiotte, and
  Lefebvre}{Blondel et~al\mbox{.}}{2008}]%
        {blondel2008fast}
\bibfield{author}{\bibinfo{person}{Vincent~D Blondel},
  \bibinfo{person}{Jean-Loup Guillaume}, \bibinfo{person}{Renaud Lambiotte},
  {and} \bibinfo{person}{Etienne Lefebvre}.} \bibinfo{year}{2008}\natexlab{}.
\newblock \showarticletitle{Fast unfolding of communities in large networks}.
\newblock \bibinfo{journal}{\emph{Journal of statistical mechanics: theory and
  experiment}} \bibinfo{volume}{2008}, \bibinfo{number}{10}
  (\bibinfo{year}{2008}), \bibinfo{pages}{P10008}.
\newblock


\bibitem[\protect\citeauthoryear{Bo, Wang, Shi, Zhu, Lu, and Cui}{Bo
  et~al\mbox{.}}{2020}]%
        {bo2020structural}
\bibfield{author}{\bibinfo{person}{Deyu Bo}, \bibinfo{person}{Xiao Wang},
  \bibinfo{person}{Chuan Shi}, \bibinfo{person}{Meiqi Zhu},
  \bibinfo{person}{Emiao Lu}, {and} \bibinfo{person}{Peng Cui}.}
  \bibinfo{year}{2020}\natexlab{}.
\newblock \showarticletitle{Structural deep clustering network}. In
  \bibinfo{booktitle}{\emph{Proceedings of The Web Conference 2020}}.
  \bibinfo{pages}{1400--1410}.
\newblock


\bibitem[\protect\citeauthoryear{Borgwardt, Ong, Sch{\"o}nauer, Vishwanathan,
  Smola, and Kriegel}{Borgwardt et~al\mbox{.}}{2005}]%
        {borgwardt2005protein}
\bibfield{author}{\bibinfo{person}{Karsten~M Borgwardt},
  \bibinfo{person}{Cheng~Soon Ong}, \bibinfo{person}{Stefan Sch{\"o}nauer},
  \bibinfo{person}{SVN Vishwanathan}, \bibinfo{person}{Alex~J Smola}, {and}
  \bibinfo{person}{Hans-Peter Kriegel}.} \bibinfo{year}{2005}\natexlab{}.
\newblock \showarticletitle{Protein function prediction via graph kernels}.
\newblock \bibinfo{journal}{\emph{Bioinformatics}} \bibinfo{volume}{21},
  \bibinfo{number}{suppl\_1} (\bibinfo{year}{2005}), \bibinfo{pages}{i47--i56}.
\newblock


\bibitem[\protect\citeauthoryear{Brbi{\'c}, Zitnik, Wang, Pisco, Altman,
  Darmanis, and Leskovec}{Brbi{\'c} et~al\mbox{.}}{2020}]%
        {brbic2020mars}
\bibfield{author}{\bibinfo{person}{Maria Brbi{\'c}}, \bibinfo{person}{Marinka
  Zitnik}, \bibinfo{person}{Sheng Wang}, \bibinfo{person}{Angela~O Pisco},
  \bibinfo{person}{Russ~B Altman}, \bibinfo{person}{Spyros Darmanis}, {and}
  \bibinfo{person}{Jure Leskovec}.} \bibinfo{year}{2020}\natexlab{}.
\newblock \showarticletitle{MARS: discovering novel cell types across
  heterogeneous single-cell experiments}.
\newblock \bibinfo{journal}{\emph{Nature methods}} \bibinfo{volume}{17},
  \bibinfo{number}{12} (\bibinfo{year}{2020}), \bibinfo{pages}{1200--1206}.
\newblock


\bibitem[\protect\citeauthoryear{Brown, Mann, Ryder, Subbiah, Kaplan, Dhariwal,
  Neelakantan, Shyam, Sastry, Askell, et~al\mbox{.}}{Brown
  et~al\mbox{.}}{2020}]%
        {brown2020language}
\bibfield{author}{\bibinfo{person}{Tom Brown}, \bibinfo{person}{Benjamin Mann},
  \bibinfo{person}{Nick Ryder}, \bibinfo{person}{Melanie Subbiah},
  \bibinfo{person}{Jared~D Kaplan}, \bibinfo{person}{Prafulla Dhariwal},
  \bibinfo{person}{Arvind Neelakantan}, \bibinfo{person}{Pranav Shyam},
  \bibinfo{person}{Girish Sastry}, \bibinfo{person}{Amanda Askell},
  {et~al\mbox{.}}} \bibinfo{year}{2020}\natexlab{}.
\newblock \showarticletitle{Language models are few-shot learners}.
\newblock \bibinfo{journal}{\emph{Advances in neural information processing
  systems}}  \bibinfo{volume}{33} (\bibinfo{year}{2020}),
  \bibinfo{pages}{1877--1901}.
\newblock


\bibitem[\protect\citeauthoryear{Buda, Maki, and Mazurowski}{Buda
  et~al\mbox{.}}{2018}]%
        {buda2018systematic}
\bibfield{author}{\bibinfo{person}{Mateusz Buda}, \bibinfo{person}{Atsuto
  Maki}, {and} \bibinfo{person}{Maciej~A Mazurowski}.}
  \bibinfo{year}{2018}\natexlab{}.
\newblock \showarticletitle{A systematic study of the class imbalance problem
  in convolutional neural networks}.
\newblock \bibinfo{journal}{\emph{Neural Networks}}  \bibinfo{volume}{106}
  (\bibinfo{year}{2018}), \bibinfo{pages}{249--259}.
\newblock


\bibitem[\protect\citeauthoryear{Cao, Ester, Qian, and Zhou}{Cao
  et~al\mbox{.}}{2006}]%
        {cao2006density}
\bibfield{author}{\bibinfo{person}{Feng Cao}, \bibinfo{person}{Martin Ester},
  \bibinfo{person}{Weining Qian}, {and} \bibinfo{person}{Aoying Zhou}.}
  \bibinfo{year}{2006}\natexlab{}.
\newblock \showarticletitle{Density-based clustering over an evolving data
  stream with noise}. In \bibinfo{booktitle}{\emph{Proceedings of the 2006 SIAM
  international conference on data mining}}. SIAM, \bibinfo{pages}{328--339}.
\newblock


\bibitem[\protect\citeauthoryear{Caron, Bojanowski, Joulin, and Douze}{Caron
  et~al\mbox{.}}{2018}]%
        {caron2018deep}
\bibfield{author}{\bibinfo{person}{Mathilde Caron}, \bibinfo{person}{Piotr
  Bojanowski}, \bibinfo{person}{Armand Joulin}, {and} \bibinfo{person}{Matthijs
  Douze}.} \bibinfo{year}{2018}\natexlab{}.
\newblock \showarticletitle{Deep clustering for unsupervised learning of visual
  features}. In \bibinfo{booktitle}{\emph{Proceedings of the European
  Conference on Computer Vision (ECCV)}}. \bibinfo{pages}{132--149}.
\newblock


\bibitem[\protect\citeauthoryear{Caron, Misra, Mairal, Goyal, Bojanowski, and
  Joulin}{Caron et~al\mbox{.}}{2020}]%
        {caron2020unsupervised}
\bibfield{author}{\bibinfo{person}{Mathilde Caron}, \bibinfo{person}{Ishan
  Misra}, \bibinfo{person}{Julien Mairal}, \bibinfo{person}{Priya Goyal},
  \bibinfo{person}{Piotr Bojanowski}, {and} \bibinfo{person}{Armand Joulin}.}
  \bibinfo{year}{2020}\natexlab{}.
\newblock \showarticletitle{Unsupervised learning of visual features by
  contrasting cluster assignments}.
\newblock \bibinfo{journal}{\emph{Advances in Neural Information Processing
  Systems}}  \bibinfo{volume}{33} (\bibinfo{year}{2020}),
  \bibinfo{pages}{9912--9924}.
\newblock


\bibitem[\protect\citeauthoryear{Celebi, Kingravi, and Vela}{Celebi
  et~al\mbox{.}}{2013}]%
        {celebi2013comparative}
\bibfield{author}{\bibinfo{person}{M~Emre Celebi}, \bibinfo{person}{Hassan~A
  Kingravi}, {and} \bibinfo{person}{Patricio~A Vela}.}
  \bibinfo{year}{2013}\natexlab{}.
\newblock \showarticletitle{A comparative study of efficient initialization
  methods for the k-means clustering algorithm}.
\newblock \bibinfo{journal}{\emph{Expert systems with applications}}
  \bibinfo{volume}{40}, \bibinfo{number}{1} (\bibinfo{year}{2013}),
  \bibinfo{pages}{200--210}.
\newblock


\bibitem[\protect\citeauthoryear{{\c{C}}elik, Dada{\c{s}}er-{\c{C}}elik, and
  Dokuz}{{\c{C}}elik et~al\mbox{.}}{2011}]%
        {ccelik2011anomaly}
\bibfield{author}{\bibinfo{person}{Mete {\c{C}}elik}, \bibinfo{person}{Filiz
  Dada{\c{s}}er-{\c{C}}elik}, {and} \bibinfo{person}{Ahmet~{\c{S}}akir Dokuz}.}
  \bibinfo{year}{2011}\natexlab{}.
\newblock \showarticletitle{Anomaly detection in temperature data using DBSCAN
  algorithm}. In \bibinfo{booktitle}{\emph{2011 international symposium on
  innovations in intelligent systems and applications}}. IEEE,
  \bibinfo{pages}{91--95}.
\newblock


\bibitem[\protect\citeauthoryear{Cerquitelli, Di~Corso, Ventura, and
  Chiusano}{Cerquitelli et~al\mbox{.}}{2017}]%
        {cerquitelli2017data}
\bibfield{author}{\bibinfo{person}{Tania Cerquitelli}, \bibinfo{person}{Evelina
  Di~Corso}, \bibinfo{person}{Francesco Ventura}, {and} \bibinfo{person}{Silvia
  Chiusano}.} \bibinfo{year}{2017}\natexlab{}.
\newblock \showarticletitle{Data miners' little helper: data transformation
  activity cues for cluster analysis on document collections}. In
  \bibinfo{booktitle}{\emph{Proceedings of the 7th International Conference on
  Web Intelligence, Mining and Semantics}}. \bibinfo{pages}{1--6}.
\newblock


\bibitem[\protect\citeauthoryear{Chang and Hsu}{Chang and Hsu}{2005}]%
        {chang2005using}
\bibfield{author}{\bibinfo{person}{Hsi-Cheng Chang} {and}
  \bibinfo{person}{Chiun-Chieh Hsu}.} \bibinfo{year}{2005}\natexlab{}.
\newblock \showarticletitle{Using topic keyword clusters for automatic document
  clustering}.
\newblock \bibinfo{journal}{\emph{IEICE TRANSACTIONS on Information and
  Systems}} \bibinfo{volume}{88}, \bibinfo{number}{8} (\bibinfo{year}{2005}),
  \bibinfo{pages}{1852--1860}.
\newblock


\bibitem[\protect\citeauthoryear{Chang, Guo, Wang, Meng, Xiang, and Pan}{Chang
  et~al\mbox{.}}{2019}]%
        {chang2019deep}
\bibfield{author}{\bibinfo{person}{Jianlong Chang}, \bibinfo{person}{Yiwen
  Guo}, \bibinfo{person}{Lingfeng Wang}, \bibinfo{person}{Gaofeng Meng},
  \bibinfo{person}{Shiming Xiang}, {and} \bibinfo{person}{Chunhong Pan}.}
  \bibinfo{year}{2019}\natexlab{}.
\newblock \showarticletitle{Deep discriminative clustering analysis}.
\newblock \bibinfo{journal}{\emph{arXiv preprint arXiv:1905.01681}}
  (\bibinfo{year}{2019}).
\newblock


\bibitem[\protect\citeauthoryear{Chang, Wang, Meng, Xiang, and Pan}{Chang
  et~al\mbox{.}}{2017}]%
        {chang2017deep}
\bibfield{author}{\bibinfo{person}{Jianlong Chang}, \bibinfo{person}{Lingfeng
  Wang}, \bibinfo{person}{Gaofeng Meng}, \bibinfo{person}{Shiming Xiang}, {and}
  \bibinfo{person}{Chunhong Pan}.} \bibinfo{year}{2017}\natexlab{}.
\newblock \showarticletitle{Deep adaptive image clustering}. In
  \bibinfo{booktitle}{\emph{Proceedings of the IEEE international conference on
  computer vision}}. \bibinfo{pages}{5879--5887}.
\newblock


\bibitem[\protect\citeauthoryear{Chen, Chu, Pan, Lu, and Wang}{Chen
  et~al\mbox{.}}{2022}]%
        {chen2022self}
\bibfield{author}{\bibinfo{person}{Nenglun Chen}, \bibinfo{person}{Lei Chu},
  \bibinfo{person}{Hao Pan}, \bibinfo{person}{Yan Lu}, {and}
  \bibinfo{person}{Wenping Wang}.} \bibinfo{year}{2022}\natexlab{}.
\newblock \showarticletitle{Self-Supervised Image Representation Learning with
  Geometric Set Consistency}.
\newblock \bibinfo{journal}{\emph{arXiv preprint arXiv:2203.15361}}
  (\bibinfo{year}{2022}).
\newblock


\bibitem[\protect\citeauthoryear{Chen, Kornblith, Norouzi, and Hinton}{Chen
  et~al\mbox{.}}{2020}]%
        {chen2020simple}
\bibfield{author}{\bibinfo{person}{Ting Chen}, \bibinfo{person}{Simon
  Kornblith}, \bibinfo{person}{Mohammad Norouzi}, {and}
  \bibinfo{person}{Geoffrey Hinton}.} \bibinfo{year}{2020}\natexlab{}.
\newblock \showarticletitle{A simple framework for contrastive learning of
  visual representations}. In \bibinfo{booktitle}{\emph{International
  conference on machine learning}}. PMLR, \bibinfo{pages}{1597--1607}.
\newblock


\bibitem[\protect\citeauthoryear{Chen and Li}{Chen and Li}{2011}]%
        {chen2011anomaly}
\bibfield{author}{\bibinfo{person}{Zhenguo Chen} {and}
  \bibinfo{person}{Yong~Fei Li}.} \bibinfo{year}{2011}\natexlab{}.
\newblock \showarticletitle{Anomaly detection based on enhanced DBScan
  algorithm}.
\newblock \bibinfo{journal}{\emph{Procedia Engineering}}  \bibinfo{volume}{15}
  (\bibinfo{year}{2011}), \bibinfo{pages}{178--182}.
\newblock


\bibitem[\protect\citeauthoryear{Chiang, Whang, and Dhillon}{Chiang
  et~al\mbox{.}}{2012}]%
        {chiang2012scalable}
\bibfield{author}{\bibinfo{person}{Kai-Yang Chiang},
  \bibinfo{person}{Joyce~Jiyoung Whang}, {and} \bibinfo{person}{Inderjit~S
  Dhillon}.} \bibinfo{year}{2012}\natexlab{}.
\newblock \showarticletitle{Scalable clustering of signed networks using
  balance normalized cut}. In \bibinfo{booktitle}{\emph{Proceedings of the 21st
  ACM international conference on Information and knowledge management}}.
  \bibinfo{pages}{615--624}.
\newblock


\bibitem[\protect\citeauthoryear{Chiang, Liu, Si, Li, Bengio, and Hsieh}{Chiang
  et~al\mbox{.}}{2019}]%
        {chiang2019cluster}
\bibfield{author}{\bibinfo{person}{Wei-Lin Chiang}, \bibinfo{person}{Xuanqing
  Liu}, \bibinfo{person}{Si Si}, \bibinfo{person}{Yang Li},
  \bibinfo{person}{Samy Bengio}, {and} \bibinfo{person}{Cho-Jui Hsieh}.}
  \bibinfo{year}{2019}\natexlab{}.
\newblock \showarticletitle{Cluster-gcn: An efficient algorithm for training
  deep and large graph convolutional networks}. In
  \bibinfo{booktitle}{\emph{Proceedings of the 25th ACM SIGKDD International
  Conference on Knowledge Discovery \& Data Mining}}.
  \bibinfo{pages}{257--266}.
\newblock


\bibitem[\protect\citeauthoryear{Chung, Hsu, Tang, and Glass}{Chung
  et~al\mbox{.}}{2019}]%
        {chung2019unsupervised}
\bibfield{author}{\bibinfo{person}{Yu-An Chung}, \bibinfo{person}{Wei-Ning
  Hsu}, \bibinfo{person}{Hao Tang}, {and} \bibinfo{person}{James Glass}.}
  \bibinfo{year}{2019}\natexlab{}.
\newblock \showarticletitle{An unsupervised autoregressive model for speech
  representation learning}.
\newblock \bibinfo{journal}{\emph{arXiv preprint arXiv:1904.03240}}
  (\bibinfo{year}{2019}).
\newblock


\bibitem[\protect\citeauthoryear{Close, Kashef, et~al\mbox{.}}{Close
  et~al\mbox{.}}{2020}]%
        {close2020combining}
\bibfield{author}{\bibinfo{person}{Liam Close}, \bibinfo{person}{Rasha Kashef},
  {et~al\mbox{.}}} \bibinfo{year}{2020}\natexlab{}.
\newblock \showarticletitle{Combining artificial immune system and clustering
  analysis: A stock market anomaly detection model}.
\newblock \bibinfo{journal}{\emph{Journal of Intelligent Learning Systems and
  Applications}} \bibinfo{volume}{12}, \bibinfo{number}{04}
  (\bibinfo{year}{2020}), \bibinfo{pages}{83}.
\newblock


\bibitem[\protect\citeauthoryear{Coleman and Andrews}{Coleman and
  Andrews}{1979}]%
        {coleman1979image}
\bibfield{author}{\bibinfo{person}{Guy~Barrett Coleman} {and}
  \bibinfo{person}{Harry~C Andrews}.} \bibinfo{year}{1979}\natexlab{}.
\newblock \showarticletitle{Image segmentation by clustering}.
\newblock \bibinfo{journal}{\emph{Proc. IEEE}} \bibinfo{volume}{67},
  \bibinfo{number}{5} (\bibinfo{year}{1979}), \bibinfo{pages}{773--785}.
\newblock


\bibitem[\protect\citeauthoryear{Comaniciu and Meer}{Comaniciu and
  Meer}{2002}]%
        {comaniciu2002mean}
\bibfield{author}{\bibinfo{person}{Dorin Comaniciu} {and}
  \bibinfo{person}{Peter Meer}.} \bibinfo{year}{2002}\natexlab{}.
\newblock \showarticletitle{Mean shift: A robust approach toward feature space
  analysis}.
\newblock \bibinfo{journal}{\emph{IEEE Transactions on pattern analysis and
  machine intelligence}} \bibinfo{volume}{24}, \bibinfo{number}{5}
  (\bibinfo{year}{2002}), \bibinfo{pages}{603--619}.
\newblock


\bibitem[\protect\citeauthoryear{Dang, Deng, Yang, and Huang}{Dang
  et~al\mbox{.}}{2021a}]%
        {dang2021doubly}
\bibfield{author}{\bibinfo{person}{Zhiyuan Dang}, \bibinfo{person}{Cheng Deng},
  \bibinfo{person}{Xu Yang}, {and} \bibinfo{person}{Heng Huang}.}
  \bibinfo{year}{2021}\natexlab{a}.
\newblock \showarticletitle{Doubly contrastive deep clustering}.
\newblock \bibinfo{journal}{\emph{arXiv preprint arXiv:2103.05484}}
  (\bibinfo{year}{2021}).
\newblock


\bibitem[\protect\citeauthoryear{Dang, Deng, Yang, Wei, and Huang}{Dang
  et~al\mbox{.}}{2021b}]%
        {dang2021nearest}
\bibfield{author}{\bibinfo{person}{Zhiyuan Dang}, \bibinfo{person}{Cheng Deng},
  \bibinfo{person}{Xu Yang}, \bibinfo{person}{Kun Wei}, {and}
  \bibinfo{person}{Heng Huang}.} \bibinfo{year}{2021}\natexlab{b}.
\newblock \showarticletitle{Nearest Neighbor Matching for Deep Clustering}. In
  \bibinfo{booktitle}{\emph{Proceedings of the IEEE/CVF Conference on Computer
  Vision and Pattern Recognition}}. \bibinfo{pages}{13693--13702}.
\newblock


\bibitem[\protect\citeauthoryear{Dao, Colak, Salari, Moser, Davicioni,
  Sch{\"o}nhuth, and Ester}{Dao et~al\mbox{.}}{2010}]%
        {dao2010inferring}
\bibfield{author}{\bibinfo{person}{Phuong Dao}, \bibinfo{person}{Recep Colak},
  \bibinfo{person}{Raheleh Salari}, \bibinfo{person}{Flavia Moser},
  \bibinfo{person}{Elai Davicioni}, \bibinfo{person}{Alexander Sch{\"o}nhuth},
  {and} \bibinfo{person}{Martin Ester}.} \bibinfo{year}{2010}\natexlab{}.
\newblock \showarticletitle{Inferring cancer subnetwork markers using
  density-constrained biclustering}.
\newblock \bibinfo{journal}{\emph{Bioinformatics}} \bibinfo{volume}{26},
  \bibinfo{number}{18} (\bibinfo{year}{2010}), \bibinfo{pages}{i625--i631}.
\newblock


\bibitem[\protect\citeauthoryear{Debnath, Lopez~de Compadre, Debnath,
  Shusterman, and Hansch}{Debnath et~al\mbox{.}}{1991}]%
        {debnath1991structure}
\bibfield{author}{\bibinfo{person}{Asim~Kumar Debnath}, \bibinfo{person}{Rosa~L
  Lopez~de Compadre}, \bibinfo{person}{Gargi Debnath}, \bibinfo{person}{Alan~J
  Shusterman}, {and} \bibinfo{person}{Corwin Hansch}.}
  \bibinfo{year}{1991}\natexlab{}.
\newblock \showarticletitle{Structure-activity relationship of mutagenic
  aromatic and heteroaromatic nitro compounds. correlation with molecular
  orbital energies and hydrophobicity}.
\newblock \bibinfo{journal}{\emph{Journal of medicinal chemistry}}
  \bibinfo{volume}{34}, \bibinfo{number}{2} (\bibinfo{year}{1991}),
  \bibinfo{pages}{786--797}.
\newblock


\bibitem[\protect\citeauthoryear{Deng, Luo, and Zhu}{Deng
  et~al\mbox{.}}{2019}]%
        {deng2019cluster}
\bibfield{author}{\bibinfo{person}{Zhijie Deng}, \bibinfo{person}{Yucen Luo},
  {and} \bibinfo{person}{Jun Zhu}.} \bibinfo{year}{2019}\natexlab{}.
\newblock \showarticletitle{Cluster alignment with a teacher for unsupervised
  domain adaptation}. In \bibinfo{booktitle}{\emph{Proceedings of the IEEE/CVF
  International Conference on Computer Vision}}. \bibinfo{pages}{9944--9953}.
\newblock


\bibitem[\protect\citeauthoryear{Devlin, Chang, Lee, and Toutanova}{Devlin
  et~al\mbox{.}}{2018}]%
        {devlin2018bert}
\bibfield{author}{\bibinfo{person}{Jacob Devlin}, \bibinfo{person}{Ming-Wei
  Chang}, \bibinfo{person}{Kenton Lee}, {and} \bibinfo{person}{Kristina
  Toutanova}.} \bibinfo{year}{2018}\natexlab{}.
\newblock \showarticletitle{Bert: Pre-training of deep bidirectional
  transformers for language understanding}.
\newblock \bibinfo{journal}{\emph{arXiv preprint arXiv:1810.04805}}
  (\bibinfo{year}{2018}).
\newblock


\bibitem[\protect\citeauthoryear{Dhillon, Guan, and Kulis}{Dhillon
  et~al\mbox{.}}{2007}]%
        {dhillon2007weighted}
\bibfield{author}{\bibinfo{person}{Inderjit~S Dhillon},
  \bibinfo{person}{Yuqiang Guan}, {and} \bibinfo{person}{Brian Kulis}.}
  \bibinfo{year}{2007}\natexlab{}.
\newblock \showarticletitle{Weighted graph cuts without eigenvectors a
  multilevel approach}.
\newblock \bibinfo{journal}{\emph{IEEE transactions on pattern analysis and
  machine intelligence}} \bibinfo{volume}{29}, \bibinfo{number}{11}
  (\bibinfo{year}{2007}), \bibinfo{pages}{1944--1957}.
\newblock


\bibitem[\protect\citeauthoryear{Dilokthanakul, Mediano, Garnelo, Lee,
  Salimbeni, Arulkumaran, and Shanahan}{Dilokthanakul et~al\mbox{.}}{2016}]%
        {dilokthanakul2016deep}
\bibfield{author}{\bibinfo{person}{Nat Dilokthanakul},
  \bibinfo{person}{Pedro~AM Mediano}, \bibinfo{person}{Marta Garnelo},
  \bibinfo{person}{Matthew~CH Lee}, \bibinfo{person}{Hugh Salimbeni},
  \bibinfo{person}{Kai Arulkumaran}, {and} \bibinfo{person}{Murray Shanahan}.}
  \bibinfo{year}{2016}\natexlab{}.
\newblock \showarticletitle{Deep unsupervised clustering with gaussian mixture
  variational autoencoders}.
\newblock \bibinfo{journal}{\emph{arXiv preprint arXiv:1611.02648}}
  (\bibinfo{year}{2016}).
\newblock


\bibitem[\protect\citeauthoryear{Do, Tran, and Venkatesh}{Do
  et~al\mbox{.}}{2021}]%
        {do2021clustering}
\bibfield{author}{\bibinfo{person}{Kien Do}, \bibinfo{person}{Truyen Tran},
  {and} \bibinfo{person}{Svetha Venkatesh}.} \bibinfo{year}{2021}\natexlab{}.
\newblock \showarticletitle{Clustering by maximizing mutual information across
  views}. In \bibinfo{booktitle}{\emph{Proceedings of the IEEE/CVF
  International Conference on Computer Vision}}. \bibinfo{pages}{9928--9938}.
\newblock


\bibitem[\protect\citeauthoryear{Dosovitskiy, Beyer, Kolesnikov, Weissenborn,
  Zhai, Unterthiner, Dehghani, Minderer, Heigold, Gelly,
  et~al\mbox{.}}{Dosovitskiy et~al\mbox{.}}{2020}]%
        {dosovitskiy2020image}
\bibfield{author}{\bibinfo{person}{Alexey Dosovitskiy}, \bibinfo{person}{Lucas
  Beyer}, \bibinfo{person}{Alexander Kolesnikov}, \bibinfo{person}{Dirk
  Weissenborn}, \bibinfo{person}{Xiaohua Zhai}, \bibinfo{person}{Thomas
  Unterthiner}, \bibinfo{person}{Mostafa Dehghani}, \bibinfo{person}{Matthias
  Minderer}, \bibinfo{person}{Georg Heigold}, \bibinfo{person}{Sylvain Gelly},
  {et~al\mbox{.}}} \bibinfo{year}{2020}\natexlab{}.
\newblock \showarticletitle{An image is worth 16x16 words: Transformers for
  image recognition at scale}.
\newblock \bibinfo{journal}{\emph{arXiv preprint arXiv:2010.11929}}
  (\bibinfo{year}{2020}).
\newblock


\bibitem[\protect\citeauthoryear{Dzisevi{\v{c}} and
  {\v{S}}e{\v{s}}ok}{Dzisevi{\v{c}} and {\v{S}}e{\v{s}}ok}{2019}]%
        {dzisevivc2019text}
\bibfield{author}{\bibinfo{person}{Robert Dzisevi{\v{c}}} {and}
  \bibinfo{person}{Dmitrij {\v{S}}e{\v{s}}ok}.}
  \bibinfo{year}{2019}\natexlab{}.
\newblock \showarticletitle{Text classification using different feature
  extraction approaches}. In \bibinfo{booktitle}{\emph{2019 Open Conference of
  Electrical, Electronic and Information Sciences (eStream)}}. IEEE,
  \bibinfo{pages}{1--4}.
\newblock


\bibitem[\protect\citeauthoryear{Eraslan, Simon, Mircea, Mueller, and
  Theis}{Eraslan et~al\mbox{.}}{2019}]%
        {eraslan2019single}
\bibfield{author}{\bibinfo{person}{G{\"o}kcen Eraslan},
  \bibinfo{person}{Lukas~M Simon}, \bibinfo{person}{Maria Mircea},
  \bibinfo{person}{Nikola~S Mueller}, {and} \bibinfo{person}{Fabian~J Theis}.}
  \bibinfo{year}{2019}\natexlab{}.
\newblock \showarticletitle{Single-cell RNA-seq denoising using a deep count
  autoencoder}.
\newblock \bibinfo{journal}{\emph{Nature communications}} \bibinfo{volume}{10},
  \bibinfo{number}{1} (\bibinfo{year}{2019}), \bibinfo{pages}{1--14}.
\newblock


\bibitem[\protect\citeauthoryear{Ester}{Ester}{2018}]%
        {ester2018density}
\bibfield{author}{\bibinfo{person}{Martin Ester}.}
  \bibinfo{year}{2018}\natexlab{}.
\newblock \showarticletitle{Density-based clustering}.
\newblock \bibinfo{journal}{\emph{Data Clustering}} (\bibinfo{year}{2018}),
  \bibinfo{pages}{111--127}.
\newblock


\bibitem[\protect\citeauthoryear{Ester, Kriegel, Sander, Xu,
  et~al\mbox{.}}{Ester et~al\mbox{.}}{1996}]%
        {ester1996density}
\bibfield{author}{\bibinfo{person}{Martin Ester}, \bibinfo{person}{Hans-Peter
  Kriegel}, \bibinfo{person}{J{\"o}rg Sander}, \bibinfo{person}{Xiaowei Xu},
  {et~al\mbox{.}}} \bibinfo{year}{1996}\natexlab{}.
\newblock \showarticletitle{A density-based algorithm for discovering clusters
  in large spatial databases with noise.}. In \bibinfo{booktitle}{\emph{kdd}},
  Vol.~\bibinfo{volume}{96}. \bibinfo{pages}{226--231}.
\newblock


\bibitem[\protect\citeauthoryear{Fang, Hu, Zhou, and Wu}{Fang
  et~al\mbox{.}}{2021}]%
        {fang2021unbalanced}
\bibfield{author}{\bibinfo{person}{Xiang Fang}, \bibinfo{person}{Yuchong Hu},
  \bibinfo{person}{Pan Zhou}, {and} \bibinfo{person}{Dapeng~Oliver Wu}.}
  \bibinfo{year}{2021}\natexlab{}.
\newblock \showarticletitle{Unbalanced Incomplete Multi-view Clustering via the
  Scheme of View Evolution: Weak Views are Meat; Strong Views do Eat}.
\newblock \bibinfo{journal}{\emph{IEEE Transactions on Emerging Topics in
  Computational Intelligence}} (\bibinfo{year}{2021}).
\newblock


\bibitem[\protect\citeauthoryear{Feichtenhofer, Fan, Xiong, Girshick, and
  He}{Feichtenhofer et~al\mbox{.}}{2021}]%
        {feichtenhofer2021large}
\bibfield{author}{\bibinfo{person}{Christoph Feichtenhofer},
  \bibinfo{person}{Haoqi Fan}, \bibinfo{person}{Bo Xiong},
  \bibinfo{person}{Ross Girshick}, {and} \bibinfo{person}{Kaiming He}.}
  \bibinfo{year}{2021}\natexlab{}.
\newblock \showarticletitle{A large-scale study on unsupervised spatiotemporal
  representation learning}. In \bibinfo{booktitle}{\emph{Proceedings of the
  IEEE/CVF Conference on Computer Vision and Pattern Recognition}}.
  \bibinfo{pages}{3299--3309}.
\newblock


\bibitem[\protect\citeauthoryear{Flake, Lawrence, Giles, and Coetzee}{Flake
  et~al\mbox{.}}{2002}]%
        {flake2002self}
\bibfield{author}{\bibinfo{person}{Gary~William Flake}, \bibinfo{person}{Steve
  Lawrence}, \bibinfo{person}{C~Lee Giles}, {and} \bibinfo{person}{Frans~M
  Coetzee}.} \bibinfo{year}{2002}\natexlab{}.
\newblock \showarticletitle{Self-organization and identification of web
  communities}.
\newblock \bibinfo{journal}{\emph{Computer}} \bibinfo{volume}{35},
  \bibinfo{number}{3} (\bibinfo{year}{2002}), \bibinfo{pages}{66--70}.
\newblock


\bibitem[\protect\citeauthoryear{Fortunato}{Fortunato}{2010}]%
        {fortunato2010community}
\bibfield{author}{\bibinfo{person}{Santo Fortunato}.}
  \bibinfo{year}{2010}\natexlab{}.
\newblock \showarticletitle{Community detection in graphs}.
\newblock \bibinfo{journal}{\emph{Physics reports}} \bibinfo{volume}{486},
  \bibinfo{number}{3-5} (\bibinfo{year}{2010}), \bibinfo{pages}{75--174}.
\newblock


\bibitem[\protect\citeauthoryear{Frades and Matthiesen}{Frades and
  Matthiesen}{2010}]%
        {frades2010overview}
\bibfield{author}{\bibinfo{person}{Itziar Frades} {and} \bibinfo{person}{Rune
  Matthiesen}.} \bibinfo{year}{2010}\natexlab{}.
\newblock \showarticletitle{Overview on techniques in cluster analysis}.
\newblock \bibinfo{journal}{\emph{Bioinformatics methods in clinical research}}
  (\bibinfo{year}{2010}), \bibinfo{pages}{81--107}.
\newblock


\bibitem[\protect\citeauthoryear{Gan, Dong, Zhou, Gao, and Dong}{Gan
  et~al\mbox{.}}{2021}]%
        {gan2021learning}
\bibfield{author}{\bibinfo{person}{Yanhai Gan}, \bibinfo{person}{Xinghui Dong},
  \bibinfo{person}{Huiyu Zhou}, \bibinfo{person}{Feng Gao}, {and}
  \bibinfo{person}{Junyu Dong}.} \bibinfo{year}{2021}\natexlab{}.
\newblock \showarticletitle{Learning the Precise Feature for Cluster
  Assignment}.
\newblock \bibinfo{journal}{\emph{IEEE Transactions on Cybernetics}}
  (\bibinfo{year}{2021}).
\newblock


\bibitem[\protect\citeauthoryear{Gao, Griffith, Ester, and Jones}{Gao
  et~al\mbox{.}}{2006}]%
        {gao2006discovering}
\bibfield{author}{\bibinfo{person}{Byron~J Gao}, \bibinfo{person}{Obi~L
  Griffith}, \bibinfo{person}{Martin Ester}, {and} \bibinfo{person}{Steven~JM
  Jones}.} \bibinfo{year}{2006}\natexlab{}.
\newblock \showarticletitle{Discovering significant opsm subspace clusters in
  massive gene expression data}. In \bibinfo{booktitle}{\emph{Proceedings of
  the 12th ACM SIGKDD international conference on knowledge discovery and data
  mining}}. \bibinfo{pages}{922--928}.
\newblock


\bibitem[\protect\citeauthoryear{Ghasedi, Wang, Deng, and Huang}{Ghasedi
  et~al\mbox{.}}{2019}]%
        {ghasedi2019balanced}
\bibfield{author}{\bibinfo{person}{Kamran Ghasedi}, \bibinfo{person}{Xiaoqian
  Wang}, \bibinfo{person}{Cheng Deng}, {and} \bibinfo{person}{Heng Huang}.}
  \bibinfo{year}{2019}\natexlab{}.
\newblock \showarticletitle{Balanced self-paced learning for generative
  adversarial clustering network}. In \bibinfo{booktitle}{\emph{Proceedings of
  the IEEE/CVF Conference on Computer Vision and Pattern Recognition}}.
  \bibinfo{pages}{4391--4400}.
\newblock


\bibitem[\protect\citeauthoryear{Ghasedi~Dizaji, Herandi, Deng, Cai, and
  Huang}{Ghasedi~Dizaji et~al\mbox{.}}{2017}]%
        {ghasedi2017deep}
\bibfield{author}{\bibinfo{person}{Kamran Ghasedi~Dizaji},
  \bibinfo{person}{Amirhossein Herandi}, \bibinfo{person}{Cheng Deng},
  \bibinfo{person}{Weidong Cai}, {and} \bibinfo{person}{Heng Huang}.}
  \bibinfo{year}{2017}\natexlab{}.
\newblock \showarticletitle{Deep clustering via joint convolutional autoencoder
  embedding and relative entropy minimization}. In
  \bibinfo{booktitle}{\emph{Proceedings of the IEEE international conference on
  computer vision}}. \bibinfo{pages}{5736--5745}.
\newblock


\bibitem[\protect\citeauthoryear{Girvan and Newman}{Girvan and Newman}{2002}]%
        {girvan2002community}
\bibfield{author}{\bibinfo{person}{Michelle Girvan} {and}
  \bibinfo{person}{Mark~EJ Newman}.} \bibinfo{year}{2002}\natexlab{}.
\newblock \showarticletitle{Community structure in social and biological
  networks}.
\newblock \bibinfo{journal}{\emph{Proceedings of the national academy of
  sciences}} \bibinfo{volume}{99}, \bibinfo{number}{12} (\bibinfo{year}{2002}),
  \bibinfo{pages}{7821--7826}.
\newblock


\bibitem[\protect\citeauthoryear{Glorot and Bengio}{Glorot and Bengio}{2010}]%
        {glorot2010understanding}
\bibfield{author}{\bibinfo{person}{Xavier Glorot} {and} \bibinfo{person}{Yoshua
  Bengio}.} \bibinfo{year}{2010}\natexlab{}.
\newblock \showarticletitle{Understanding the difficulty of training deep
  feedforward neural networks}. In \bibinfo{booktitle}{\emph{Proceedings of the
  thirteenth international conference on artificial intelligence and
  statistics}}. JMLR Workshop and Conference Proceedings,
  \bibinfo{pages}{249--256}.
\newblock


\bibitem[\protect\citeauthoryear{Goodfellow, Pouget-Abadie, Mirza, Xu,
  Warde-Farley, Ozair, Courville, and Bengio}{Goodfellow et~al\mbox{.}}{2014}]%
        {goodfellow2014generative}
\bibfield{author}{\bibinfo{person}{Ian Goodfellow}, \bibinfo{person}{Jean
  Pouget-Abadie}, \bibinfo{person}{Mehdi Mirza}, \bibinfo{person}{Bing Xu},
  \bibinfo{person}{David Warde-Farley}, \bibinfo{person}{Sherjil Ozair},
  \bibinfo{person}{Aaron Courville}, {and} \bibinfo{person}{Yoshua Bengio}.}
  \bibinfo{year}{2014}\natexlab{}.
\newblock \showarticletitle{Generative adversarial nets}.
\newblock \bibinfo{journal}{\emph{Advances in neural information processing
  systems}}  \bibinfo{volume}{27} (\bibinfo{year}{2014}).
\newblock


\bibitem[\protect\citeauthoryear{Govindasamy, Arumugam, Zhuang, Kelley, and
  Vellangany}{Govindasamy et~al\mbox{.}}{2018}]%
        {govindasamy2018cluster}
\bibfield{author}{\bibinfo{person}{Ramu Govindasamy},
  \bibinfo{person}{Surendran Arumugam}, \bibinfo{person}{Jingkun Zhuang},
  \bibinfo{person}{Kathleen~M Kelley}, {and} \bibinfo{person}{Isaac
  Vellangany}.} \bibinfo{year}{2018}\natexlab{}.
\newblock \showarticletitle{Cluster analysis of wine market segmentation--a
  consumer based study in the mid-atlantic usa}.
\newblock \bibinfo{journal}{\emph{Economic Affairs}} \bibinfo{volume}{63},
  \bibinfo{number}{1} (\bibinfo{year}{2018}), \bibinfo{pages}{151--157}.
\newblock


\bibitem[\protect\citeauthoryear{Grill, Strub, Altch{\'e}, Tallec, Richemond,
  Buchatskaya, Doersch, Avila~Pires, Guo, Gheshlaghi~Azar, et~al\mbox{.}}{Grill
  et~al\mbox{.}}{2020}]%
        {grill2020bootstrap}
\bibfield{author}{\bibinfo{person}{Jean-Bastien Grill},
  \bibinfo{person}{Florian Strub}, \bibinfo{person}{Florent Altch{\'e}},
  \bibinfo{person}{Corentin Tallec}, \bibinfo{person}{Pierre Richemond},
  \bibinfo{person}{Elena Buchatskaya}, \bibinfo{person}{Carl Doersch},
  \bibinfo{person}{Bernardo Avila~Pires}, \bibinfo{person}{Zhaohan Guo},
  \bibinfo{person}{Mohammad Gheshlaghi~Azar}, {et~al\mbox{.}}}
  \bibinfo{year}{2020}\natexlab{}.
\newblock \showarticletitle{Bootstrap your own latent-a new approach to
  self-supervised learning}.
\newblock \bibinfo{journal}{\emph{Advances in Neural Information Processing
  Systems}}  \bibinfo{volume}{33} (\bibinfo{year}{2020}),
  \bibinfo{pages}{21271--21284}.
\newblock


\bibitem[\protect\citeauthoryear{Guha, Rastogi, and Shim}{Guha
  et~al\mbox{.}}{1998}]%
        {guha1998cure}
\bibfield{author}{\bibinfo{person}{Sudipto Guha}, \bibinfo{person}{Rajeev
  Rastogi}, {and} \bibinfo{person}{Kyuseok Shim}.}
  \bibinfo{year}{1998}\natexlab{}.
\newblock \showarticletitle{CURE: An efficient clustering algorithm for large
  databases}.
\newblock \bibinfo{journal}{\emph{ACM Sigmod record}} \bibinfo{volume}{27},
  \bibinfo{number}{2} (\bibinfo{year}{1998}), \bibinfo{pages}{73--84}.
\newblock


\bibitem[\protect\citeauthoryear{Guo, Wu, Zhu, and Zhang}{Guo
  et~al\mbox{.}}{2017b}]%
        {guo2017combining}
\bibfield{author}{\bibinfo{person}{Ting Guo}, \bibinfo{person}{Jia Wu},
  \bibinfo{person}{Xingquan Zhu}, {and} \bibinfo{person}{Chengqi Zhang}.}
  \bibinfo{year}{2017}\natexlab{b}.
\newblock \showarticletitle{Combining structured node content and topology
  information for networked graph clustering}.
\newblock \bibinfo{journal}{\emph{ACM Transactions on Knowledge Discovery from
  Data (TKDD)}} \bibinfo{volume}{11}, \bibinfo{number}{3}
  (\bibinfo{year}{2017}), \bibinfo{pages}{1--29}.
\newblock


\bibitem[\protect\citeauthoryear{Guo, Gao, Liu, and Yin}{Guo
  et~al\mbox{.}}{2017a}]%
        {guo2017improved}
\bibfield{author}{\bibinfo{person}{Xifeng Guo}, \bibinfo{person}{Long Gao},
  \bibinfo{person}{Xinwang Liu}, {and} \bibinfo{person}{Jianping Yin}.}
  \bibinfo{year}{2017}\natexlab{a}.
\newblock \showarticletitle{Improved Deep Embedded Clustering with Local
  Structure Preservation.}. In \bibinfo{booktitle}{\emph{Ijcai}}.
  \bibinfo{pages}{1753--1759}.
\newblock


\bibitem[\protect\citeauthoryear{Guo, Zhu, Liu, and Yin}{Guo
  et~al\mbox{.}}{2018}]%
        {guo2018deep}
\bibfield{author}{\bibinfo{person}{Xifeng Guo}, \bibinfo{person}{En Zhu},
  \bibinfo{person}{Xinwang Liu}, {and} \bibinfo{person}{Jianping Yin}.}
  \bibinfo{year}{2018}\natexlab{}.
\newblock \showarticletitle{Deep embedded clustering with data augmentation}.
  In \bibinfo{booktitle}{\emph{Asian conference on machine learning}}. PMLR,
  \bibinfo{pages}{550--565}.
\newblock


\bibitem[\protect\citeauthoryear{Hamilton, Ying, and Leskovec}{Hamilton
  et~al\mbox{.}}{2017}]%
        {hamilton2017inductive}
\bibfield{author}{\bibinfo{person}{Will Hamilton}, \bibinfo{person}{Zhitao
  Ying}, {and} \bibinfo{person}{Jure Leskovec}.}
  \bibinfo{year}{2017}\natexlab{}.
\newblock \showarticletitle{Inductive representation learning on large graphs}.
\newblock \bibinfo{journal}{\emph{Advances in neural information processing
  systems}}  \bibinfo{volume}{30} (\bibinfo{year}{2017}).
\newblock


\bibitem[\protect\citeauthoryear{Hassan, Rashid, and Hamarashid}{Hassan
  et~al\mbox{.}}{2021}]%
        {hassan2021novel}
\bibfield{author}{\bibinfo{person}{Bryar~A Hassan}, \bibinfo{person}{Tarik~A
  Rashid}, {and} \bibinfo{person}{Hozan~K Hamarashid}.}
  \bibinfo{year}{2021}\natexlab{}.
\newblock \showarticletitle{A novel cluster detection of COVID-19 patients and
  medical disease conditions using improved evolutionary clustering algorithm
  star}.
\newblock \bibinfo{journal}{\emph{Computers in biology and medicine}}
  \bibinfo{volume}{138} (\bibinfo{year}{2021}), \bibinfo{pages}{104866}.
\newblock


\bibitem[\protect\citeauthoryear{He, Fan, Wu, Xie, and Girshick}{He
  et~al\mbox{.}}{2020}]%
        {he2020momentum}
\bibfield{author}{\bibinfo{person}{Kaiming He}, \bibinfo{person}{Haoqi Fan},
  \bibinfo{person}{Yuxin Wu}, \bibinfo{person}{Saining Xie}, {and}
  \bibinfo{person}{Ross Girshick}.} \bibinfo{year}{2020}\natexlab{}.
\newblock \showarticletitle{Momentum contrast for unsupervised visual
  representation learning}. In \bibinfo{booktitle}{\emph{Proceedings of the
  IEEE/CVF conference on computer vision and pattern recognition}}.
  \bibinfo{pages}{9729--9738}.
\newblock


\bibitem[\protect\citeauthoryear{He, Zhang, Ren, and Sun}{He
  et~al\mbox{.}}{2016}]%
        {he2016deep}
\bibfield{author}{\bibinfo{person}{Kaiming He}, \bibinfo{person}{Xiangyu
  Zhang}, \bibinfo{person}{Shaoqing Ren}, {and} \bibinfo{person}{Jian Sun}.}
  \bibinfo{year}{2016}\natexlab{}.
\newblock \showarticletitle{Deep residual learning for image recognition}. In
  \bibinfo{booktitle}{\emph{Proceedings of the IEEE conference on computer
  vision and pattern recognition}}. \bibinfo{pages}{770--778}.
\newblock


\bibitem[\protect\citeauthoryear{Hinton, Vinyals, Dean, et~al\mbox{.}}{Hinton
  et~al\mbox{.}}{2015}]%
        {hinton2015distilling}
\bibfield{author}{\bibinfo{person}{Geoffrey Hinton}, \bibinfo{person}{Oriol
  Vinyals}, \bibinfo{person}{Jeff Dean}, {et~al\mbox{.}}}
  \bibinfo{year}{2015}\natexlab{}.
\newblock \showarticletitle{Distilling the knowledge in a neural network}.
\newblock \bibinfo{journal}{\emph{arXiv preprint arXiv:1503.02531}}
  \bibinfo{volume}{2}, \bibinfo{number}{7} (\bibinfo{year}{2015}).
\newblock


\bibitem[\protect\citeauthoryear{Hjelm, Fedorov, Lavoie-Marchildon, Grewal,
  Bachman, Trischler, and Bengio}{Hjelm et~al\mbox{.}}{2018}]%
        {hjelm2018learning}
\bibfield{author}{\bibinfo{person}{R~Devon Hjelm}, \bibinfo{person}{Alex
  Fedorov}, \bibinfo{person}{Samuel Lavoie-Marchildon}, \bibinfo{person}{Karan
  Grewal}, \bibinfo{person}{Phil Bachman}, \bibinfo{person}{Adam Trischler},
  {and} \bibinfo{person}{Yoshua Bengio}.} \bibinfo{year}{2018}\natexlab{}.
\newblock \showarticletitle{Learning deep representations by mutual information
  estimation and maximization}.
\newblock \bibinfo{journal}{\emph{arXiv preprint arXiv:1808.06670}}
  (\bibinfo{year}{2018}).
\newblock


\bibitem[\protect\citeauthoryear{Hofmann}{Hofmann}{2013}]%
        {hofmann2013probabilistic}
\bibfield{author}{\bibinfo{person}{Thomas Hofmann}.}
  \bibinfo{year}{2013}\natexlab{}.
\newblock \showarticletitle{Probabilistic latent semantic analysis}.
\newblock \bibinfo{journal}{\emph{arXiv preprint arXiv:1301.6705}}
  (\bibinfo{year}{2013}).
\newblock


\bibitem[\protect\citeauthoryear{Hsu and Lin}{Hsu and Lin}{2017}]%
        {hsu2017cnn}
\bibfield{author}{\bibinfo{person}{Chih-Chung Hsu} {and}
  \bibinfo{person}{Chia-Wen Lin}.} \bibinfo{year}{2017}\natexlab{}.
\newblock \showarticletitle{Cnn-based joint clustering and representation
  learning with feature drift compensation for large-scale image data}.
\newblock \bibinfo{journal}{\emph{IEEE Transactions on Multimedia}}
  \bibinfo{volume}{20}, \bibinfo{number}{2} (\bibinfo{year}{2017}),
  \bibinfo{pages}{421--429}.
\newblock


\bibitem[\protect\citeauthoryear{Hu, Li, Hu, Lyu, Susztak, and Li}{Hu
  et~al\mbox{.}}{2020}]%
        {hu2020iterative}
\bibfield{author}{\bibinfo{person}{Jian Hu}, \bibinfo{person}{Xiangjie Li},
  \bibinfo{person}{Gang Hu}, \bibinfo{person}{Yafei Lyu},
  \bibinfo{person}{Katalin Susztak}, {and} \bibinfo{person}{Mingyao Li}.}
  \bibinfo{year}{2020}\natexlab{}.
\newblock \showarticletitle{Iterative transfer learning with neural network for
  clustering and cell type classification in single-cell RNA-seq analysis}.
\newblock \bibinfo{journal}{\emph{Nature machine intelligence}}
  \bibinfo{volume}{2}, \bibinfo{number}{10} (\bibinfo{year}{2020}),
  \bibinfo{pages}{607--618}.
\newblock


\bibitem[\protect\citeauthoryear{Huang and Gong}{Huang and Gong}{2021}]%
        {huang2021deep}
\bibfield{author}{\bibinfo{person}{Jiabo Huang} {and} \bibinfo{person}{Shaogang
  Gong}.} \bibinfo{year}{2021}\natexlab{}.
\newblock \showarticletitle{Deep clustering by semantic contrastive learning}.
\newblock \bibinfo{journal}{\emph{arXiv preprint arXiv:2103.02662}}
  (\bibinfo{year}{2021}).
\newblock


\bibitem[\protect\citeauthoryear{Huang, Gong, and Zhu}{Huang
  et~al\mbox{.}}{2020}]%
        {huang2020deep}
\bibfield{author}{\bibinfo{person}{Jiabo Huang}, \bibinfo{person}{Shaogang
  Gong}, {and} \bibinfo{person}{Xiatian Zhu}.} \bibinfo{year}{2020}\natexlab{}.
\newblock \showarticletitle{Deep semantic clustering by partition confidence
  maximisation}. In \bibinfo{booktitle}{\emph{Proceedings of the IEEE/CVF
  Conference on Computer Vision and Pattern Recognition}}.
  \bibinfo{pages}{8849--8858}.
\newblock


\bibitem[\protect\citeauthoryear{Huang, Huang, Wang, and Wang}{Huang
  et~al\mbox{.}}{2014}]%
        {huang2014deep}
\bibfield{author}{\bibinfo{person}{Peihao Huang}, \bibinfo{person}{Yan Huang},
  \bibinfo{person}{Wei Wang}, {and} \bibinfo{person}{Liang Wang}.}
  \bibinfo{year}{2014}\natexlab{}.
\newblock \showarticletitle{Deep embedding network for clustering}. In
  \bibinfo{booktitle}{\emph{2014 22nd International conference on pattern
  recognition}}. IEEE, \bibinfo{pages}{1532--1537}.
\newblock


\bibitem[\protect\citeauthoryear{Huang, Chen, Zhang, and Shan}{Huang
  et~al\mbox{.}}{2021}]%
        {huang2021exploring}
\bibfield{author}{\bibinfo{person}{Zhizhong Huang}, \bibinfo{person}{Jie Chen},
  \bibinfo{person}{Junping Zhang}, {and} \bibinfo{person}{Hongming Shan}.}
  \bibinfo{year}{2021}\natexlab{}.
\newblock \showarticletitle{Exploring Non-Contrastive Representation Learning
  for Deep Clustering}.
\newblock \bibinfo{journal}{\emph{arXiv preprint arXiv:2111.11821}}
  (\bibinfo{year}{2021}).
\newblock


\bibitem[\protect\citeauthoryear{Jabbar, Li, and Omar}{Jabbar
  et~al\mbox{.}}{2021}]%
        {jabbar2021survey}
\bibfield{author}{\bibinfo{person}{Abdul Jabbar}, \bibinfo{person}{Xi Li},
  {and} \bibinfo{person}{Bourahla Omar}.} \bibinfo{year}{2021}\natexlab{}.
\newblock \showarticletitle{A survey on generative adversarial networks:
  Variants, applications, and training}.
\newblock \bibinfo{journal}{\emph{ACM Computing Surveys (CSUR)}}
  \bibinfo{volume}{54}, \bibinfo{number}{8} (\bibinfo{year}{2021}),
  \bibinfo{pages}{1--49}.
\newblock


\bibitem[\protect\citeauthoryear{Jain, Murty, and Flynn}{Jain
  et~al\mbox{.}}{1999}]%
        {jain1999data}
\bibfield{author}{\bibinfo{person}{Anil~K Jain}, \bibinfo{person}{M~Narasimha
  Murty}, {and} \bibinfo{person}{Patrick~J Flynn}.}
  \bibinfo{year}{1999}\natexlab{}.
\newblock \showarticletitle{Data clustering: a review}.
\newblock \bibinfo{journal}{\emph{ACM computing surveys (CSUR)}}
  \bibinfo{volume}{31}, \bibinfo{number}{3} (\bibinfo{year}{1999}),
  \bibinfo{pages}{264--323}.
\newblock


\bibitem[\protect\citeauthoryear{Jaiswal, Babu, Zadeh, Banerjee, and
  Makedon}{Jaiswal et~al\mbox{.}}{2020a}]%
        {jaiswal2020survey}
\bibfield{author}{\bibinfo{person}{Ashish Jaiswal},
  \bibinfo{person}{Ashwin~Ramesh Babu}, \bibinfo{person}{Mohammad~Zaki Zadeh},
  \bibinfo{person}{Debapriya Banerjee}, {and} \bibinfo{person}{Fillia
  Makedon}.} \bibinfo{year}{2020}\natexlab{a}.
\newblock \showarticletitle{A survey on contrastive self-supervised learning}.
\newblock \bibinfo{journal}{\emph{Technologies}} \bibinfo{volume}{9},
  \bibinfo{number}{1} (\bibinfo{year}{2020}), \bibinfo{pages}{2}.
\newblock


\bibitem[\protect\citeauthoryear{Jaiswal, Kaushal, Singh, and Biswas}{Jaiswal
  et~al\mbox{.}}{2020b}]%
        {jaiswal2020green}
\bibfield{author}{\bibinfo{person}{Deepak Jaiswal}, \bibinfo{person}{Vikrant
  Kaushal}, \bibinfo{person}{Pankaj~Kumar Singh}, {and}
  \bibinfo{person}{Abhijeet Biswas}.} \bibinfo{year}{2020}\natexlab{b}.
\newblock \showarticletitle{Green market segmentation and consumer profiling: a
  cluster approach to an emerging consumer market}.
\newblock \bibinfo{journal}{\emph{Benchmarking: An International Journal}}
  (\bibinfo{year}{2020}).
\newblock


\bibitem[\protect\citeauthoryear{Janani and Vijayarani}{Janani and
  Vijayarani}{2019}]%
        {janani2019text}
\bibfield{author}{\bibinfo{person}{R Janani} {and} \bibinfo{person}{S
  Vijayarani}.} \bibinfo{year}{2019}\natexlab{}.
\newblock \showarticletitle{Text document clustering using spectral clustering
  algorithm with particle swarm optimization}.
\newblock \bibinfo{journal}{\emph{Expert Systems with Applications}}
  \bibinfo{volume}{134} (\bibinfo{year}{2019}), \bibinfo{pages}{192--200}.
\newblock


\bibitem[\protect\citeauthoryear{Ji, Zhang, Li, Salzmann, and Reid}{Ji
  et~al\mbox{.}}{2017}]%
        {ji2017deep}
\bibfield{author}{\bibinfo{person}{Pan Ji}, \bibinfo{person}{Tong Zhang},
  \bibinfo{person}{Hongdong Li}, \bibinfo{person}{Mathieu Salzmann}, {and}
  \bibinfo{person}{Ian Reid}.} \bibinfo{year}{2017}\natexlab{}.
\newblock \showarticletitle{Deep subspace clustering networks}.
\newblock \bibinfo{journal}{\emph{Advances in neural information processing
  systems}}  \bibinfo{volume}{30} (\bibinfo{year}{2017}).
\newblock


\bibitem[\protect\citeauthoryear{Ji, Sun, Gao, Hu, and Yin}{Ji
  et~al\mbox{.}}{2021a}]%
        {ji2021decoder}
\bibfield{author}{\bibinfo{person}{Qiang Ji}, \bibinfo{person}{Yanfeng Sun},
  \bibinfo{person}{Junbin Gao}, \bibinfo{person}{Yongli Hu}, {and}
  \bibinfo{person}{Baocai Yin}.} \bibinfo{year}{2021}\natexlab{a}.
\newblock \showarticletitle{A Decoder-Free Variational Deep Embedding for
  Unsupervised Clustering}.
\newblock \bibinfo{journal}{\emph{IEEE Transactions on Neural Networks and
  Learning Systems}} (\bibinfo{year}{2021}).
\newblock


\bibitem[\protect\citeauthoryear{Ji, Sun, Hu, and Yin}{Ji
  et~al\mbox{.}}{2021b}]%
        {ji2021variational}
\bibfield{author}{\bibinfo{person}{Qiang Ji}, \bibinfo{person}{Yanfeng Sun},
  \bibinfo{person}{Yongli Hu}, {and} \bibinfo{person}{Baocai Yin}.}
  \bibinfo{year}{2021}\natexlab{b}.
\newblock \showarticletitle{Variational Deep Embedding Clustering by Augmented
  Mutual Information Maximization}. In \bibinfo{booktitle}{\emph{2020 25th
  International Conference on Pattern Recognition (ICPR)}}. IEEE,
  \bibinfo{pages}{2196--2202}.
\newblock


\bibitem[\protect\citeauthoryear{Ji, Henriques, and Vedaldi}{Ji
  et~al\mbox{.}}{2019}]%
        {ji2019invariant}
\bibfield{author}{\bibinfo{person}{Xu Ji}, \bibinfo{person}{Joao~F Henriques},
  {and} \bibinfo{person}{Andrea Vedaldi}.} \bibinfo{year}{2019}\natexlab{}.
\newblock \showarticletitle{Invariant information clustering for unsupervised
  image classification and segmentation}. In
  \bibinfo{booktitle}{\emph{Proceedings of the IEEE/CVF International
  Conference on Computer Vision}}. \bibinfo{pages}{9865--9874}.
\newblock


\bibitem[\protect\citeauthoryear{Jia, Zhang, Zhang, and Wang}{Jia
  et~al\mbox{.}}{2019}]%
        {jia2019communitygan}
\bibfield{author}{\bibinfo{person}{Yuting Jia}, \bibinfo{person}{Qinqin Zhang},
  \bibinfo{person}{Weinan Zhang}, {and} \bibinfo{person}{Xinbing Wang}.}
  \bibinfo{year}{2019}\natexlab{}.
\newblock \showarticletitle{Communitygan: Community detection with generative
  adversarial nets}. In \bibinfo{booktitle}{\emph{The World Wide Web
  Conference}}. \bibinfo{pages}{784--794}.
\newblock


\bibitem[\protect\citeauthoryear{Jiang, Zheng, Tan, Tang, and Zhou}{Jiang
  et~al\mbox{.}}{2016}]%
        {jiang2016variational}
\bibfield{author}{\bibinfo{person}{Zhuxi Jiang}, \bibinfo{person}{Yin Zheng},
  \bibinfo{person}{Huachun Tan}, \bibinfo{person}{Bangsheng Tang}, {and}
  \bibinfo{person}{Hanning Zhou}.} \bibinfo{year}{2016}\natexlab{}.
\newblock \showarticletitle{Variational deep embedding: An unsupervised and
  generative approach to clustering}.
\newblock \bibinfo{journal}{\emph{arXiv preprint arXiv:1611.05148}}
  (\bibinfo{year}{2016}).
\newblock


\bibitem[\protect\citeauthoryear{Jin, Yu, Jiao, Pan, He, Wu, Yu, and Zhang}{Jin
  et~al\mbox{.}}{2021}]%
        {jin2021survey}
\bibfield{author}{\bibinfo{person}{Di Jin}, \bibinfo{person}{Zhizhi Yu},
  \bibinfo{person}{Pengfei Jiao}, \bibinfo{person}{Shirui Pan},
  \bibinfo{person}{Dongxiao He}, \bibinfo{person}{Jia Wu},
  \bibinfo{person}{Philip Yu}, {and} \bibinfo{person}{Weixiong Zhang}.}
  \bibinfo{year}{2021}\natexlab{}.
\newblock \showarticletitle{A survey of community detection approaches: From
  statistical modeling to deep learning}.
\newblock \bibinfo{journal}{\emph{IEEE Transactions on Knowledge and Data
  Engineering}} (\bibinfo{year}{2021}).
\newblock


\bibitem[\protect\citeauthoryear{Jing and Tian}{Jing and Tian}{2020}]%
        {jing2020self}
\bibfield{author}{\bibinfo{person}{Longlong Jing} {and} \bibinfo{person}{Yingli
  Tian}.} \bibinfo{year}{2020}\natexlab{}.
\newblock \showarticletitle{Self-supervised visual feature learning with deep
  neural networks: A survey}.
\newblock \bibinfo{journal}{\emph{IEEE transactions on pattern analysis and
  machine intelligence}} \bibinfo{volume}{43}, \bibinfo{number}{11}
  (\bibinfo{year}{2020}), \bibinfo{pages}{4037--4058}.
\newblock


\bibitem[\protect\citeauthoryear{Johnson}{Johnson}{1967}]%
        {johnson1967hierarchical}
\bibfield{author}{\bibinfo{person}{Stephen~C Johnson}.}
  \bibinfo{year}{1967}\natexlab{}.
\newblock \showarticletitle{Hierarchical clustering schemes}.
\newblock \bibinfo{journal}{\emph{Psychometrika}} \bibinfo{volume}{32},
  \bibinfo{number}{3} (\bibinfo{year}{1967}), \bibinfo{pages}{241--254}.
\newblock


\bibitem[\protect\citeauthoryear{Kalos and Whitlock}{Kalos and
  Whitlock}{2009}]%
        {kalos2009monte}
\bibfield{author}{\bibinfo{person}{Malvin~H Kalos} {and}
  \bibinfo{person}{Paula~A Whitlock}.} \bibinfo{year}{2009}\natexlab{}.
\newblock \bibinfo{booktitle}{\emph{Monte carlo methods}}.
\newblock \bibinfo{publisher}{John Wiley \& Sons}.
\newblock


\bibitem[\protect\citeauthoryear{K{\'a}roly, Full{\'e}r, and
  Galambos}{K{\'a}roly et~al\mbox{.}}{2018}]%
        {karoly2018unsupervised}
\bibfield{author}{\bibinfo{person}{Art{\'u}r~Istv{\'a}n K{\'a}roly},
  \bibinfo{person}{R{\'o}bert Full{\'e}r}, {and} \bibinfo{person}{P{\'e}ter
  Galambos}.} \bibinfo{year}{2018}\natexlab{}.
\newblock \showarticletitle{Unsupervised clustering for deep learning: A
  tutorial survey}.
\newblock \bibinfo{journal}{\emph{Acta Polytechnica Hungarica}}
  \bibinfo{volume}{15}, \bibinfo{number}{8} (\bibinfo{year}{2018}),
  \bibinfo{pages}{29--53}.
\newblock


\bibitem[\protect\citeauthoryear{Kart, Bai, Glocker, and Rueckert}{Kart
  et~al\mbox{.}}{2021}]%
        {kart2021deepmcat}
\bibfield{author}{\bibinfo{person}{Turkay Kart}, \bibinfo{person}{Wenjia Bai},
  \bibinfo{person}{Ben Glocker}, {and} \bibinfo{person}{Daniel Rueckert}.}
  \bibinfo{year}{2021}\natexlab{}.
\newblock \showarticletitle{DeepMCAT: Large-Scale Deep Clustering for Medical
  Image Categorization}.
\newblock In \bibinfo{booktitle}{\emph{Deep Generative Models, and Data
  Augmentation, Labelling, and Imperfections}}. \bibinfo{publisher}{Springer},
  \bibinfo{pages}{259--267}.
\newblock


\bibitem[\protect\citeauthoryear{Karypis and Kumar}{Karypis and Kumar}{1998}]%
        {karypis1998fast}
\bibfield{author}{\bibinfo{person}{George Karypis} {and} \bibinfo{person}{Vipin
  Kumar}.} \bibinfo{year}{1998}\natexlab{}.
\newblock \showarticletitle{A fast and high quality multilevel scheme for
  partitioning irregular graphs}.
\newblock \bibinfo{journal}{\emph{SIAM Journal on scientific Computing}}
  \bibinfo{volume}{20}, \bibinfo{number}{1} (\bibinfo{year}{1998}),
  \bibinfo{pages}{359--392}.
\newblock


\bibitem[\protect\citeauthoryear{Khosla, Teterwak, Wang, Sarna, Tian, Isola,
  Maschinot, Liu, and Krishnan}{Khosla et~al\mbox{.}}{2020}]%
        {khosla2020supervised}
\bibfield{author}{\bibinfo{person}{Prannay Khosla}, \bibinfo{person}{Piotr
  Teterwak}, \bibinfo{person}{Chen Wang}, \bibinfo{person}{Aaron Sarna},
  \bibinfo{person}{Yonglong Tian}, \bibinfo{person}{Phillip Isola},
  \bibinfo{person}{Aaron Maschinot}, \bibinfo{person}{Ce Liu}, {and}
  \bibinfo{person}{Dilip Krishnan}.} \bibinfo{year}{2020}\natexlab{}.
\newblock \showarticletitle{Supervised contrastive learning}.
\newblock \bibinfo{journal}{\emph{Advances in Neural Information Processing
  Systems}}  \bibinfo{volume}{33} (\bibinfo{year}{2020}),
  \bibinfo{pages}{18661--18673}.
\newblock


\bibitem[\protect\citeauthoryear{Kingma and Welling}{Kingma and
  Welling}{2013}]%
        {kingma2013auto}
\bibfield{author}{\bibinfo{person}{Diederik~P Kingma} {and}
  \bibinfo{person}{Max Welling}.} \bibinfo{year}{2013}\natexlab{}.
\newblock \showarticletitle{Auto-encoding variational bayes}.
\newblock \bibinfo{journal}{\emph{arXiv preprint arXiv:1312.6114}}
  (\bibinfo{year}{2013}).
\newblock


\bibitem[\protect\citeauthoryear{Kinney and Atwal}{Kinney and Atwal}{2014}]%
        {kinney2014equitability}
\bibfield{author}{\bibinfo{person}{Justin~B Kinney} {and}
  \bibinfo{person}{Gurinder~S Atwal}.} \bibinfo{year}{2014}\natexlab{}.
\newblock \showarticletitle{Equitability, mutual information, and the maximal
  information coefficient}.
\newblock \bibinfo{journal}{\emph{Proceedings of the National Academy of
  Sciences}} \bibinfo{volume}{111}, \bibinfo{number}{9} (\bibinfo{year}{2014}),
  \bibinfo{pages}{3354--3359}.
\newblock


\bibitem[\protect\citeauthoryear{Kipf and Welling}{Kipf and Welling}{2016}]%
        {kipf2016semi}
\bibfield{author}{\bibinfo{person}{Thomas~N Kipf} {and} \bibinfo{person}{Max
  Welling}.} \bibinfo{year}{2016}\natexlab{}.
\newblock \showarticletitle{Semi-supervised classification with graph
  convolutional networks}.
\newblock \bibinfo{journal}{\emph{arXiv preprint arXiv:1609.02907}}
  (\bibinfo{year}{2016}).
\newblock


\bibitem[\protect\citeauthoryear{Kong, Bendersky, Najork, Vargo, and
  Colagrosso}{Kong et~al\mbox{.}}{2020}]%
        {kong2020learning}
\bibfield{author}{\bibinfo{person}{Weize Kong}, \bibinfo{person}{Michael
  Bendersky}, \bibinfo{person}{Marc Najork}, \bibinfo{person}{Brandon Vargo},
  {and} \bibinfo{person}{Mike Colagrosso}.} \bibinfo{year}{2020}\natexlab{}.
\newblock \showarticletitle{Learning to cluster documents into workspaces using
  large scale activity logs}. In \bibinfo{booktitle}{\emph{Proceedings of the
  26th ACM SIGKDD International Conference on Knowledge Discovery \& Data
  Mining}}. \bibinfo{pages}{2416--2424}.
\newblock


\bibitem[\protect\citeauthoryear{Kraskov, St{\"o}gbauer, and
  Grassberger}{Kraskov et~al\mbox{.}}{2004}]%
        {kraskov2004estimating}
\bibfield{author}{\bibinfo{person}{Alexander Kraskov}, \bibinfo{person}{Harald
  St{\"o}gbauer}, {and} \bibinfo{person}{Peter Grassberger}.}
  \bibinfo{year}{2004}\natexlab{}.
\newblock \showarticletitle{Estimating mutual information}.
\newblock \bibinfo{journal}{\emph{Physical review E}} \bibinfo{volume}{69},
  \bibinfo{number}{6} (\bibinfo{year}{2004}), \bibinfo{pages}{066138}.
\newblock


\bibitem[\protect\citeauthoryear{Kuhn}{Kuhn}{1955}]%
        {kuhn1955hungarian}
\bibfield{author}{\bibinfo{person}{Harold~W Kuhn}.}
  \bibinfo{year}{1955}\natexlab{}.
\newblock \showarticletitle{The Hungarian method for the assignment problem}.
\newblock \bibinfo{journal}{\emph{Naval research logistics quarterly}}
  \bibinfo{volume}{2}, \bibinfo{number}{1-2} (\bibinfo{year}{1955}),
  \bibinfo{pages}{83--97}.
\newblock


\bibitem[\protect\citeauthoryear{Le and Mikolov}{Le and Mikolov}{2014}]%
        {le2014distributed}
\bibfield{author}{\bibinfo{person}{Quoc Le} {and} \bibinfo{person}{Tomas
  Mikolov}.} \bibinfo{year}{2014}\natexlab{}.
\newblock \showarticletitle{Distributed representations of sentences and
  documents}. In \bibinfo{booktitle}{\emph{International conference on machine
  learning}}. PMLR, \bibinfo{pages}{1188--1196}.
\newblock


\bibitem[\protect\citeauthoryear{Le-Khac, Healy, and Smeaton}{Le-Khac
  et~al\mbox{.}}{2020}]%
        {le2020contrastive}
\bibfield{author}{\bibinfo{person}{Phuc~H Le-Khac}, \bibinfo{person}{Graham
  Healy}, {and} \bibinfo{person}{Alan~F Smeaton}.}
  \bibinfo{year}{2020}\natexlab{}.
\newblock \showarticletitle{Contrastive representation learning: A framework
  and review}.
\newblock \bibinfo{journal}{\emph{IEEE Access}}  \bibinfo{volume}{8}
  (\bibinfo{year}{2020}), \bibinfo{pages}{193907--193934}.
\newblock


\bibitem[\protect\citeauthoryear{LeCun, Bottou, Bengio, and Haffner}{LeCun
  et~al\mbox{.}}{1998}]%
        {lecun1998gradient}
\bibfield{author}{\bibinfo{person}{Yann LeCun}, \bibinfo{person}{L{\'e}on
  Bottou}, \bibinfo{person}{Yoshua Bengio}, {and} \bibinfo{person}{Patrick
  Haffner}.} \bibinfo{year}{1998}\natexlab{}.
\newblock \showarticletitle{Gradient-based learning applied to document
  recognition}.
\newblock \bibinfo{journal}{\emph{Proc. IEEE}} \bibinfo{volume}{86},
  \bibinfo{number}{11} (\bibinfo{year}{1998}), \bibinfo{pages}{2278--2324}.
\newblock


\bibitem[\protect\citeauthoryear{Lee et~al\mbox{.}}{Lee et~al\mbox{.}}{2013}]%
        {lee2013pseudo}
\bibfield{author}{\bibinfo{person}{Dong-Hyun Lee} {et~al\mbox{.}}}
  \bibinfo{year}{2013}\natexlab{}.
\newblock \showarticletitle{Pseudo-label: The simple and efficient
  semi-supervised learning method for deep neural networks}. In
  \bibinfo{booktitle}{\emph{Workshop on challenges in representation learning,
  ICML}}, Vol.~\bibinfo{volume}{3}. \bibinfo{pages}{896}.
\newblock


\bibitem[\protect\citeauthoryear{Li, Qiao, and Zhang}{Li et~al\mbox{.}}{2018}]%
        {li2018discriminatively}
\bibfield{author}{\bibinfo{person}{Fengfu Li}, \bibinfo{person}{Hong Qiao},
  {and} \bibinfo{person}{Bo Zhang}.} \bibinfo{year}{2018}\natexlab{}.
\newblock \showarticletitle{Discriminatively boosted image clustering with
  fully convolutional auto-encoders}.
\newblock \bibinfo{journal}{\emph{Pattern Recognition}}  \bibinfo{volume}{83}
  (\bibinfo{year}{2018}), \bibinfo{pages}{161--173}.
\newblock


\bibitem[\protect\citeauthoryear{Li, Wang, Zhang, Yuan, Li, and Zhu}{Li
  et~al\mbox{.}}{2021b}]%
        {li2021disentangled}
\bibfield{author}{\bibinfo{person}{Haoyang Li}, \bibinfo{person}{Xin Wang},
  \bibinfo{person}{Ziwei Zhang}, \bibinfo{person}{Zehuan Yuan},
  \bibinfo{person}{Hang Li}, {and} \bibinfo{person}{Wenwu Zhu}.}
  \bibinfo{year}{2021}\natexlab{b}.
\newblock \showarticletitle{Disentangled Contrastive Learning on Graphs}.
\newblock \bibinfo{journal}{\emph{Advances in Neural Information Processing
  Systems}}  \bibinfo{volume}{34} (\bibinfo{year}{2021}).
\newblock


\bibitem[\protect\citeauthoryear{Li, Zhou, Xiong, and Hoi}{Li
  et~al\mbox{.}}{2020d}]%
        {li2020prototypical}
\bibfield{author}{\bibinfo{person}{Junnan Li}, \bibinfo{person}{Pan Zhou},
  \bibinfo{person}{Caiming Xiong}, {and} \bibinfo{person}{Steven~CH Hoi}.}
  \bibinfo{year}{2020}\natexlab{d}.
\newblock \showarticletitle{Prototypical contrastive learning of unsupervised
  representations}.
\newblock \bibinfo{journal}{\emph{arXiv preprint arXiv:2005.04966}}
  (\bibinfo{year}{2020}).
\newblock


\bibitem[\protect\citeauthoryear{Li, Wu, Ester, Kao, Wang, and Zheng}{Li
  et~al\mbox{.}}{2017}]%
        {li2017semi}
\bibfield{author}{\bibinfo{person}{Xiang Li}, \bibinfo{person}{Yao Wu},
  \bibinfo{person}{Martin Ester}, \bibinfo{person}{Ben Kao},
  \bibinfo{person}{Xin Wang}, {and} \bibinfo{person}{Yudian Zheng}.}
  \bibinfo{year}{2017}\natexlab{}.
\newblock \showarticletitle{Semi-supervised clustering in attributed
  heterogeneous information networks}. In \bibinfo{booktitle}{\emph{Proceedings
  of the 26th international conference on world wide web}}.
  \bibinfo{pages}{1621--1629}.
\newblock


\bibitem[\protect\citeauthoryear{Li, Wu, Ester, Kao, Wang, and Zheng}{Li
  et~al\mbox{.}}{2020b}]%
        {li2020schain}
\bibfield{author}{\bibinfo{person}{Xiang Li}, \bibinfo{person}{Yao Wu},
  \bibinfo{person}{Martin Ester}, \bibinfo{person}{Ben Kao},
  \bibinfo{person}{Xin Wang}, {and} \bibinfo{person}{Yudian Zheng}.}
  \bibinfo{year}{2020}\natexlab{b}.
\newblock \showarticletitle{Schain-iram: An efficient and effective
  semi-supervised clustering algorithm for attributed heterogeneous information
  networks}.
\newblock \bibinfo{journal}{\emph{IEEE Transactions on knowledge and data
  engineering}} (\bibinfo{year}{2020}).
\newblock


\bibitem[\protect\citeauthoryear{Li, Cai, and Wang}{Li et~al\mbox{.}}{2020a}]%
        {li2020text}
\bibfield{author}{\bibinfo{person}{Yutong Li}, \bibinfo{person}{Juanjuan Cai},
  {and} \bibinfo{person}{Jingling Wang}.} \bibinfo{year}{2020}\natexlab{a}.
\newblock \showarticletitle{A text document clustering method based on weighted
  Bert model}. In \bibinfo{booktitle}{\emph{2020 IEEE 4th Information
  Technology, Networking, Electronic and Automation Control Conference
  (ITNEC)}}, Vol.~\bibinfo{volume}{1}. IEEE, \bibinfo{pages}{1426--1430}.
\newblock


\bibitem[\protect\citeauthoryear{Li, Hu, Liu, Peng, Zhou, and Peng}{Li
  et~al\mbox{.}}{2021a}]%
        {li2021contrastive}
\bibfield{author}{\bibinfo{person}{Yunfan Li}, \bibinfo{person}{Peng Hu},
  \bibinfo{person}{Zitao Liu}, \bibinfo{person}{Dezhong Peng},
  \bibinfo{person}{Joey~Tianyi Zhou}, {and} \bibinfo{person}{Xi Peng}.}
  \bibinfo{year}{2021}\natexlab{a}.
\newblock \showarticletitle{Contrastive clustering}. In
  \bibinfo{booktitle}{\emph{2021 AAAI Conference on Artificial Intelligence
  (AAAI)}}.
\newblock


\bibitem[\protect\citeauthoryear{Li, Zhao, Xu, Chen, Xu, Li, and Pei}{Li
  et~al\mbox{.}}{2020c}]%
        {li2020unsupervised}
\bibfield{author}{\bibinfo{person}{Zhihan Li}, \bibinfo{person}{Youjian Zhao},
  \bibinfo{person}{Haowen Xu}, \bibinfo{person}{Wenxiao Chen},
  \bibinfo{person}{Shangqing Xu}, \bibinfo{person}{Yilin Li}, {and}
  \bibinfo{person}{Dan Pei}.} \bibinfo{year}{2020}\natexlab{c}.
\newblock \showarticletitle{Unsupervised clustering through gaussian mixture
  variational autoencoder with non-reparameterized variational inference and
  std annealing}. In \bibinfo{booktitle}{\emph{2020 International Joint
  Conference on Neural Networks (IJCNN)}}. IEEE, \bibinfo{pages}{1--8}.
\newblock


\bibitem[\protect\citeauthoryear{Liang, Jiang, Li, Xue, and Wang}{Liang
  et~al\mbox{.}}{2020}]%
        {liang2020lr}
\bibfield{author}{\bibinfo{person}{XW Liang}, \bibinfo{person}{AP Jiang},
  \bibinfo{person}{T Li}, \bibinfo{person}{YY Xue}, {and} \bibinfo{person}{GT
  Wang}.} \bibinfo{year}{2020}\natexlab{}.
\newblock \showarticletitle{LR-SMOTE—An improved unbalanced data set
  oversampling based on K-means and SVM}.
\newblock \bibinfo{journal}{\emph{Knowledge-Based Systems}}
  \bibinfo{volume}{196} (\bibinfo{year}{2020}), \bibinfo{pages}{105845}.
\newblock


\bibitem[\protect\citeauthoryear{Liu, Xue, Wu, Zhou, Hu, Paris, Nepal, Yang,
  and Yu}{Liu et~al\mbox{.}}{2020}]%
        {liu2020deep}
\bibfield{author}{\bibinfo{person}{Fanzhen Liu}, \bibinfo{person}{Shan Xue},
  \bibinfo{person}{Jia Wu}, \bibinfo{person}{Chuan Zhou},
  \bibinfo{person}{Wenbin Hu}, \bibinfo{person}{Cecile Paris},
  \bibinfo{person}{Surya Nepal}, \bibinfo{person}{Jian Yang}, {and}
  \bibinfo{person}{Philip~S Yu}.} \bibinfo{year}{2020}\natexlab{}.
\newblock \showarticletitle{Deep learning for community detection: progress,
  challenges and opportunities}.
\newblock \bibinfo{journal}{\emph{arXiv preprint arXiv:2005.08225}}
  (\bibinfo{year}{2020}).
\newblock


\bibitem[\protect\citeauthoryear{Liu, Li, Wu, and Fu}{Liu
  et~al\mbox{.}}{2019}]%
        {liu2019clustering}
\bibfield{author}{\bibinfo{person}{Hongfu Liu}, \bibinfo{person}{Jun Li},
  \bibinfo{person}{Yue Wu}, {and} \bibinfo{person}{Yun Fu}.}
  \bibinfo{year}{2019}\natexlab{}.
\newblock \showarticletitle{Clustering with outlier removal}.
\newblock \bibinfo{journal}{\emph{IEEE transactions on knowledge and data
  engineering}} \bibinfo{volume}{33}, \bibinfo{number}{6}
  (\bibinfo{year}{2019}), \bibinfo{pages}{2369--2379}.
\newblock


\bibitem[\protect\citeauthoryear{Liu, Wang, Shen, and Tsang}{Liu
  et~al\mbox{.}}{2021a}]%
        {liu2021emerging}
\bibfield{author}{\bibinfo{person}{Weiwei Liu}, \bibinfo{person}{Haobo Wang},
  \bibinfo{person}{Xiaobo Shen}, {and} \bibinfo{person}{Ivor Tsang}.}
  \bibinfo{year}{2021}\natexlab{a}.
\newblock \showarticletitle{The emerging trends of multi-label learning}.
\newblock \bibinfo{journal}{\emph{IEEE transactions on pattern analysis and
  machine intelligence}} (\bibinfo{year}{2021}).
\newblock


\bibitem[\protect\citeauthoryear{Liu, Zhang, Hou, Mian, Wang, Zhang, and
  Tang}{Liu et~al\mbox{.}}{2021b}]%
        {liu2021self}
\bibfield{author}{\bibinfo{person}{Xiao Liu}, \bibinfo{person}{Fanjin Zhang},
  \bibinfo{person}{Zhenyu Hou}, \bibinfo{person}{Li Mian},
  \bibinfo{person}{Zhaoyu Wang}, \bibinfo{person}{Jing Zhang}, {and}
  \bibinfo{person}{Jie Tang}.} \bibinfo{year}{2021}\natexlab{b}.
\newblock \showarticletitle{Self-supervised learning: Generative or
  contrastive}.
\newblock \bibinfo{journal}{\emph{IEEE Transactions on Knowledge and Data
  Engineering}} (\bibinfo{year}{2021}).
\newblock


\bibitem[\protect\citeauthoryear{Lloyd}{Lloyd}{1982}]%
        {lloyd1982least}
\bibfield{author}{\bibinfo{person}{Stuart Lloyd}.}
  \bibinfo{year}{1982}\natexlab{}.
\newblock \showarticletitle{Least squares quantization in PCM}.
\newblock \bibinfo{journal}{\emph{IEEE transactions on information theory}}
  \bibinfo{volume}{28}, \bibinfo{number}{2} (\bibinfo{year}{1982}),
  \bibinfo{pages}{129--137}.
\newblock


\bibitem[\protect\citeauthoryear{Long, Cao, Wang, and Jordan}{Long
  et~al\mbox{.}}{2015}]%
        {long2015learning}
\bibfield{author}{\bibinfo{person}{Mingsheng Long}, \bibinfo{person}{Yue Cao},
  \bibinfo{person}{Jianmin Wang}, {and} \bibinfo{person}{Michael Jordan}.}
  \bibinfo{year}{2015}\natexlab{}.
\newblock \showarticletitle{Learning transferable features with deep adaptation
  networks}. In \bibinfo{booktitle}{\emph{International conference on machine
  learning}}. PMLR, \bibinfo{pages}{97--105}.
\newblock


\bibitem[\protect\citeauthoryear{Lv, Kang, Lu, and Xu}{Lv
  et~al\mbox{.}}{2021}]%
        {lv2021pseudo}
\bibfield{author}{\bibinfo{person}{Juncheng Lv}, \bibinfo{person}{Zhao Kang},
  \bibinfo{person}{Xiao Lu}, {and} \bibinfo{person}{Zenglin Xu}.}
  \bibinfo{year}{2021}\natexlab{}.
\newblock \showarticletitle{Pseudo-supervised deep subspace clustering}.
\newblock \bibinfo{journal}{\emph{IEEE Transactions on Image Processing}}
  \bibinfo{volume}{30} (\bibinfo{year}{2021}), \bibinfo{pages}{5252--5263}.
\newblock


\bibitem[\protect\citeauthoryear{Ma, Cui, Kuang, Wang, and Zhu}{Ma
  et~al\mbox{.}}{2019}]%
        {ma2019disentangled}
\bibfield{author}{\bibinfo{person}{Jianxin Ma}, \bibinfo{person}{Peng Cui},
  \bibinfo{person}{Kun Kuang}, \bibinfo{person}{Xin Wang}, {and}
  \bibinfo{person}{Wenwu Zhu}.} \bibinfo{year}{2019}\natexlab{}.
\newblock \showarticletitle{Disentangled graph convolutional networks}. In
  \bibinfo{booktitle}{\emph{International conference on machine learning}}.
  PMLR, \bibinfo{pages}{4212--4221}.
\newblock


\bibitem[\protect\citeauthoryear{Ma, Wu, Xue, Yang, Zhou, Sheng, Xiong, and
  Akoglu}{Ma et~al\mbox{.}}{2021}]%
        {ma2021comprehensive}
\bibfield{author}{\bibinfo{person}{Xiaoxiao Ma}, \bibinfo{person}{Jia Wu},
  \bibinfo{person}{Shan Xue}, \bibinfo{person}{Jian Yang},
  \bibinfo{person}{Chuan Zhou}, \bibinfo{person}{Quan~Z Sheng},
  \bibinfo{person}{Hui Xiong}, {and} \bibinfo{person}{Leman Akoglu}.}
  \bibinfo{year}{2021}\natexlab{}.
\newblock \showarticletitle{A comprehensive survey on graph anomaly detection
  with deep learning}.
\newblock \bibinfo{journal}{\emph{IEEE Transactions on Knowledge and Data
  Engineering}} (\bibinfo{year}{2021}).
\newblock


\bibitem[\protect\citeauthoryear{Markovitz, Sharir, Friedman, Zelnik-Manor, and
  Avidan}{Markovitz et~al\mbox{.}}{2020}]%
        {markovitz2020graph}
\bibfield{author}{\bibinfo{person}{Amir Markovitz}, \bibinfo{person}{Gilad
  Sharir}, \bibinfo{person}{Itamar Friedman}, \bibinfo{person}{Lihi
  Zelnik-Manor}, {and} \bibinfo{person}{Shai Avidan}.}
  \bibinfo{year}{2020}\natexlab{}.
\newblock \showarticletitle{Graph embedded pose clustering for anomaly
  detection}. In \bibinfo{booktitle}{\emph{Proceedings of the IEEE/CVF
  Conference on Computer Vision and Pattern Recognition}}.
  \bibinfo{pages}{10539--10547}.
\newblock


\bibitem[\protect\citeauthoryear{Meier, Van De~Geer, and B{\"u}hlmann}{Meier
  et~al\mbox{.}}{2008}]%
        {meier2008group}
\bibfield{author}{\bibinfo{person}{Lukas Meier}, \bibinfo{person}{Sara Van
  De~Geer}, {and} \bibinfo{person}{Peter B{\"u}hlmann}.}
  \bibinfo{year}{2008}\natexlab{}.
\newblock \showarticletitle{The group lasso for logistic regression}.
\newblock \bibinfo{journal}{\emph{Journal of the Royal Statistical Society:
  Series B (Statistical Methodology)}} \bibinfo{volume}{70},
  \bibinfo{number}{1} (\bibinfo{year}{2008}), \bibinfo{pages}{53--71}.
\newblock


\bibitem[\protect\citeauthoryear{Menapace, Lathuili{\`e}re, and Ricci}{Menapace
  et~al\mbox{.}}{2020}]%
        {menapace2020learning}
\bibfield{author}{\bibinfo{person}{Willi Menapace},
  \bibinfo{person}{St{\'e}phane Lathuili{\`e}re}, {and} \bibinfo{person}{Elisa
  Ricci}.} \bibinfo{year}{2020}\natexlab{}.
\newblock \showarticletitle{Learning to cluster under domain shift}. In
  \bibinfo{booktitle}{\emph{European Conference on Computer Vision}}. Springer,
  \bibinfo{pages}{736--752}.
\newblock


\bibitem[\protect\citeauthoryear{Meng, Zhang, Huang, Zhang, Zhang, and
  Han}{Meng et~al\mbox{.}}{2020}]%
        {meng2020hierarchical}
\bibfield{author}{\bibinfo{person}{Yu Meng}, \bibinfo{person}{Yunyi Zhang},
  \bibinfo{person}{Jiaxin Huang}, \bibinfo{person}{Yu Zhang},
  \bibinfo{person}{Chao Zhang}, {and} \bibinfo{person}{Jiawei Han}.}
  \bibinfo{year}{2020}\natexlab{}.
\newblock \showarticletitle{Hierarchical topic mining via joint spherical tree
  and text embedding}. In \bibinfo{booktitle}{\emph{Proceedings of the 26th ACM
  SIGKDD international conference on knowledge discovery \& data mining}}.
  \bibinfo{pages}{1908--1917}.
\newblock


\bibitem[\protect\citeauthoryear{Mikolov, Chen, Corrado, and Dean}{Mikolov
  et~al\mbox{.}}{2013}]%
        {mikolov2013efficient}
\bibfield{author}{\bibinfo{person}{Tomas Mikolov}, \bibinfo{person}{Kai Chen},
  \bibinfo{person}{Greg Corrado}, {and} \bibinfo{person}{Jeffrey Dean}.}
  \bibinfo{year}{2013}\natexlab{}.
\newblock \showarticletitle{Efficient estimation of word representations in
  vector space}.
\newblock \bibinfo{journal}{\emph{arXiv preprint arXiv:1301.3781}}
  (\bibinfo{year}{2013}).
\newblock


\bibitem[\protect\citeauthoryear{Min, Guo, Liu, Zhang, Cui, and Long}{Min
  et~al\mbox{.}}{2018}]%
        {min2018survey}
\bibfield{author}{\bibinfo{person}{Erxue Min}, \bibinfo{person}{Xifeng Guo},
  \bibinfo{person}{Qiang Liu}, \bibinfo{person}{Gen Zhang},
  \bibinfo{person}{Jianjing Cui}, {and} \bibinfo{person}{Jun Long}.}
  \bibinfo{year}{2018}\natexlab{}.
\newblock \showarticletitle{A survey of clustering with deep learning: From the
  perspective of network architecture}.
\newblock \bibinfo{journal}{\emph{IEEE Access}}  \bibinfo{volume}{6}
  (\bibinfo{year}{2018}), \bibinfo{pages}{39501--39514}.
\newblock


\bibitem[\protect\citeauthoryear{Mittal, Pandey, Pal, and Tripathi}{Mittal
  et~al\mbox{.}}{2021}]%
        {mittal2021new}
\bibfield{author}{\bibinfo{person}{Himanshu Mittal},
  \bibinfo{person}{Avinash~Chandra Pandey}, \bibinfo{person}{Raju Pal}, {and}
  \bibinfo{person}{Ashish Tripathi}.} \bibinfo{year}{2021}\natexlab{}.
\newblock \showarticletitle{A new clustering method for the diagnosis of
  CoVID19 using medical images}.
\newblock \bibinfo{journal}{\emph{Applied Intelligence}} \bibinfo{volume}{51},
  \bibinfo{number}{5} (\bibinfo{year}{2021}), \bibinfo{pages}{2988--3011}.
\newblock


\bibitem[\protect\citeauthoryear{Moser, Ge, and Ester}{Moser
  et~al\mbox{.}}{2007}]%
        {moser2007joint}
\bibfield{author}{\bibinfo{person}{Flavia Moser}, \bibinfo{person}{Rong Ge},
  {and} \bibinfo{person}{Martin Ester}.} \bibinfo{year}{2007}\natexlab{}.
\newblock \showarticletitle{Joint cluster analysis of attribute and
  relationship data withouta-priori specification of the number of clusters}.
  In \bibinfo{booktitle}{\emph{Proceedings of the 13th ACM SIGKDD international
  conference on Knowledge discovery and data mining}}.
  \bibinfo{pages}{510--519}.
\newblock


\bibitem[\protect\citeauthoryear{Mukherjee, Asnani, Lin, and Kannan}{Mukherjee
  et~al\mbox{.}}{2019}]%
        {mukherjee2019clustergan}
\bibfield{author}{\bibinfo{person}{Sudipto Mukherjee},
  \bibinfo{person}{Himanshu Asnani}, \bibinfo{person}{Eugene Lin}, {and}
  \bibinfo{person}{Sreeram Kannan}.} \bibinfo{year}{2019}\natexlab{}.
\newblock \showarticletitle{Clustergan: Latent space clustering in generative
  adversarial networks}. In \bibinfo{booktitle}{\emph{Proceedings of the AAAI
  Conference on Artificial Intelligence}}, Vol.~\bibinfo{volume}{33}.
  \bibinfo{pages}{4610--4617}.
\newblock


\bibitem[\protect\citeauthoryear{Nasrazadani, Fatemi, and
  Nematbakhsh}{Nasrazadani et~al\mbox{.}}{2022}]%
        {nasrazadani2022sign}
\bibfield{author}{\bibinfo{person}{Mina Nasrazadani}, \bibinfo{person}{Afsaneh
  Fatemi}, {and} \bibinfo{person}{Mohammadali Nematbakhsh}.}
  \bibinfo{year}{2022}\natexlab{}.
\newblock \showarticletitle{Sign prediction in sparse social networks using
  clustering and collaborative filtering}.
\newblock \bibinfo{journal}{\emph{The Journal of Supercomputing}}
  \bibinfo{volume}{78}, \bibinfo{number}{1} (\bibinfo{year}{2022}),
  \bibinfo{pages}{596--615}.
\newblock


\bibitem[\protect\citeauthoryear{Ng, Jordan, and Weiss}{Ng
  et~al\mbox{.}}{2001}]%
        {ng2001spectral}
\bibfield{author}{\bibinfo{person}{Andrew Ng}, \bibinfo{person}{Michael
  Jordan}, {and} \bibinfo{person}{Yair Weiss}.}
  \bibinfo{year}{2001}\natexlab{}.
\newblock \showarticletitle{On spectral clustering: Analysis and an algorithm}.
\newblock \bibinfo{journal}{\emph{Advances in neural information processing
  systems}}  \bibinfo{volume}{14} (\bibinfo{year}{2001}).
\newblock


\bibitem[\protect\citeauthoryear{Niu, Shan, and Wang}{Niu
  et~al\mbox{.}}{2021}]%
        {niu2021spice}
\bibfield{author}{\bibinfo{person}{Chuang Niu}, \bibinfo{person}{Hongming
  Shan}, {and} \bibinfo{person}{Ge Wang}.} \bibinfo{year}{2021}\natexlab{}.
\newblock \showarticletitle{Spice: Semantic pseudo-labeling for image
  clustering}.
\newblock \bibinfo{journal}{\emph{arXiv preprint arXiv:2103.09382}}
  (\bibinfo{year}{2021}).
\newblock


\bibitem[\protect\citeauthoryear{Niu, Zhang, Wang, and Liang}{Niu
  et~al\mbox{.}}{2020}]%
        {niu2020gatcluster}
\bibfield{author}{\bibinfo{person}{Chuang Niu}, \bibinfo{person}{Jun Zhang},
  \bibinfo{person}{Ge Wang}, {and} \bibinfo{person}{Jimin Liang}.}
  \bibinfo{year}{2020}\natexlab{}.
\newblock \showarticletitle{Gatcluster: Self-supervised gaussian-attention
  network for image clustering}. In \bibinfo{booktitle}{\emph{European
  Conference on Computer Vision}}. Springer, \bibinfo{pages}{735--751}.
\newblock


\bibitem[\protect\citeauthoryear{Nowozin, Cseke, and Tomioka}{Nowozin
  et~al\mbox{.}}{2016}]%
        {nowozin2016f}
\bibfield{author}{\bibinfo{person}{Sebastian Nowozin}, \bibinfo{person}{Botond
  Cseke}, {and} \bibinfo{person}{Ryota Tomioka}.}
  \bibinfo{year}{2016}\natexlab{}.
\newblock \showarticletitle{f-gan: Training generative neural samplers using
  variational divergence minimization}.
\newblock \bibinfo{journal}{\emph{Advances in neural information processing
  systems}}  \bibinfo{volume}{29} (\bibinfo{year}{2016}).
\newblock


\bibitem[\protect\citeauthoryear{Ntelemis, Jin, and Thomas}{Ntelemis
  et~al\mbox{.}}{2021a}]%
        {ntelemis2021image}
\bibfield{author}{\bibinfo{person}{Foivos Ntelemis}, \bibinfo{person}{Yaochu
  Jin}, {and} \bibinfo{person}{Spencer~A Thomas}.}
  \bibinfo{year}{2021}\natexlab{a}.
\newblock \showarticletitle{Image clustering using an augmented generative
  adversarial network and information maximization}.
\newblock \bibinfo{journal}{\emph{IEEE Transactions on Neural Networks and
  Learning Systems}} (\bibinfo{year}{2021}).
\newblock


\bibitem[\protect\citeauthoryear{Ntelemis, Jin, and Thomas}{Ntelemis
  et~al\mbox{.}}{2021b}]%
        {ntelemis2021information}
\bibfield{author}{\bibinfo{person}{Foivos Ntelemis}, \bibinfo{person}{Yaochu
  Jin}, {and} \bibinfo{person}{Spencer~A Thomas}.}
  \bibinfo{year}{2021}\natexlab{b}.
\newblock \showarticletitle{Information Maximization Clustering via Multi-View
  Self-Labelling}.
\newblock \bibinfo{journal}{\emph{arXiv preprint arXiv:2103.07368}}
  (\bibinfo{year}{2021}).
\newblock


\bibitem[\protect\citeauthoryear{Nutakki, Abdollahi, Sun, and Nasraoui}{Nutakki
  et~al\mbox{.}}{2019}]%
        {nutakki2019introduction}
\bibfield{author}{\bibinfo{person}{Gopi~Chand Nutakki},
  \bibinfo{person}{Behnoush Abdollahi}, \bibinfo{person}{Wenlong Sun}, {and}
  \bibinfo{person}{Olfa Nasraoui}.} \bibinfo{year}{2019}\natexlab{}.
\newblock \showarticletitle{An introduction to deep clustering}.
\newblock In \bibinfo{booktitle}{\emph{Clustering Methods for Big Data
  Analytics}}. \bibinfo{publisher}{Springer}, \bibinfo{pages}{73--89}.
\newblock


\bibitem[\protect\citeauthoryear{Olive and Morio}{Olive and Morio}{2019}]%
        {olive2019trajectory}
\bibfield{author}{\bibinfo{person}{Xavier Olive} {and}
  \bibinfo{person}{J{\'e}r{\^o}me Morio}.} \bibinfo{year}{2019}\natexlab{}.
\newblock \showarticletitle{Trajectory clustering of air traffic flows around
  airports}.
\newblock \bibinfo{journal}{\emph{Aerospace Science and Technology}}
  \bibinfo{volume}{84} (\bibinfo{year}{2019}), \bibinfo{pages}{776--781}.
\newblock


\bibitem[\protect\citeauthoryear{Oord, Li, and Vinyals}{Oord
  et~al\mbox{.}}{2018}]%
        {oord2018representation}
\bibfield{author}{\bibinfo{person}{Aaron van~den Oord}, \bibinfo{person}{Yazhe
  Li}, {and} \bibinfo{person}{Oriol Vinyals}.} \bibinfo{year}{2018}\natexlab{}.
\newblock \showarticletitle{Representation learning with contrastive predictive
  coding}.
\newblock \bibinfo{journal}{\emph{arXiv preprint arXiv:1807.03748}}
  (\bibinfo{year}{2018}).
\newblock


\bibitem[\protect\citeauthoryear{Orkphol and Yang}{Orkphol and Yang}{2019}]%
        {orkphol2019sentiment}
\bibfield{author}{\bibinfo{person}{Korawit Orkphol} {and} \bibinfo{person}{Wu
  Yang}.} \bibinfo{year}{2019}\natexlab{}.
\newblock \showarticletitle{Sentiment analysis on microblogging with K-means
  clustering and artificial bee colony}.
\newblock \bibinfo{journal}{\emph{International Journal of Computational
  Intelligence and Applications}} \bibinfo{volume}{18}, \bibinfo{number}{03}
  (\bibinfo{year}{2019}), \bibinfo{pages}{1950017}.
\newblock


\bibitem[\protect\citeauthoryear{Pan, Hu, Long, Jiang, Yao, and Zhang}{Pan
  et~al\mbox{.}}{2018}]%
        {pan2018adversarially}
\bibfield{author}{\bibinfo{person}{Shirui Pan}, \bibinfo{person}{Ruiqi Hu},
  \bibinfo{person}{Guodong Long}, \bibinfo{person}{Jing Jiang},
  \bibinfo{person}{Lina Yao}, {and} \bibinfo{person}{Chengqi Zhang}.}
  \bibinfo{year}{2018}\natexlab{}.
\newblock \showarticletitle{Adversarially regularized graph autoencoder for
  graph embedding}.
\newblock \bibinfo{journal}{\emph{arXiv preprint arXiv:1802.04407}}
  (\bibinfo{year}{2018}).
\newblock


\bibitem[\protect\citeauthoryear{Pan and Yang}{Pan and Yang}{2009}]%
        {pan2009survey}
\bibfield{author}{\bibinfo{person}{Sinno~Jialin Pan} {and}
  \bibinfo{person}{Qiang Yang}.} \bibinfo{year}{2009}\natexlab{}.
\newblock \showarticletitle{A survey on transfer learning}.
\newblock \bibinfo{journal}{\emph{IEEE Transactions on knowledge and data
  engineering}} \bibinfo{volume}{22}, \bibinfo{number}{10}
  (\bibinfo{year}{2009}), \bibinfo{pages}{1345--1359}.
\newblock


\bibitem[\protect\citeauthoryear{Park and Jun}{Park and Jun}{2009}]%
        {park2009simple}
\bibfield{author}{\bibinfo{person}{Hae-Sang Park} {and}
  \bibinfo{person}{Chi-Hyuck Jun}.} \bibinfo{year}{2009}\natexlab{}.
\newblock \showarticletitle{A simple and fast algorithm for K-medoids
  clustering}.
\newblock \bibinfo{journal}{\emph{Expert systems with applications}}
  \bibinfo{volume}{36}, \bibinfo{number}{2} (\bibinfo{year}{2009}),
  \bibinfo{pages}{3336--3341}.
\newblock


\bibitem[\protect\citeauthoryear{Park, Han, Kim, Kim, Park, Hong, and Cha}{Park
  et~al\mbox{.}}{2021}]%
        {park2021improving}
\bibfield{author}{\bibinfo{person}{Sungwon Park}, \bibinfo{person}{Sungwon
  Han}, \bibinfo{person}{Sundong Kim}, \bibinfo{person}{Danu Kim},
  \bibinfo{person}{Sungkyu Park}, \bibinfo{person}{Seunghoon Hong}, {and}
  \bibinfo{person}{Meeyoung Cha}.} \bibinfo{year}{2021}\natexlab{}.
\newblock \showarticletitle{Improving unsupervised image clustering with robust
  learning}. In \bibinfo{booktitle}{\emph{Proceedings of the IEEE/CVF
  Conference on Computer Vision and Pattern Recognition}}.
  \bibinfo{pages}{12278--12287}.
\newblock


\bibitem[\protect\citeauthoryear{Pedregosa, Varoquaux, Gramfort, Michel,
  Thirion, Grisel, Blondel, Prettenhofer, Weiss, Dubourg, Vanderplas, Passos,
  Cournapeau, Brucher, Perrot, and Duchesnay}{Pedregosa et~al\mbox{.}}{2011}]%
        {scikit-learn}
\bibfield{author}{\bibinfo{person}{F. Pedregosa}, \bibinfo{person}{G.
  Varoquaux}, \bibinfo{person}{A. Gramfort}, \bibinfo{person}{V. Michel},
  \bibinfo{person}{B. Thirion}, \bibinfo{person}{O. Grisel},
  \bibinfo{person}{M. Blondel}, \bibinfo{person}{P. Prettenhofer},
  \bibinfo{person}{R. Weiss}, \bibinfo{person}{V. Dubourg}, \bibinfo{person}{J.
  Vanderplas}, \bibinfo{person}{A. Passos}, \bibinfo{person}{D. Cournapeau},
  \bibinfo{person}{M. Brucher}, \bibinfo{person}{M. Perrot}, {and}
  \bibinfo{person}{E. Duchesnay}.} \bibinfo{year}{2011}\natexlab{}.
\newblock \showarticletitle{Scikit-learn: Machine Learning in {P}ython}.
\newblock \bibinfo{journal}{\emph{Journal of Machine Learning Research}}
  \bibinfo{volume}{12} (\bibinfo{year}{2011}), \bibinfo{pages}{2825--2830}.
\newblock


\bibitem[\protect\citeauthoryear{Peng, Lei, Fu, Jia, Zhang, and Li}{Peng
  et~al\mbox{.}}{2021}]%
        {peng2021deep}
\bibfield{author}{\bibinfo{person}{Bo Peng}, \bibinfo{person}{Jianjun Lei},
  \bibinfo{person}{Huazhu Fu}, \bibinfo{person}{Yalong Jia},
  \bibinfo{person}{Zongqian Zhang}, {and} \bibinfo{person}{Yi Li}.}
  \bibinfo{year}{2021}\natexlab{}.
\newblock \showarticletitle{Deep video action clustering via spatio-temporal
  feature learning}.
\newblock \bibinfo{journal}{\emph{Neurocomputing}}  \bibinfo{volume}{456}
  (\bibinfo{year}{2021}), \bibinfo{pages}{519--527}.
\newblock


\bibitem[\protect\citeauthoryear{Peng, Feng, Lu, Yau, and Yi}{Peng
  et~al\mbox{.}}{2017}]%
        {peng2017cascade}
\bibfield{author}{\bibinfo{person}{Xi Peng}, \bibinfo{person}{Jiashi Feng},
  \bibinfo{person}{Jiwen Lu}, \bibinfo{person}{Wei-Yun Yau}, {and}
  \bibinfo{person}{Zhang Yi}.} \bibinfo{year}{2017}\natexlab{}.
\newblock \showarticletitle{Cascade subspace clustering}. In
  \bibinfo{booktitle}{\emph{Thirty-First AAAI conference on artificial
  intelligence}}.
\newblock


\bibitem[\protect\citeauthoryear{Peng, Xiao, Feng, Yau, and Yi}{Peng
  et~al\mbox{.}}{2016}]%
        {peng2016deep}
\bibfield{author}{\bibinfo{person}{Xi Peng}, \bibinfo{person}{Shijie Xiao},
  \bibinfo{person}{Jiashi Feng}, \bibinfo{person}{Wei-Yun Yau}, {and}
  \bibinfo{person}{Zhang Yi}.} \bibinfo{year}{2016}\natexlab{}.
\newblock \showarticletitle{Deep subspace clustering with sparsity prior.}. In
  \bibinfo{booktitle}{\emph{IJCAI}}. \bibinfo{pages}{1925--1931}.
\newblock


\bibitem[\protect\citeauthoryear{Prasad, Mohammed, and Noorullah}{Prasad
  et~al\mbox{.}}{2019}]%
        {prasad2019hybrid}
\bibfield{author}{\bibinfo{person}{K~Rajendra Prasad}, \bibinfo{person}{Moulana
  Mohammed}, {and} \bibinfo{person}{RM Noorullah}.}
  \bibinfo{year}{2019}\natexlab{}.
\newblock \showarticletitle{Hybrid topic cluster models for social healthcare
  data}.
\newblock \bibinfo{journal}{\emph{Int J Adv Comput Sci Appl}}
  \bibinfo{volume}{10}, \bibinfo{number}{11} (\bibinfo{year}{2019}),
  \bibinfo{pages}{490--506}.
\newblock


\bibitem[\protect\citeauthoryear{Prasad, Das, and Bhowmick}{Prasad
  et~al\mbox{.}}{2020}]%
        {prasad2020variational}
\bibfield{author}{\bibinfo{person}{Vignesh Prasad}, \bibinfo{person}{Dipanjan
  Das}, {and} \bibinfo{person}{Brojeshwar Bhowmick}.}
  \bibinfo{year}{2020}\natexlab{}.
\newblock \showarticletitle{Variational clustering: Leveraging variational
  autoencoders for image clustering}. In \bibinfo{booktitle}{\emph{2020
  International Joint Conference on Neural Networks (IJCNN)}}. IEEE,
  \bibinfo{pages}{1--10}.
\newblock


\bibitem[\protect\citeauthoryear{Qian, Meng, Gong, Yang, Wang, Belongie, and
  Cui}{Qian et~al\mbox{.}}{2021}]%
        {qian2021spatiotemporal}
\bibfield{author}{\bibinfo{person}{Rui Qian}, \bibinfo{person}{Tianjian Meng},
  \bibinfo{person}{Boqing Gong}, \bibinfo{person}{Ming-Hsuan Yang},
  \bibinfo{person}{Huisheng Wang}, \bibinfo{person}{Serge Belongie}, {and}
  \bibinfo{person}{Yin Cui}.} \bibinfo{year}{2021}\natexlab{}.
\newblock \showarticletitle{Spatiotemporal contrastive video representation
  learning}. In \bibinfo{booktitle}{\emph{Proceedings of the IEEE/CVF
  Conference on Computer Vision and Pattern Recognition}}.
  \bibinfo{pages}{6964--6974}.
\newblock


\bibitem[\protect\citeauthoryear{Qiu, Sun, Xu, Shao, Dai, and Huang}{Qiu
  et~al\mbox{.}}{2020}]%
        {qiu2020pre}
\bibfield{author}{\bibinfo{person}{Xipeng Qiu}, \bibinfo{person}{Tianxiang
  Sun}, \bibinfo{person}{Yige Xu}, \bibinfo{person}{Yunfan Shao},
  \bibinfo{person}{Ning Dai}, {and} \bibinfo{person}{Xuanjing Huang}.}
  \bibinfo{year}{2020}\natexlab{}.
\newblock \showarticletitle{Pre-trained models for natural language processing:
  A survey}.
\newblock \bibinfo{journal}{\emph{Science China Technological Sciences}}
  \bibinfo{volume}{63}, \bibinfo{number}{10} (\bibinfo{year}{2020}),
  \bibinfo{pages}{1872--1897}.
\newblock


\bibitem[\protect\citeauthoryear{Rasmussen}{Rasmussen}{1999}]%
        {rasmussen1999infinite}
\bibfield{author}{\bibinfo{person}{Carl Rasmussen}.}
  \bibinfo{year}{1999}\natexlab{}.
\newblock \showarticletitle{The infinite Gaussian mixture model}.
\newblock \bibinfo{journal}{\emph{Advances in neural information processing
  systems}}  \bibinfo{volume}{12} (\bibinfo{year}{1999}).
\newblock


\bibitem[\protect\citeauthoryear{Reynolds}{Reynolds}{2009}]%
        {reynolds2009gaussian}
\bibfield{author}{\bibinfo{person}{Douglas~A Reynolds}.}
  \bibinfo{year}{2009}\natexlab{}.
\newblock \showarticletitle{Gaussian mixture models.}
\newblock \bibinfo{journal}{\emph{Encyclopedia of biometrics}}
  \bibinfo{volume}{741}, \bibinfo{number}{659-663} (\bibinfo{year}{2009}).
\newblock


\bibitem[\protect\citeauthoryear{Rezaei, Dorigatti, Ruegamer, and
  Bischl}{Rezaei et~al\mbox{.}}{2021}]%
        {rezaei2021learning}
\bibfield{author}{\bibinfo{person}{Mina Rezaei}, \bibinfo{person}{Emilio
  Dorigatti}, \bibinfo{person}{David Ruegamer}, {and} \bibinfo{person}{Bernd
  Bischl}.} \bibinfo{year}{2021}\natexlab{}.
\newblock \showarticletitle{Learning Statistical Representation with Joint Deep
  Embedded Clustering}.
\newblock \bibinfo{journal}{\emph{arXiv preprint arXiv:2109.05232}}
  (\bibinfo{year}{2021}).
\newblock


\bibitem[\protect\citeauthoryear{Rumelhart, Hinton, and Williams}{Rumelhart
  et~al\mbox{.}}{1985}]%
        {rumelhart1985learning}
\bibfield{author}{\bibinfo{person}{David~E Rumelhart},
  \bibinfo{person}{Geoffrey~E Hinton}, {and} \bibinfo{person}{Ronald~J
  Williams}.} \bibinfo{year}{1985}\natexlab{}.
\newblock \bibinfo{booktitle}{\emph{Learning internal representations by error
  propagation}}.
\newblock \bibinfo{type}{{T}echnical {R}eport}.
  \bibinfo{institution}{California Univ San Diego La Jolla Inst for Cognitive
  Science}.
\newblock


\bibitem[\protect\citeauthoryear{Saeedi~Emadi and Mazinani}{Saeedi~Emadi and
  Mazinani}{2018}]%
        {saeedi2018novel}
\bibfield{author}{\bibinfo{person}{Hossein Saeedi~Emadi} {and}
  \bibinfo{person}{Sayyed~Majid Mazinani}.} \bibinfo{year}{2018}\natexlab{}.
\newblock \showarticletitle{A novel anomaly detection algorithm using DBSCAN
  and SVM in wireless sensor networks}.
\newblock \bibinfo{journal}{\emph{Wireless Personal Communications}}
  \bibinfo{volume}{98}, \bibinfo{number}{2} (\bibinfo{year}{2018}),
  \bibinfo{pages}{2025--2035}.
\newblock


\bibitem[\protect\citeauthoryear{Sander, Ester, Kriegel, and Xu}{Sander
  et~al\mbox{.}}{1998}]%
        {sander1998density}
\bibfield{author}{\bibinfo{person}{J{\"o}rg Sander}, \bibinfo{person}{Martin
  Ester}, \bibinfo{person}{Hans-Peter Kriegel}, {and} \bibinfo{person}{Xiaowei
  Xu}.} \bibinfo{year}{1998}\natexlab{}.
\newblock \showarticletitle{Density-based clustering in spatial databases: The
  algorithm gdbscan and its applications}.
\newblock \bibinfo{journal}{\emph{Data mining and knowledge discovery}}
  \bibinfo{volume}{2}, \bibinfo{number}{2} (\bibinfo{year}{1998}),
  \bibinfo{pages}{169--194}.
\newblock


\bibitem[\protect\citeauthoryear{Schubert, Sander, Ester, Kriegel, and
  Xu}{Schubert et~al\mbox{.}}{2017}]%
        {schubert2017dbscan}
\bibfield{author}{\bibinfo{person}{Erich Schubert}, \bibinfo{person}{J{\"o}rg
  Sander}, \bibinfo{person}{Martin Ester}, \bibinfo{person}{Hans~Peter
  Kriegel}, {and} \bibinfo{person}{Xiaowei Xu}.}
  \bibinfo{year}{2017}\natexlab{}.
\newblock \showarticletitle{DBSCAN revisited, revisited: why and how you should
  (still) use DBSCAN}.
\newblock \bibinfo{journal}{\emph{ACM Transactions on Database Systems (TODS)}}
  \bibinfo{volume}{42}, \bibinfo{number}{3} (\bibinfo{year}{2017}),
  \bibinfo{pages}{1--21}.
\newblock


\bibitem[\protect\citeauthoryear{Shah and Koltun}{Shah and Koltun}{2017}]%
        {shah2017robust}
\bibfield{author}{\bibinfo{person}{Sohil~Atul Shah} {and}
  \bibinfo{person}{Vladlen Koltun}.} \bibinfo{year}{2017}\natexlab{}.
\newblock \showarticletitle{Robust continuous clustering}.
\newblock \bibinfo{journal}{\emph{Proceedings of the National Academy of
  Sciences}} \bibinfo{volume}{114}, \bibinfo{number}{37}
  (\bibinfo{year}{2017}), \bibinfo{pages}{9814--9819}.
\newblock


\bibitem[\protect\citeauthoryear{Shah and Koltun}{Shah and Koltun}{2018}]%
        {shah2018deep}
\bibfield{author}{\bibinfo{person}{Sohil~Atul Shah} {and}
  \bibinfo{person}{Vladlen Koltun}.} \bibinfo{year}{2018}\natexlab{}.
\newblock \showarticletitle{Deep continuous clustering}.
\newblock \bibinfo{journal}{\emph{arXiv preprint arXiv:1803.01449}}
  (\bibinfo{year}{2018}).
\newblock


\bibitem[\protect\citeauthoryear{Shen, Shen, Wang, Qin, Torr, and Shao}{Shen
  et~al\mbox{.}}{2021}]%
        {shen2021you}
\bibfield{author}{\bibinfo{person}{Yuming Shen}, \bibinfo{person}{Ziyi Shen},
  \bibinfo{person}{Menghan Wang}, \bibinfo{person}{Jie Qin},
  \bibinfo{person}{Philip Torr}, {and} \bibinfo{person}{Ling Shao}.}
  \bibinfo{year}{2021}\natexlab{}.
\newblock \showarticletitle{You never cluster alone}.
\newblock \bibinfo{journal}{\emph{Advances in Neural Information Processing
  Systems}}  \bibinfo{volume}{34} (\bibinfo{year}{2021}).
\newblock


\bibitem[\protect\citeauthoryear{Shen, Zhang, Zhang, Huang, Lu, and Lei}{Shen
  et~al\mbox{.}}{2018}]%
        {shen2018improving}
\bibfield{author}{\bibinfo{person}{Ying Shen}, \bibinfo{person}{Qiang Zhang},
  \bibinfo{person}{Jin Zhang}, \bibinfo{person}{Jiyue Huang},
  \bibinfo{person}{Yuming Lu}, {and} \bibinfo{person}{Kai Lei}.}
  \bibinfo{year}{2018}\natexlab{}.
\newblock \showarticletitle{Improving medical short text classification with
  semantic expansion using word-cluster embedding}. In
  \bibinfo{booktitle}{\emph{International Conference on Information Science and
  Applications}}. Springer, \bibinfo{pages}{401--411}.
\newblock


\bibitem[\protect\citeauthoryear{Shi and Malik}{Shi and Malik}{2000}]%
        {shi2000normalized}
\bibfield{author}{\bibinfo{person}{Jianbo Shi} {and} \bibinfo{person}{Jitendra
  Malik}.} \bibinfo{year}{2000}\natexlab{}.
\newblock \showarticletitle{Normalized cuts and image segmentation}.
\newblock \bibinfo{journal}{\emph{IEEE Transactions on pattern analysis and
  machine intelligence}} \bibinfo{volume}{22}, \bibinfo{number}{8}
  (\bibinfo{year}{2000}), \bibinfo{pages}{888--905}.
\newblock


\bibitem[\protect\citeauthoryear{Shun, Roosta-Khorasani, Fountoulakis, and
  Mahoney}{Shun et~al\mbox{.}}{2016}]%
        {shun2016parallel}
\bibfield{author}{\bibinfo{person}{Julian Shun}, \bibinfo{person}{Farbod
  Roosta-Khorasani}, \bibinfo{person}{Kimon Fountoulakis}, {and}
  \bibinfo{person}{Michael~W Mahoney}.} \bibinfo{year}{2016}\natexlab{}.
\newblock \showarticletitle{Parallel local graph clustering}.
\newblock \bibinfo{journal}{\emph{arXiv preprint arXiv:1604.07515}}
  (\bibinfo{year}{2016}).
\newblock


\bibitem[\protect\citeauthoryear{Song, Liu, Huang, Wang, and Tan}{Song
  et~al\mbox{.}}{2013}]%
        {song2013auto}
\bibfield{author}{\bibinfo{person}{Chunfeng Song}, \bibinfo{person}{Feng Liu},
  \bibinfo{person}{Yongzhen Huang}, \bibinfo{person}{Liang Wang}, {and}
  \bibinfo{person}{Tieniu Tan}.} \bibinfo{year}{2013}\natexlab{}.
\newblock \showarticletitle{Auto-encoder based data clustering}. In
  \bibinfo{booktitle}{\emph{Iberoamerican congress on pattern recognition}}.
  Springer, \bibinfo{pages}{117--124}.
\newblock


\bibitem[\protect\citeauthoryear{Srivastava, Mansimov, and
  Salakhudinov}{Srivastava et~al\mbox{.}}{2015}]%
        {srivastava2015unsupervised}
\bibfield{author}{\bibinfo{person}{Nitish Srivastava}, \bibinfo{person}{Elman
  Mansimov}, {and} \bibinfo{person}{Ruslan Salakhudinov}.}
  \bibinfo{year}{2015}\natexlab{}.
\newblock \showarticletitle{Unsupervised learning of video representations
  using lstms}. In \bibinfo{booktitle}{\emph{International conference on
  machine learning}}. PMLR, \bibinfo{pages}{843--852}.
\newblock


\bibitem[\protect\citeauthoryear{Steinbach, Karypis, and Kumar}{Steinbach
  et~al\mbox{.}}{2000}]%
        {steinbach2000comparison}
\bibfield{author}{\bibinfo{person}{Michael Steinbach}, \bibinfo{person}{George
  Karypis}, {and} \bibinfo{person}{Vipin Kumar}.}
  \bibinfo{year}{2000}\natexlab{}.
\newblock \showarticletitle{A comparison of document clustering techniques}.
\newblock  (\bibinfo{year}{2000}).
\newblock


\bibitem[\protect\citeauthoryear{Su, Xue, Liu, Wu, Yang, Zhou, Hu, Paris,
  Nepal, Jin, et~al\mbox{.}}{Su et~al\mbox{.}}{2022}]%
        {su2022comprehensive}
\bibfield{author}{\bibinfo{person}{Xing Su}, \bibinfo{person}{Shan Xue},
  \bibinfo{person}{Fanzhen Liu}, \bibinfo{person}{Jia Wu},
  \bibinfo{person}{Jian Yang}, \bibinfo{person}{Chuan Zhou},
  \bibinfo{person}{Wenbin Hu}, \bibinfo{person}{Cecile Paris},
  \bibinfo{person}{Surya Nepal}, \bibinfo{person}{Di Jin}, {et~al\mbox{.}}}
  \bibinfo{year}{2022}\natexlab{}.
\newblock \showarticletitle{A comprehensive survey on community detection with
  deep learning}.
\newblock \bibinfo{journal}{\emph{IEEE Transactions on Neural Networks and
  Learning Systems}} (\bibinfo{year}{2022}).
\newblock


\bibitem[\protect\citeauthoryear{Sundareswaran, Herrera-Gerena, Just, and
  Jannesari}{Sundareswaran et~al\mbox{.}}{2021}]%
        {sundareswaran2021cluster}
\bibfield{author}{\bibinfo{person}{Ramakrishnan Sundareswaran},
  \bibinfo{person}{Jansel Herrera-Gerena}, \bibinfo{person}{John Just}, {and}
  \bibinfo{person}{Ali Jannesari}.} \bibinfo{year}{2021}\natexlab{}.
\newblock \showarticletitle{Cluster Analysis with Deep Embeddings and
  Contrastive Learning}.
\newblock \bibinfo{journal}{\emph{arXiv preprint arXiv:2109.12714}}
  (\bibinfo{year}{2021}).
\newblock


\bibitem[\protect\citeauthoryear{Tang, Chen, and Jia}{Tang
  et~al\mbox{.}}{2020}]%
        {tang2020unsupervised}
\bibfield{author}{\bibinfo{person}{Hui Tang}, \bibinfo{person}{Ke Chen}, {and}
  \bibinfo{person}{Kui Jia}.} \bibinfo{year}{2020}\natexlab{}.
\newblock \showarticletitle{Unsupervised domain adaptation via structurally
  regularized deep clustering}. In \bibinfo{booktitle}{\emph{Proceedings of the
  IEEE/CVF conference on computer vision and pattern recognition}}.
  \bibinfo{pages}{8725--8735}.
\newblock


\bibitem[\protect\citeauthoryear{Tao, Takagi, and Nakata}{Tao
  et~al\mbox{.}}{2021}]%
        {tao2021clustering}
\bibfield{author}{\bibinfo{person}{Yaling Tao}, \bibinfo{person}{Kentaro
  Takagi}, {and} \bibinfo{person}{Kouta Nakata}.}
  \bibinfo{year}{2021}\natexlab{}.
\newblock \showarticletitle{Clustering-friendly representation learning via
  instance discrimination and feature decorrelation}.
\newblock \bibinfo{journal}{\emph{arXiv preprint arXiv:2106.00131}}
  (\bibinfo{year}{2021}).
\newblock


\bibitem[\protect\citeauthoryear{Thirumoorthy and Muneeswaran}{Thirumoorthy and
  Muneeswaran}{2021}]%
        {thirumoorthy2021hybrid}
\bibfield{author}{\bibinfo{person}{Karpagalingam Thirumoorthy} {and}
  \bibinfo{person}{Karuppaiah Muneeswaran}.} \bibinfo{year}{2021}\natexlab{}.
\newblock \showarticletitle{A hybrid approach for text document clustering
  using Jaya optimization algorithm}.
\newblock \bibinfo{journal}{\emph{Expert Systems with Applications}}
  \bibinfo{volume}{178} (\bibinfo{year}{2021}), \bibinfo{pages}{115040}.
\newblock


\bibitem[\protect\citeauthoryear{Tian, Gao, Cui, Chen, and Liu}{Tian
  et~al\mbox{.}}{2014}]%
        {tian2014learning}
\bibfield{author}{\bibinfo{person}{Fei Tian}, \bibinfo{person}{Bin Gao},
  \bibinfo{person}{Qing Cui}, \bibinfo{person}{Enhong Chen}, {and}
  \bibinfo{person}{Tie-Yan Liu}.} \bibinfo{year}{2014}\natexlab{}.
\newblock \showarticletitle{Learning deep representations for graph
  clustering}. In \bibinfo{booktitle}{\emph{Proceedings of the AAAI Conference
  on Artificial Intelligence}}, Vol.~\bibinfo{volume}{28}.
\newblock


\bibitem[\protect\citeauthoryear{Tian, Wan, Song, and Wei}{Tian
  et~al\mbox{.}}{2019}]%
        {tian2019clustering}
\bibfield{author}{\bibinfo{person}{Tian Tian}, \bibinfo{person}{Ji Wan},
  \bibinfo{person}{Qi Song}, {and} \bibinfo{person}{Zhi Wei}.}
  \bibinfo{year}{2019}\natexlab{}.
\newblock \showarticletitle{Clustering single-cell RNA-seq data with a
  model-based deep learning approach}.
\newblock \bibinfo{journal}{\emph{Nature Machine Intelligence}}
  \bibinfo{volume}{1}, \bibinfo{number}{4} (\bibinfo{year}{2019}),
  \bibinfo{pages}{191--198}.
\newblock


\bibitem[\protect\citeauthoryear{Tsai, Li, and Zhu}{Tsai et~al\mbox{.}}{2020}]%
        {tsai2020mice}
\bibfield{author}{\bibinfo{person}{Tsung~Wei Tsai}, \bibinfo{person}{Chongxuan
  Li}, {and} \bibinfo{person}{Jun Zhu}.} \bibinfo{year}{2020}\natexlab{}.
\newblock \showarticletitle{Mice: Mixture of contrastive experts for
  unsupervised image clustering}. In \bibinfo{booktitle}{\emph{International
  Conference on Learning Representations}}.
\newblock


\bibitem[\protect\citeauthoryear{Tseng}{Tseng}{2010}]%
        {tseng2010generic}
\bibfield{author}{\bibinfo{person}{Yuen-Hsien Tseng}.}
  \bibinfo{year}{2010}\natexlab{}.
\newblock \showarticletitle{Generic title labeling for clustered documents}.
\newblock \bibinfo{journal}{\emph{Expert Systems with Applications}}
  \bibinfo{volume}{37}, \bibinfo{number}{3} (\bibinfo{year}{2010}),
  \bibinfo{pages}{2247--2254}.
\newblock


\bibitem[\protect\citeauthoryear{Tsitsulin, Palowitch, Perozzi, and
  M{\"u}ller}{Tsitsulin et~al\mbox{.}}{2020}]%
        {tsitsulin2020graph}
\bibfield{author}{\bibinfo{person}{Anton Tsitsulin}, \bibinfo{person}{John
  Palowitch}, \bibinfo{person}{Bryan Perozzi}, {and} \bibinfo{person}{Emmanuel
  M{\"u}ller}.} \bibinfo{year}{2020}\natexlab{}.
\newblock \showarticletitle{Graph clustering with graph neural networks}.
\newblock \bibinfo{journal}{\emph{arXiv preprint arXiv:2006.16904}}
  (\bibinfo{year}{2020}).
\newblock


\bibitem[\protect\citeauthoryear{Usman, Jan, He, and Chen}{Usman
  et~al\mbox{.}}{2019}]%
        {usman2019survey}
\bibfield{author}{\bibinfo{person}{Muhammad Usman}, \bibinfo{person}{Mian~Ahmad
  Jan}, \bibinfo{person}{Xiangjian He}, {and} \bibinfo{person}{Jinjun Chen}.}
  \bibinfo{year}{2019}\natexlab{}.
\newblock \showarticletitle{A survey on representation learning efforts in
  cybersecurity domain}.
\newblock \bibinfo{journal}{\emph{ACM Computing Surveys (CSUR)}}
  \bibinfo{volume}{52}, \bibinfo{number}{6} (\bibinfo{year}{2019}),
  \bibinfo{pages}{1--28}.
\newblock


\bibitem[\protect\citeauthoryear{Van~der Maaten and Hinton}{Van~der Maaten and
  Hinton}{2008}]%
        {van2008visualizing}
\bibfield{author}{\bibinfo{person}{Laurens Van~der Maaten} {and}
  \bibinfo{person}{Geoffrey Hinton}.} \bibinfo{year}{2008}\natexlab{}.
\newblock \showarticletitle{Visualizing data using t-SNE.}
\newblock \bibinfo{journal}{\emph{Journal of machine learning research}}
  \bibinfo{volume}{9}, \bibinfo{number}{11} (\bibinfo{year}{2008}).
\newblock


\bibitem[\protect\citeauthoryear{Van~Gansbeke, Vandenhende, Georgoulis,
  Proesmans, and Van~Gool}{Van~Gansbeke et~al\mbox{.}}{2020}]%
        {van2020scan}
\bibfield{author}{\bibinfo{person}{Wouter Van~Gansbeke}, \bibinfo{person}{Simon
  Vandenhende}, \bibinfo{person}{Stamatios Georgoulis}, \bibinfo{person}{Marc
  Proesmans}, {and} \bibinfo{person}{Luc Van~Gool}.}
  \bibinfo{year}{2020}\natexlab{}.
\newblock \showarticletitle{Scan: Learning to classify images without labels}.
  In \bibinfo{booktitle}{\emph{European Conference on Computer Vision}}.
  Springer, \bibinfo{pages}{268--285}.
\newblock


\bibitem[\protect\citeauthoryear{Veli{\v{c}}kovi{\'c}, Cucurull, Casanova,
  Romero, Lio, and Bengio}{Veli{\v{c}}kovi{\'c} et~al\mbox{.}}{2017}]%
        {velivckovic2017graph}
\bibfield{author}{\bibinfo{person}{Petar Veli{\v{c}}kovi{\'c}},
  \bibinfo{person}{Guillem Cucurull}, \bibinfo{person}{Arantxa Casanova},
  \bibinfo{person}{Adriana Romero}, \bibinfo{person}{Pietro Lio}, {and}
  \bibinfo{person}{Yoshua Bengio}.} \bibinfo{year}{2017}\natexlab{}.
\newblock \showarticletitle{Graph attention networks}.
\newblock \bibinfo{journal}{\emph{arXiv preprint arXiv:1710.10903}}
  (\bibinfo{year}{2017}).
\newblock


\bibitem[\protect\citeauthoryear{Vidal}{Vidal}{2009}]%
        {vidal2009sparse}
\bibfield{author}{\bibinfo{person}{Ehsan Elhamifar~Ren{\'e} Vidal}.}
  \bibinfo{year}{2009}\natexlab{}.
\newblock \showarticletitle{Sparse subspace clustering}. In
  \bibinfo{booktitle}{\emph{2009 IEEE Conference on Computer Vision and Pattern
  Recognition (CVPR), vol. 00}}, Vol.~\bibinfo{volume}{6}.
  \bibinfo{pages}{2790--2797}.
\newblock


\bibitem[\protect\citeauthoryear{Von~Luxburg}{Von~Luxburg}{2007}]%
        {von2007tutorial}
\bibfield{author}{\bibinfo{person}{Ulrike Von~Luxburg}.}
  \bibinfo{year}{2007}\natexlab{}.
\newblock \showarticletitle{A tutorial on spectral clustering}.
\newblock \bibinfo{journal}{\emph{Statistics and computing}}
  \bibinfo{volume}{17}, \bibinfo{number}{4} (\bibinfo{year}{2007}),
  \bibinfo{pages}{395--416}.
\newblock


\bibitem[\protect\citeauthoryear{Wang, Cui, and Zhu}{Wang
  et~al\mbox{.}}{2016a}]%
        {wang2016structural}
\bibfield{author}{\bibinfo{person}{Daixin Wang}, \bibinfo{person}{Peng Cui},
  {and} \bibinfo{person}{Wenwu Zhu}.} \bibinfo{year}{2016}\natexlab{a}.
\newblock \showarticletitle{Structural deep network embedding}. In
  \bibinfo{booktitle}{\emph{Proceedings of the 22nd ACM SIGKDD international
  conference on Knowledge discovery and data mining}}.
  \bibinfo{pages}{1225--1234}.
\newblock


\bibitem[\protect\citeauthoryear{Wang, Liu, Wu, Cao, Meng, and Kennedy}{Wang
  et~al\mbox{.}}{2016b}]%
        {wang2016training}
\bibfield{author}{\bibinfo{person}{Shoujin Wang}, \bibinfo{person}{Wei Liu},
  \bibinfo{person}{Jia Wu}, \bibinfo{person}{Longbing Cao},
  \bibinfo{person}{Qinxue Meng}, {and} \bibinfo{person}{Paul~J Kennedy}.}
  \bibinfo{year}{2016}\natexlab{b}.
\newblock \showarticletitle{Training deep neural networks on imbalanced data
  sets}. In \bibinfo{booktitle}{\emph{2016 international joint conference on
  neural networks (IJCNN)}}. IEEE, \bibinfo{pages}{4368--4374}.
\newblock


\bibitem[\protect\citeauthoryear{Wang and Isola}{Wang and Isola}{2020}]%
        {wang2020understanding}
\bibfield{author}{\bibinfo{person}{Tongzhou Wang} {and}
  \bibinfo{person}{Phillip Isola}.} \bibinfo{year}{2020}\natexlab{}.
\newblock \showarticletitle{Understanding contrastive representation learning
  through alignment and uniformity on the hypersphere}. In
  \bibinfo{booktitle}{\emph{International Conference on Machine Learning}}.
  PMLR, \bibinfo{pages}{9929--9939}.
\newblock


\bibitem[\protect\citeauthoryear{Wang, Bao, and Guo}{Wang
  et~al\mbox{.}}{2022}]%
        {wang2022neural}
\bibfield{author}{\bibinfo{person}{Wenqing Wang}, \bibinfo{person}{Junpeng
  Bao}, {and} \bibinfo{person}{Siyao Guo}.} \bibinfo{year}{2022}\natexlab{}.
\newblock \showarticletitle{Neural generative model for clustering by
  separating particularity and commonality}.
\newblock \bibinfo{journal}{\emph{Information Sciences}}  \bibinfo{volume}{589}
  (\bibinfo{year}{2022}), \bibinfo{pages}{813--826}.
\newblock


\bibitem[\protect\citeauthoryear{Wang, Fan, Kuang, and Zhu}{Wang
  et~al\mbox{.}}{2021a}]%
        {wang2021explainable}
\bibfield{author}{\bibinfo{person}{Xin Wang}, \bibinfo{person}{Shuyi Fan},
  \bibinfo{person}{Kun Kuang}, {and} \bibinfo{person}{Wenwu Zhu}.}
  \bibinfo{year}{2021}\natexlab{a}.
\newblock \showarticletitle{Explainable automated graph representation learning
  with hyperparameter importance}. In \bibinfo{booktitle}{\emph{International
  Conference on Machine Learning}}. PMLR, \bibinfo{pages}{10727--10737}.
\newblock


\bibitem[\protect\citeauthoryear{Wang, Zhang, Wang, Yang, and Lin}{Wang
  et~al\mbox{.}}{2021b}]%
        {wang2021chaos}
\bibfield{author}{\bibinfo{person}{Yifei Wang}, \bibinfo{person}{Qi Zhang},
  \bibinfo{person}{Yisen Wang}, \bibinfo{person}{Jiansheng Yang}, {and}
  \bibinfo{person}{Zhouchen Lin}.} \bibinfo{year}{2021}\natexlab{b}.
\newblock \showarticletitle{Chaos is a ladder: A new understanding of
  contrastive learning}. In \bibinfo{booktitle}{\emph{International Conference
  on Learning Representations}}.
\newblock


\bibitem[\protect\citeauthoryear{Wang, Zou, and Zhang}{Wang
  et~al\mbox{.}}{2020}]%
        {wang2020cluster}
\bibfield{author}{\bibinfo{person}{Ziming Wang}, \bibinfo{person}{Yuexian Zou},
  {and} \bibinfo{person}{Zeming Zhang}.} \bibinfo{year}{2020}\natexlab{}.
\newblock \showarticletitle{Cluster attention contrast for video anomaly
  detection}. In \bibinfo{booktitle}{\emph{Proceedings of the 28th ACM
  International Conference on Multimedia}}. \bibinfo{pages}{2463--2471}.
\newblock


\bibitem[\protect\citeauthoryear{Wibisono, Anwar, Supriyanto, and
  Amin}{Wibisono et~al\mbox{.}}{2021}]%
        {wibisono2021multivariate}
\bibfield{author}{\bibinfo{person}{S Wibisono}, \bibinfo{person}{MT Anwar},
  \bibinfo{person}{A Supriyanto}, {and} \bibinfo{person}{IHA Amin}.}
  \bibinfo{year}{2021}\natexlab{}.
\newblock \showarticletitle{Multivariate weather anomaly detection using DBSCAN
  clustering algorithm}. In \bibinfo{booktitle}{\emph{Journal of Physics:
  Conference Series}}, Vol.~\bibinfo{volume}{1869}. IOP Publishing,
  \bibinfo{pages}{012077}.
\newblock


\bibitem[\protect\citeauthoryear{Wu, Long, Wang, Qian, Li, Lin, and Zha}{Wu
  et~al\mbox{.}}{2019}]%
        {wu2019deep}
\bibfield{author}{\bibinfo{person}{Jianlong Wu}, \bibinfo{person}{Keyu Long},
  \bibinfo{person}{Fei Wang}, \bibinfo{person}{Chen Qian},
  \bibinfo{person}{Cheng Li}, \bibinfo{person}{Zhouchen Lin}, {and}
  \bibinfo{person}{Hongbin Zha}.} \bibinfo{year}{2019}\natexlab{}.
\newblock \showarticletitle{Deep comprehensive correlation mining for image
  clustering}. In \bibinfo{booktitle}{\emph{Proceedings of the IEEE/CVF
  International Conference on Computer Vision}}. \bibinfo{pages}{8150--8159}.
\newblock


\bibitem[\protect\citeauthoryear{Wu, Sheng, Zhang, Li, Dadakova, Swisher, Cui,
  and Zhao}{Wu et~al\mbox{.}}{2020}]%
        {wu2020multi}
\bibfield{author}{\bibinfo{person}{Jian Wu}, \bibinfo{person}{Victor~S Sheng},
  \bibinfo{person}{Jing Zhang}, \bibinfo{person}{Hua Li},
  \bibinfo{person}{Tetiana Dadakova}, \bibinfo{person}{Christine~Leon Swisher},
  \bibinfo{person}{Zhiming Cui}, {and} \bibinfo{person}{Pengpeng Zhao}.}
  \bibinfo{year}{2020}\natexlab{}.
\newblock \showarticletitle{Multi-label active learning algorithms for image
  classification: Overview and future promise}.
\newblock \bibinfo{journal}{\emph{ACM Computing Surveys (CSUR)}}
  \bibinfo{volume}{53}, \bibinfo{number}{2} (\bibinfo{year}{2020}),
  \bibinfo{pages}{1--35}.
\newblock


\bibitem[\protect\citeauthoryear{Wu, Zhu, Zhang, and Philip}{Wu
  et~al\mbox{.}}{2014}]%
        {wu2014bag}
\bibfield{author}{\bibinfo{person}{Jia Wu}, \bibinfo{person}{Xingquan Zhu},
  \bibinfo{person}{Chengqi Zhang}, {and} \bibinfo{person}{S~Yu Philip}.}
  \bibinfo{year}{2014}\natexlab{}.
\newblock \showarticletitle{Bag constrained structure pattern mining for
  multi-graph classification}.
\newblock \bibinfo{journal}{\emph{Ieee transactions on knowledge and data
  engineering}} \bibinfo{volume}{26}, \bibinfo{number}{10}
  (\bibinfo{year}{2014}), \bibinfo{pages}{2382--2396}.
\newblock


\bibitem[\protect\citeauthoryear{Wu, Liu, Xie, Ester, and Yang}{Wu
  et~al\mbox{.}}{2016}]%
        {wu2016cccf}
\bibfield{author}{\bibinfo{person}{Yao Wu}, \bibinfo{person}{Xudong Liu},
  \bibinfo{person}{Min Xie}, \bibinfo{person}{Martin Ester}, {and}
  \bibinfo{person}{Qing Yang}.} \bibinfo{year}{2016}\natexlab{}.
\newblock \showarticletitle{CCCF: Improving collaborative filtering via
  scalable user-item co-clustering}. In \bibinfo{booktitle}{\emph{Proceedings
  of the ninth ACM international conference on web search and data mining}}.
  \bibinfo{pages}{73--82}.
\newblock


\bibitem[\protect\citeauthoryear{Xie, Girshick, and Farhadi}{Xie
  et~al\mbox{.}}{2016}]%
        {xie2016unsupervised}
\bibfield{author}{\bibinfo{person}{Junyuan Xie}, \bibinfo{person}{Ross
  Girshick}, {and} \bibinfo{person}{Ali Farhadi}.}
  \bibinfo{year}{2016}\natexlab{}.
\newblock \showarticletitle{Unsupervised deep embedding for clustering
  analysis}. In \bibinfo{booktitle}{\emph{International conference on machine
  learning}}. PMLR, \bibinfo{pages}{478--487}.
\newblock


\bibitem[\protect\citeauthoryear{Xie, Kelley, and Szymanski}{Xie
  et~al\mbox{.}}{2013}]%
        {xie2013overlapping}
\bibfield{author}{\bibinfo{person}{Jierui Xie}, \bibinfo{person}{Stephen
  Kelley}, {and} \bibinfo{person}{Boleslaw~K Szymanski}.}
  \bibinfo{year}{2013}\natexlab{}.
\newblock \showarticletitle{Overlapping community detection in networks: The
  state-of-the-art and comparative study}.
\newblock \bibinfo{journal}{\emph{Acm computing surveys (csur)}}
  \bibinfo{volume}{45}, \bibinfo{number}{4} (\bibinfo{year}{2013}),
  \bibinfo{pages}{1--35}.
\newblock


\bibitem[\protect\citeauthoryear{Xu and Tian}{Xu and Tian}{2015}]%
        {xu2015comprehensive}
\bibfield{author}{\bibinfo{person}{Dongkuan Xu} {and} \bibinfo{person}{Yingjie
  Tian}.} \bibinfo{year}{2015}\natexlab{}.
\newblock \showarticletitle{A comprehensive survey of clustering algorithms}.
\newblock \bibinfo{journal}{\emph{Annals of Data Science}} \bibinfo{volume}{2},
  \bibinfo{number}{2} (\bibinfo{year}{2015}), \bibinfo{pages}{165--193}.
\newblock


\bibitem[\protect\citeauthoryear{Xu and Wunsch}{Xu and Wunsch}{2005}]%
        {xu2005survey}
\bibfield{author}{\bibinfo{person}{Rui Xu} {and} \bibinfo{person}{Donald
  Wunsch}.} \bibinfo{year}{2005}\natexlab{}.
\newblock \showarticletitle{Survey of clustering algorithms}.
\newblock \bibinfo{journal}{\emph{IEEE Transactions on neural networks}}
  \bibinfo{volume}{16}, \bibinfo{number}{3} (\bibinfo{year}{2005}),
  \bibinfo{pages}{645--678}.
\newblock


\bibitem[\protect\citeauthoryear{Xu, Ester, Kriegel, and Sander}{Xu
  et~al\mbox{.}}{1998}]%
        {xu1998distribution}
\bibfield{author}{\bibinfo{person}{Xiaowei Xu}, \bibinfo{person}{Martin Ester},
  \bibinfo{person}{H-P Kriegel}, {and} \bibinfo{person}{J{\"o}rg Sander}.}
  \bibinfo{year}{1998}\natexlab{}.
\newblock \showarticletitle{A distribution-based clustering algorithm for
  mining in large spatial databases}. In \bibinfo{booktitle}{\emph{Proceedings
  14th International Conference on Data Engineering}}. IEEE,
  \bibinfo{pages}{324--331}.
\newblock


\bibitem[\protect\citeauthoryear{Xue, Lu, Wu, Zhang, and Xiong}{Xue
  et~al\mbox{.}}{2016}]%
        {xue2016multi}
\bibfield{author}{\bibinfo{person}{Shan Xue}, \bibinfo{person}{Jie Lu},
  \bibinfo{person}{Jia Wu}, \bibinfo{person}{Guangquan Zhang}, {and}
  \bibinfo{person}{Li Xiong}.} \bibinfo{year}{2016}\natexlab{}.
\newblock \showarticletitle{Multi-instance graphical transfer clustering for
  traffic data learning}. In \bibinfo{booktitle}{\emph{2016 International Joint
  Conference on Neural Networks (IJCNN)}}. IEEE, \bibinfo{pages}{4390--4395}.
\newblock


\bibitem[\protect\citeauthoryear{Yang, Fu, Sidiropoulos, and Hong}{Yang
  et~al\mbox{.}}{2017}]%
        {yang2017towards}
\bibfield{author}{\bibinfo{person}{Bo Yang}, \bibinfo{person}{Xiao Fu},
  \bibinfo{person}{Nicholas~D Sidiropoulos}, {and} \bibinfo{person}{Mingyi
  Hong}.} \bibinfo{year}{2017}\natexlab{}.
\newblock \showarticletitle{Towards k-means-friendly spaces: Simultaneous deep
  learning and clustering}. In \bibinfo{booktitle}{\emph{international
  conference on machine learning}}. PMLR, \bibinfo{pages}{3861--3870}.
\newblock


\bibitem[\protect\citeauthoryear{Yang, Lu, Lee, Batra, and Parikh}{Yang
  et~al\mbox{.}}{2018}]%
        {yang2018graph}
\bibfield{author}{\bibinfo{person}{Jianwei Yang}, \bibinfo{person}{Jiasen Lu},
  \bibinfo{person}{Stefan Lee}, \bibinfo{person}{Dhruv Batra}, {and}
  \bibinfo{person}{Devi Parikh}.} \bibinfo{year}{2018}\natexlab{}.
\newblock \showarticletitle{Graph r-cnn for scene graph generation}. In
  \bibinfo{booktitle}{\emph{Proceedings of the European conference on computer
  vision (ECCV)}}. \bibinfo{pages}{670--685}.
\newblock


\bibitem[\protect\citeauthoryear{Yang, Parikh, and Batra}{Yang
  et~al\mbox{.}}{2016}]%
        {yang2016joint}
\bibfield{author}{\bibinfo{person}{Jianwei Yang}, \bibinfo{person}{Devi
  Parikh}, {and} \bibinfo{person}{Dhruv Batra}.}
  \bibinfo{year}{2016}\natexlab{}.
\newblock \showarticletitle{Joint unsupervised learning of deep representations
  and image clusters}. In \bibinfo{booktitle}{\emph{Proceedings of the IEEE
  conference on computer vision and pattern recognition}}.
  \bibinfo{pages}{5147--5156}.
\newblock


\bibitem[\protect\citeauthoryear{Yang, Cheung, Li, and Fang}{Yang
  et~al\mbox{.}}{2019}]%
        {yang2019deep}
\bibfield{author}{\bibinfo{person}{Linxiao Yang}, \bibinfo{person}{Ngai-Man
  Cheung}, \bibinfo{person}{Jiaying Li}, {and} \bibinfo{person}{Jun Fang}.}
  \bibinfo{year}{2019}\natexlab{}.
\newblock \showarticletitle{Deep clustering by gaussian mixture variational
  autoencoders with graph embedding}. In \bibinfo{booktitle}{\emph{Proceedings
  of the IEEE/CVF International Conference on Computer Vision}}.
  \bibinfo{pages}{6440--6449}.
\newblock


\bibitem[\protect\citeauthoryear{Yi and Moon}{Yi and Moon}{2012}]%
        {yi2012image}
\bibfield{author}{\bibinfo{person}{Faliu Yi} {and} \bibinfo{person}{Inkyu
  Moon}.} \bibinfo{year}{2012}\natexlab{}.
\newblock \showarticletitle{Image segmentation: A survey of graph-cut methods}.
  In \bibinfo{booktitle}{\emph{2012 international conference on systems and
  informatics (ICSAI2012)}}. IEEE, \bibinfo{pages}{1936--1941}.
\newblock


\bibitem[\protect\citeauthoryear{You, Chen, Sui, Chen, Wang, and Shen}{You
  et~al\mbox{.}}{2020}]%
        {you2020graph}
\bibfield{author}{\bibinfo{person}{Yuning You}, \bibinfo{person}{Tianlong
  Chen}, \bibinfo{person}{Yongduo Sui}, \bibinfo{person}{Ting Chen},
  \bibinfo{person}{Zhangyang Wang}, {and} \bibinfo{person}{Yang Shen}.}
  \bibinfo{year}{2020}\natexlab{}.
\newblock \showarticletitle{Graph contrastive learning with augmentations}.
\newblock \bibinfo{journal}{\emph{Advances in Neural Information Processing
  Systems}}  \bibinfo{volume}{33} (\bibinfo{year}{2020}),
  \bibinfo{pages}{5812--5823}.
\newblock


\bibitem[\protect\citeauthoryear{Yu and Zhou}{Yu and Zhou}{2018}]%
        {yu2018mixture}
\bibfield{author}{\bibinfo{person}{Yang Yu} {and} \bibinfo{person}{Wen-Ji
  Zhou}.} \bibinfo{year}{2018}\natexlab{}.
\newblock \showarticletitle{Mixture of GANs for Clustering.}. In
  \bibinfo{booktitle}{\emph{IJCAI}}. \bibinfo{pages}{3047--3053}.
\newblock


\bibitem[\protect\citeauthoryear{Yue, Chen, Li, Zuo, and Yin}{Yue
  et~al\mbox{.}}{2019}]%
        {yue2019survey}
\bibfield{author}{\bibinfo{person}{Lin Yue}, \bibinfo{person}{Weitong Chen},
  \bibinfo{person}{Xue Li}, \bibinfo{person}{Wanli Zuo}, {and}
  \bibinfo{person}{Minghao Yin}.} \bibinfo{year}{2019}\natexlab{}.
\newblock \showarticletitle{A survey of sentiment analysis in social media}.
\newblock \bibinfo{journal}{\emph{Knowledge and Information Systems}}
  \bibinfo{volume}{60}, \bibinfo{number}{2} (\bibinfo{year}{2019}),
  \bibinfo{pages}{617--663}.
\newblock


\bibitem[\protect\citeauthoryear{Zhai, Zhang, Chen, and He}{Zhai
  et~al\mbox{.}}{2018}]%
        {zhai2018autoencoder}
\bibfield{author}{\bibinfo{person}{Junhai Zhai}, \bibinfo{person}{Sufang
  Zhang}, \bibinfo{person}{Junfen Chen}, {and} \bibinfo{person}{Qiang He}.}
  \bibinfo{year}{2018}\natexlab{}.
\newblock \showarticletitle{Autoencoder and its various variants}. In
  \bibinfo{booktitle}{\emph{2018 IEEE International Conference on Systems, Man,
  and Cybernetics (SMC)}}. IEEE, \bibinfo{pages}{415--419}.
\newblock


\bibitem[\protect\citeauthoryear{Zhai, Lu, Ye, Shan, Chen, Ji, and Tian}{Zhai
  et~al\mbox{.}}{2020}]%
        {zhai2020ad}
\bibfield{author}{\bibinfo{person}{Yunpeng Zhai}, \bibinfo{person}{Shijian Lu},
  \bibinfo{person}{Qixiang Ye}, \bibinfo{person}{Xuebo Shan},
  \bibinfo{person}{Jie Chen}, \bibinfo{person}{Rongrong Ji}, {and}
  \bibinfo{person}{Yonghong Tian}.} \bibinfo{year}{2020}\natexlab{}.
\newblock \showarticletitle{Ad-cluster: Augmented discriminative clustering for
  domain adaptive person re-identification}. In
  \bibinfo{booktitle}{\emph{Proceedings of the IEEE/CVF Conference on Computer
  Vision and Pattern Recognition}}. \bibinfo{pages}{9021--9030}.
\newblock


\bibitem[\protect\citeauthoryear{Zhang, Nan, Wei, Li, Zhu, McKeown, Nallapati,
  Arnold, and Xiang}{Zhang et~al\mbox{.}}{2021b}]%
        {zhang2021supporting}
\bibfield{author}{\bibinfo{person}{Dejiao Zhang}, \bibinfo{person}{Feng Nan},
  \bibinfo{person}{Xiaokai Wei}, \bibinfo{person}{Shangwen Li},
  \bibinfo{person}{Henghui Zhu}, \bibinfo{person}{Kathleen McKeown},
  \bibinfo{person}{Ramesh Nallapati}, \bibinfo{person}{Andrew Arnold}, {and}
  \bibinfo{person}{Bing Xiang}.} \bibinfo{year}{2021}\natexlab{b}.
\newblock \showarticletitle{Supporting clustering with contrastive learning}.
\newblock \bibinfo{journal}{\emph{arXiv preprint arXiv:2103.12953}}
  (\bibinfo{year}{2021}).
\newblock


\bibitem[\protect\citeauthoryear{Zhang and Davidson}{Zhang and
  Davidson}{2021}]%
        {zhang2021deep}
\bibfield{author}{\bibinfo{person}{Hongjing Zhang} {and} \bibinfo{person}{Ian
  Davidson}.} \bibinfo{year}{2021}\natexlab{}.
\newblock \showarticletitle{Deep Descriptive Clustering}.
\newblock \bibinfo{journal}{\emph{arXiv preprint arXiv:2105.11549}}
  (\bibinfo{year}{2021}).
\newblock


\bibitem[\protect\citeauthoryear{Zhang, Chang, Yu, and Dhillon}{Zhang
  et~al\mbox{.}}{2021a}]%
        {zhang2021fast}
\bibfield{author}{\bibinfo{person}{Jiong Zhang}, \bibinfo{person}{Wei-cheng
  Chang}, \bibinfo{person}{Hsiang-fu Yu}, {and} \bibinfo{person}{Inderjit
  Dhillon}.} \bibinfo{year}{2021}\natexlab{a}.
\newblock \showarticletitle{Fast multi-resolution transformer fine-tuning for
  extreme multi-label text classification}.
\newblock \bibinfo{journal}{\emph{Advances in Neural Information Processing
  Systems}}  \bibinfo{volume}{34} (\bibinfo{year}{2021}).
\newblock


\bibitem[\protect\citeauthoryear{Zhang, Li, You, Qi, Zhang, Guo, and Lin}{Zhang
  et~al\mbox{.}}{2019c}]%
        {zhang2019self}
\bibfield{author}{\bibinfo{person}{Junjian Zhang}, \bibinfo{person}{Chun-Guang
  Li}, \bibinfo{person}{Chong You}, \bibinfo{person}{Xianbiao Qi},
  \bibinfo{person}{Honggang Zhang}, \bibinfo{person}{Jun Guo}, {and}
  \bibinfo{person}{Zhouchen Lin}.} \bibinfo{year}{2019}\natexlab{c}.
\newblock \showarticletitle{Self-supervised convolutional subspace clustering
  network}. In \bibinfo{booktitle}{\emph{Proceedings of the IEEE/CVF Conference
  on Computer Vision and Pattern Recognition}}. \bibinfo{pages}{5473--5482}.
\newblock


\bibitem[\protect\citeauthoryear{Zhang and Zhou}{Zhang and Zhou}{2013}]%
        {zhang2013review}
\bibfield{author}{\bibinfo{person}{Min-Ling Zhang} {and}
  \bibinfo{person}{Zhi-Hua Zhou}.} \bibinfo{year}{2013}\natexlab{}.
\newblock \showarticletitle{A review on multi-label learning algorithms}.
\newblock \bibinfo{journal}{\emph{IEEE transactions on knowledge and data
  engineering}} \bibinfo{volume}{26}, \bibinfo{number}{8}
  (\bibinfo{year}{2013}), \bibinfo{pages}{1819--1837}.
\newblock


\bibitem[\protect\citeauthoryear{Zhang, Wu, Yang, Tian, and Zhang}{Zhang
  et~al\mbox{.}}{2016}]%
        {zhang2016unsupervised}
\bibfield{author}{\bibinfo{person}{Qin Zhang}, \bibinfo{person}{Jia Wu},
  \bibinfo{person}{Hong Yang}, \bibinfo{person}{Yingjie Tian}, {and}
  \bibinfo{person}{Chengqi Zhang}.} \bibinfo{year}{2016}\natexlab{}.
\newblock \showarticletitle{Unsupervised Feature Learning from Time Series.}.
  In \bibinfo{booktitle}{\emph{IJCAI}}. New York, USA,
  \bibinfo{pages}{2322--2328}.
\newblock


\bibitem[\protect\citeauthoryear{Zhang, Wu, Zhang, Long, and Zhang}{Zhang
  et~al\mbox{.}}{2018a}]%
        {zhang2018salient}
\bibfield{author}{\bibinfo{person}{Qin Zhang}, \bibinfo{person}{Jia Wu},
  \bibinfo{person}{Peng Zhang}, \bibinfo{person}{Guodong Long}, {and}
  \bibinfo{person}{Chengqi Zhang}.} \bibinfo{year}{2018}\natexlab{a}.
\newblock \showarticletitle{Salient subsequence learning for time series
  clustering}.
\newblock \bibinfo{journal}{\emph{IEEE transactions on pattern analysis and
  machine intelligence}} \bibinfo{volume}{41}, \bibinfo{number}{9}
  (\bibinfo{year}{2018}), \bibinfo{pages}{2193--2207}.
\newblock


\bibitem[\protect\citeauthoryear{Zhang, You, Vidal, and Li}{Zhang
  et~al\mbox{.}}{2021c}]%
        {zhang2021learning}
\bibfield{author}{\bibinfo{person}{Shangzhi Zhang}, \bibinfo{person}{Chong
  You}, \bibinfo{person}{Ren{\'e} Vidal}, {and} \bibinfo{person}{Chun-Guang
  Li}.} \bibinfo{year}{2021}\natexlab{c}.
\newblock \showarticletitle{Learning a self-expressive network for subspace
  clustering}. In \bibinfo{booktitle}{\emph{Proceedings of the IEEE/CVF
  Conference on Computer Vision and Pattern Recognition}}.
  \bibinfo{pages}{12393--12403}.
\newblock


\bibitem[\protect\citeauthoryear{Zhang, Ji, Harandi, Huang, and Li}{Zhang
  et~al\mbox{.}}{2019b}]%
        {zhang2019neural}
\bibfield{author}{\bibinfo{person}{Tong Zhang}, \bibinfo{person}{Pan Ji},
  \bibinfo{person}{Mehrtash Harandi}, \bibinfo{person}{Wenbing Huang}, {and}
  \bibinfo{person}{Hongdong Li}.} \bibinfo{year}{2019}\natexlab{b}.
\newblock \showarticletitle{Neural collaborative subspace clustering}. In
  \bibinfo{booktitle}{\emph{International Conference on Machine Learning}}.
  PMLR, \bibinfo{pages}{7384--7393}.
\newblock


\bibitem[\protect\citeauthoryear{Zhang, Ramakrishnan, and Livny}{Zhang
  et~al\mbox{.}}{1996}]%
        {zhang1996birch}
\bibfield{author}{\bibinfo{person}{Tian Zhang}, \bibinfo{person}{Raghu
  Ramakrishnan}, {and} \bibinfo{person}{Miron Livny}.}
  \bibinfo{year}{1996}\natexlab{}.
\newblock \showarticletitle{BIRCH: an efficient data clustering method for very
  large databases}.
\newblock \bibinfo{journal}{\emph{ACM sigmod record}} \bibinfo{volume}{25},
  \bibinfo{number}{2} (\bibinfo{year}{1996}), \bibinfo{pages}{103--114}.
\newblock


\bibitem[\protect\citeauthoryear{Zhang, Bu, Ester, Zhang, Yao, Li, and
  Wang}{Zhang et~al\mbox{.}}{2020a}]%
        {zhang2020learning}
\bibfield{author}{\bibinfo{person}{Zhen Zhang}, \bibinfo{person}{Jiajun Bu},
  \bibinfo{person}{Martin Ester}, \bibinfo{person}{Jianfeng Zhang},
  \bibinfo{person}{Chengwei Yao}, \bibinfo{person}{Zhao Li}, {and}
  \bibinfo{person}{Can Wang}.} \bibinfo{year}{2020}\natexlab{a}.
\newblock \showarticletitle{Learning temporal interaction graph embedding via
  coupled memory networks}. In \bibinfo{booktitle}{\emph{Proceedings of the web
  conference 2020}}. \bibinfo{pages}{3049--3055}.
\newblock


\bibitem[\protect\citeauthoryear{Zhang, Bu, Ester, Zhang, Yao, Yu, and
  Wang}{Zhang et~al\mbox{.}}{2019a}]%
        {zhang2019hierarchical}
\bibfield{author}{\bibinfo{person}{Zhen Zhang}, \bibinfo{person}{Jiajun Bu},
  \bibinfo{person}{Martin Ester}, \bibinfo{person}{Jianfeng Zhang},
  \bibinfo{person}{Chengwei Yao}, \bibinfo{person}{Zhi Yu}, {and}
  \bibinfo{person}{Can Wang}.} \bibinfo{year}{2019}\natexlab{a}.
\newblock \showarticletitle{Hierarchical graph pooling with structure
  learning}.
\newblock \bibinfo{journal}{\emph{arXiv preprint arXiv:1911.05954}}
  (\bibinfo{year}{2019}).
\newblock


\bibitem[\protect\citeauthoryear{Zhang, Cui, and Zhu}{Zhang
  et~al\mbox{.}}{2020b}]%
        {zhang2020deep}
\bibfield{author}{\bibinfo{person}{Ziwei Zhang}, \bibinfo{person}{Peng Cui},
  {and} \bibinfo{person}{Wenwu Zhu}.} \bibinfo{year}{2020}\natexlab{b}.
\newblock \showarticletitle{Deep learning on graphs: A survey}.
\newblock \bibinfo{journal}{\emph{IEEE Transactions on Knowledge and Data
  Engineering}} (\bibinfo{year}{2020}).
\newblock


\bibitem[\protect\citeauthoryear{Zhang, Yang, Bu, Zhou, Yu, Zhang, Ester, and
  Wang}{Zhang et~al\mbox{.}}{2018b}]%
        {zhang2018anrl}
\bibfield{author}{\bibinfo{person}{Zhen Zhang}, \bibinfo{person}{Hongxia Yang},
  \bibinfo{person}{Jiajun Bu}, \bibinfo{person}{Sheng Zhou},
  \bibinfo{person}{Pinggang Yu}, \bibinfo{person}{Jianwei Zhang},
  \bibinfo{person}{Martin Ester}, {and} \bibinfo{person}{Can Wang}.}
  \bibinfo{year}{2018}\natexlab{b}.
\newblock \showarticletitle{ANRL: Attributed Network Representation Learning
  via Deep Neural Networks.}. In \bibinfo{booktitle}{\emph{Ijcai}},
  Vol.~\bibinfo{volume}{18}. \bibinfo{pages}{3155--3161}.
\newblock


\bibitem[\protect\citeauthoryear{Zhong, Chen, Jin, and Hua}{Zhong
  et~al\mbox{.}}{2020}]%
        {zhong2020deep}
\bibfield{author}{\bibinfo{person}{Huasong Zhong}, \bibinfo{person}{Chong
  Chen}, \bibinfo{person}{Zhongming Jin}, {and} \bibinfo{person}{Xian-Sheng
  Hua}.} \bibinfo{year}{2020}\natexlab{}.
\newblock \showarticletitle{Deep robust clustering by contrastive learning}.
\newblock \bibinfo{journal}{\emph{arXiv preprint arXiv:2008.03030}}
  (\bibinfo{year}{2020}).
\newblock


\bibitem[\protect\citeauthoryear{Zhong, Wu, Chen, Huang, Deng, Nie, Lin, and
  Hua}{Zhong et~al\mbox{.}}{2021}]%
        {zhong2021graph}
\bibfield{author}{\bibinfo{person}{Huasong Zhong}, \bibinfo{person}{Jianlong
  Wu}, \bibinfo{person}{Chong Chen}, \bibinfo{person}{Jianqiang Huang},
  \bibinfo{person}{Minghua Deng}, \bibinfo{person}{Liqiang Nie},
  \bibinfo{person}{Zhouchen Lin}, {and} \bibinfo{person}{Xian-Sheng Hua}.}
  \bibinfo{year}{2021}\natexlab{}.
\newblock \showarticletitle{Graph contrastive clustering}. In
  \bibinfo{booktitle}{\emph{Proceedings of the IEEE/CVF International
  Conference on Computer Vision}}. \bibinfo{pages}{9224--9233}.
\newblock


\bibitem[\protect\citeauthoryear{Zhou, Liu, Wei, and Fan}{Zhou
  et~al\mbox{.}}{2022}]%
        {zhou2022network}
\bibfield{author}{\bibinfo{person}{Jingya Zhou}, \bibinfo{person}{Ling Liu},
  \bibinfo{person}{Wenqi Wei}, {and} \bibinfo{person}{Jianxi Fan}.}
  \bibinfo{year}{2022}\natexlab{}.
\newblock \showarticletitle{Network Representation Learning: From
  Preprocessing, Feature Extraction to Node Embedding}.
\newblock \bibinfo{journal}{\emph{ACM Computing Surveys (CSUR)}}
  \bibinfo{volume}{55}, \bibinfo{number}{2} (\bibinfo{year}{2022}),
  \bibinfo{pages}{1--35}.
\newblock


\bibitem[\protect\citeauthoryear{Zhou, Hou, and Feng}{Zhou
  et~al\mbox{.}}{2018}]%
        {zhou2018deep}
\bibfield{author}{\bibinfo{person}{Pan Zhou}, \bibinfo{person}{Yunqing Hou},
  {and} \bibinfo{person}{Jiashi Feng}.} \bibinfo{year}{2018}\natexlab{}.
\newblock \showarticletitle{Deep adversarial subspace clustering}. In
  \bibinfo{booktitle}{\emph{Proceedings of the IEEE Conference on Computer
  Vision and Pattern Recognition}}. \bibinfo{pages}{1596--1604}.
\newblock


\bibitem[\protect\citeauthoryear{Zhou, Wang, et~al\mbox{.}}{Zhou
  et~al\mbox{.}}{2021b}]%
        {zhou2021cluster}
\bibfield{author}{\bibinfo{person}{Qiang Zhou}, \bibinfo{person}{Shirui Wang},
  {et~al\mbox{.}}} \bibinfo{year}{2021}\natexlab{b}.
\newblock \showarticletitle{Cluster adaptation networks for unsupervised domain
  adaptation}.
\newblock \bibinfo{journal}{\emph{Image and Vision Computing}}
  \bibinfo{volume}{108} (\bibinfo{year}{2021}), \bibinfo{pages}{104137}.
\newblock


\bibitem[\protect\citeauthoryear{Zhou, Bu, Wang, Chen, and Wang}{Zhou
  et~al\mbox{.}}{2019}]%
        {zhou2019hahe}
\bibfield{author}{\bibinfo{person}{Sheng Zhou}, \bibinfo{person}{Jiajun Bu},
  \bibinfo{person}{Xin Wang}, \bibinfo{person}{Jiawei Chen}, {and}
  \bibinfo{person}{Can Wang}.} \bibinfo{year}{2019}\natexlab{}.
\newblock \showarticletitle{HAHE: Hierarchical attentive heterogeneous
  information network embedding}.
\newblock \bibinfo{journal}{\emph{arXiv preprint arXiv:1902.01475}}
  (\bibinfo{year}{2019}).
\newblock


\bibitem[\protect\citeauthoryear{Zhou, Bu, Zhang, Wang, Ma, and Zhang}{Zhou
  et~al\mbox{.}}{2020a}]%
        {zhou2020cross}
\bibfield{author}{\bibinfo{person}{Sheng Zhou}, \bibinfo{person}{Jiajun Bu},
  \bibinfo{person}{Zhen Zhang}, \bibinfo{person}{Can Wang},
  \bibinfo{person}{Lingzhou Ma}, {and} \bibinfo{person}{Jianfeng Zhang}.}
  \bibinfo{year}{2020}\natexlab{a}.
\newblock \showarticletitle{Cross multi-type objects clustering in attributed
  heterogeneous information network}.
\newblock \bibinfo{journal}{\emph{Knowledge-Based Systems}}
  \bibinfo{volume}{194} (\bibinfo{year}{2020}), \bibinfo{pages}{105458}.
\newblock


\bibitem[\protect\citeauthoryear{Zhou, Wang, Bu, Ester, Yu, Chen, Shi, and
  Wang}{Zhou et~al\mbox{.}}{2020b}]%
        {zhou2020dge}
\bibfield{author}{\bibinfo{person}{Sheng Zhou}, \bibinfo{person}{Xin Wang},
  \bibinfo{person}{Jiajun Bu}, \bibinfo{person}{Martin Ester},
  \bibinfo{person}{Pinggang Yu}, \bibinfo{person}{Jiawei Chen},
  \bibinfo{person}{Qihao Shi}, {and} \bibinfo{person}{Can Wang}.}
  \bibinfo{year}{2020}\natexlab{b}.
\newblock \showarticletitle{DGE: Deep generative network embedding based on
  commonality and individuality}. In \bibinfo{booktitle}{\emph{Proceedings of
  the AAAI Conference on Artificial Intelligence}}, Vol.~\bibinfo{volume}{34}.
  \bibinfo{pages}{6949--6956}.
\newblock


\bibitem[\protect\citeauthoryear{Zhou, Wang, Chen, Chen, Wang, Wang, and
  Bu}{Zhou et~al\mbox{.}}{2021a}]%
        {Zhou_2021_ICCV}
\bibfield{author}{\bibinfo{person}{Sheng Zhou}, \bibinfo{person}{Yucheng Wang},
  \bibinfo{person}{Defang Chen}, \bibinfo{person}{Jiawei Chen},
  \bibinfo{person}{Xin Wang}, \bibinfo{person}{Can Wang}, {and}
  \bibinfo{person}{Jiajun Bu}.} \bibinfo{year}{2021}\natexlab{a}.
\newblock \showarticletitle{Distilling Holistic Knowledge With Graph Neural
  Networks}. In \bibinfo{booktitle}{\emph{Proceedings of the IEEE/CVF
  International Conference on Computer Vision (ICCV)}}.
  \bibinfo{pages}{10387--10396}.
\newblock


\bibitem[\protect\citeauthoryear{Zhu, Galstyan, Cheng, and Lerman}{Zhu
  et~al\mbox{.}}{2014}]%
        {zhu2014tripartite}
\bibfield{author}{\bibinfo{person}{Linhong Zhu}, \bibinfo{person}{Aram
  Galstyan}, \bibinfo{person}{James Cheng}, {and} \bibinfo{person}{Kristina
  Lerman}.} \bibinfo{year}{2014}\natexlab{}.
\newblock \showarticletitle{Tripartite graph clustering for dynamic sentiment
  analysis on social media}. In \bibinfo{booktitle}{\emph{Proceedings of the
  2014 ACM SIGMOD international conference on Management of data}}.
  \bibinfo{pages}{1531--1542}.
\newblock


\bibitem[\protect\citeauthoryear{Zhu, Wang, and Cui}{Zhu et~al\mbox{.}}{2020}]%
        {zhu2020deep}
\bibfield{author}{\bibinfo{person}{Wenwu Zhu}, \bibinfo{person}{Xin Wang},
  {and} \bibinfo{person}{Peng Cui}.} \bibinfo{year}{2020}\natexlab{}.
\newblock \showarticletitle{Deep learning for learning graph representations}.
\newblock In \bibinfo{booktitle}{\emph{Deep learning: concepts and
  architectures}}. \bibinfo{publisher}{Springer}, \bibinfo{pages}{169--210}.
\newblock


\end{thebibliography}
